\documentclass[journal,compsoc,cspaper,final]{IEEEtran}
\usepackage{xspace}
\usepackage{graphicx}
\usepackage{amsmath}
\usepackage{amssymb}
\usepackage{booktabs}
\usepackage{caption}
\usepackage[nocompress]{cite}
\usepackage{algorithm}
\usepackage{algorithmic}
\usepackage{makecell}
\usepackage{tikz}
\usepackage{pifont}
\usetikzlibrary{spy}
\usepackage{comment}
\usepackage{tabularx}
\usepackage{amsmath,amssymb} 
\usepackage{multirow}
\usepackage[pagebackref,breaklinks,colorlinks]{hyperref}
\usepackage{xcolor}
\usepackage[export]{adjustbox}
\usepackage{graphicx}
\usepackage{amsmath}
\usepackage{amssymb}
\usepackage{boxedminipage}
\usepackage{booktabs}
\usepackage{colortbl}
\usepackage{algorithm}
\usepackage{algorithmic}
\usepackage{stfloats}
\usepackage{makecell}
\usepackage{flushend} 
\usepackage{tikz}
\usetikzlibrary{spy}
\usepackage{comment}
\usepackage{ragged2e}
\usepackage{amsmath,amssymb}
\usepackage[capitalize]{cleveref}
\crefname{section}{Sec.}{Secs.}
\Crefname{section}{Section}{Sections}
\Crefname{table}{Table}{Tables}
\crefname{table}{Tab.}{Tabs.}

\newcommand{\ie}{{\emph{i.e.}}\xspace}
\newcommand{\vs}{{\emph{vs.}}\xspace}

\newcommand{\eg}{{\emph{e.g.}}\xspace}

\newcommand{\etal}{{\emph{et al.}}}

\newcommand{\website}{https://w3un.github.io/selfunroll/}

\newcommand{\lei}[1]{\textcolor{black}{#1}}
\newcommand{\zhang}[1]{\textcolor{black}{#1}}
\newcommand{\wang}[1]{\textcolor{black}{#1}}
\newcommand{\lin}[1]{\textcolor{black}{#1}}
\newcommand{\myname}[0]{SelfUnroll}

\begin{document}
%


\title{Self-Supervised Scene Dynamic Recovery from \\ Rolling Shutter Images and Events}

%
%
%
%

\author{Yangguang Wang, Xiang Zhang, Mingyuan Lin, Lei Yu, Boxin Shi, Wen Yang, and Gui-Song Xia
        \IEEEcompsocitemizethanks{
\IEEEcompsocthanksitem Y. Wang, X. Zhang, M. Lin, L. Yu, and W. Yang are with the School of Electronic Information, Wuhan University, Wuhan 430072, China. E-mail: \{wyg.eis, xiangz, linmingyuan, ly.wd, yangwen\}@whu.edu.cn.
\IEEEcompsocthanksitem G.-S. Xia is with the School of Computer Science, Wuhan University, Wuhan 430072, China. E-mail: guisong.xia@whu.edu.cn.
\IEEEcompsocthanksitem B. Shi is with the National Key Laboratory for Multimedia Information Processing and the National Engineering Research Center of Visual Technology, School of Computer Science, Peking University. E-mail: shiboxin@pku.edu.cn.
\IEEEcompsocthanksitem The research was partially supported by the National Natural Science Foundation of China under Grants 62271354 and 61871297, and the Natural Science Foundation of Hubei Province, China under Grant 2021CFB467.
\IEEEcompsocthanksitem Corresponding authors: L. Yu and G.-S. Xia.
}
}

\markboth{IEEE TRANSACTIONS ON PATTERN ANALYSIS AND MACHINE INTELLIGENCE}{}
%



\IEEEtitleabstractindextext{%
\begin{abstract}
\justifying
Scene Dynamic Recovery (SDR) by inverting distorted Rolling Shutter (RS) images to an undistorted high frame-rate Global Shutter (GS) video is a severely ill-posed problem 
due to the missing temporal dynamic information in both RS intra-frame scanlines and inter-frame exposures,
particularly when prior knowledge about camera/object motions is unavailable.
Commonly used artificial assumptions on scenes/motions and data-specific characteristics
are prone to producing sub-optimal solutions in real-world scenarios.
To address this challenge, we propose an event-based SDR network within a self-supervised learning paradigm, \ie, {\it SelfUnroll}. We leverage the extremely high temporal resolution of event cameras to provide accurate inter/intra-frame dynamic information. 
Specifically, an Event-based Inter/intra-frame Compensator (E-IC) is proposed to predict the per-pixel dynamic between arbitrary time intervals, including the temporal transition and spatial translation. Exploring connections in terms of RS-RS, RS-GS, and GS-RS, we explicitly formulate mutual constraints with the proposed E-IC, resulting in supervisions without ground-truth GS images. Extensive evaluations over synthetic and real datasets demonstrate that the proposed method achieves state-of-the-art and shows remarkable performance for event-based RS2GS inversion in real-world scenarios. The dataset and code are available at \url{\website}.

%
%


%

\end{abstract}

}

\maketitle

\IEEEdisplaynontitleabstractindextext

%
\IEEEpeerreviewmaketitle

\IEEEraisesectionheading{\section{Introduction}\label{sec:introduction}}

%
%
%
%
%


\IEEEPARstart{T}{he} row-by-row exposure mechanism in the Rolling Shutter (RS) camera significantly reduces the data transfer rate, making it an affordable solution for high-speed imaging~\cite{zhong2022bringing,choi2022analysis,sheinin2022dual}. However, such an approach produces so-called RS effects, which appear as spatial distortions in dynamic scenes (\eg, as wobble and skew, as shown in \cref{mul_image_real}), especially when high-speed camera-to-object motions are involved~\cite{baker2010removing}. Although correcting RS effects is essential for practical applications, it is not sufficient to completely capture the underlying scene dynamics. Alternatively, inverting the distorted RS observations to an undistorted continuous-time {Global Shutter (GS) video} can achieve the full Scene Dynamic Recovery (SDR) \cite{xue2016visual}, which however is inherently challenging due to the missing temporal dynamic information in both RS intra-frame scanlines and inter-frame exposures.

%

%

%
%
%

%
%
%

%
%

 %

\def\imwidth{0.196}
\def\cimwid{0.1345}
\def\ccimwid{0.1313}
\begin{figure}[t]
\footnotesize
	\centering

    \begin{minipage}[t]{\imwidth\linewidth}
    		\centering
			\begin{tikzpicture}[spy using outlines={rectangle,green,magnification=\ssmag,size=\ssizz},inner sep=0]
				\node {\includegraphics[width=\linewidth]{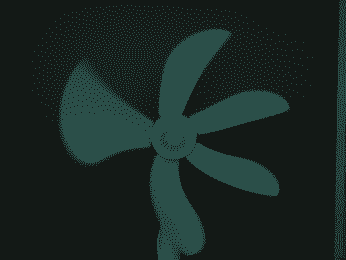}};
				\end{tikzpicture}
    \vspace{0.5em}
            RS image \vspace{0.5em}
    	\end{minipage}%
    \hfill
    	\begin{minipage}[t]{\imwidth\linewidth}
    		\centering
    		\begin{tikzpicture}[spy using outlines={rectangle,green,magnification=\ssmag,size=\ssizz},inner sep=0]
				\node {\includegraphics[width=\linewidth]{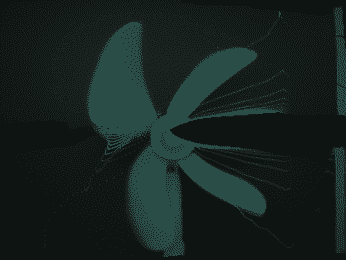}};
				\end{tikzpicture}

			(a) RSSR\vspace{0.5em}
    	\end{minipage}%
    \hfill
    	\begin{minipage}[t]{\imwidth\linewidth}
    		\centering
    	    \begin{tikzpicture}[spy using outlines={green,magnification=\ssmag,size=\ssizz},inner sep=0]
				\node {\includegraphics[width=\linewidth]{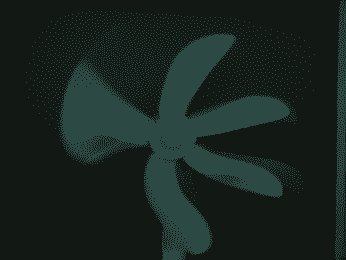}};
				\end{tikzpicture}
			(b) CVR\vspace{0.5em}
    	\end{minipage}%
    \hfill
    	\begin{minipage}[t]{\imwidth\linewidth}
    		\centering
    	    \begin{tikzpicture}[spy using outlines={green,magnification=\ssmag,size=\ssizz},inner sep=0]
				\node {\includegraphics[width=\linewidth]{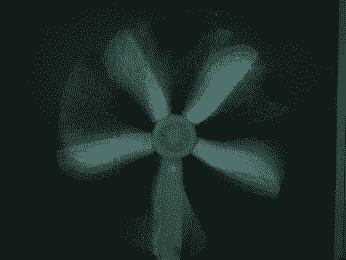}};
				\end{tikzpicture}
			(c) EvUnroll\vspace{0.5em}
    	\end{minipage}%
    \hfill
		\begin{minipage}[t]{\imwidth\linewidth}
			\centering
			\begin{tikzpicture}[spy using outlines={green,magnification=\ssmag,size=\ssizz},inner sep=0]
				\node {\includegraphics[width=\linewidth]{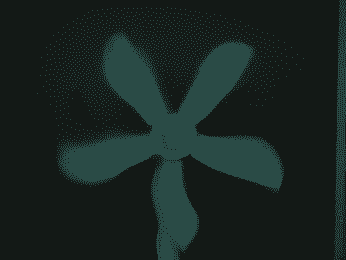}};
				\end{tikzpicture}
			 {\bf (d) Ours}\vspace{0.5em}
		\end{minipage}%
	\\
 \vspace{.2mm}
\begin{minipage}[t]{\linewidth}
    		\centering
     \begin{tikzpicture}[inner sep=0]
            \node[
            label= {[label distance=-0.20cm,text depth=2.3ex,rotate=90,align=center] left:\textcolor{black}{\footnotesize {(a)}}}]{\includegraphics[width=\cimwid\linewidth,trim={0 0 0 0},clip]{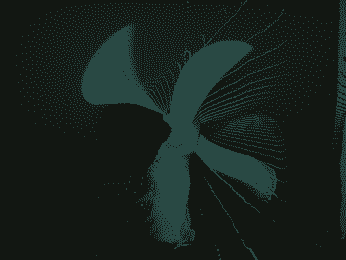}};
            \end{tikzpicture}\hfill
			\includegraphics[width=\cimwid\linewidth,trim={0 0 0 0},clip]{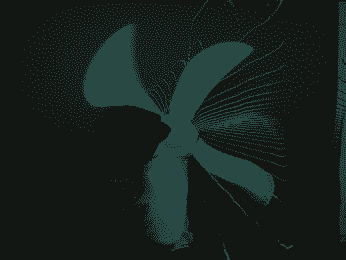}\hfill
			\includegraphics[width=\cimwid\linewidth,trim={0 0 0 0},clip]{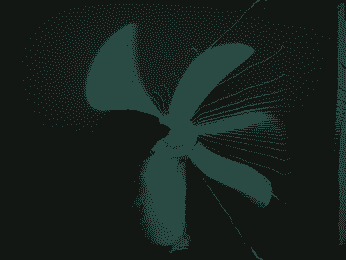}\hfill
			\includegraphics[width=\cimwid\linewidth,trim={0 0 0 0},clip]{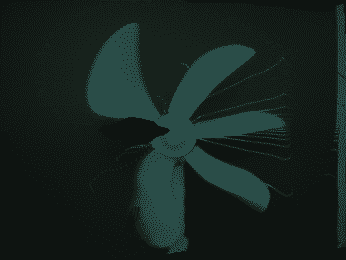}\hfill
			\includegraphics[width=\cimwid\linewidth,trim={0 0 0 0},clip]{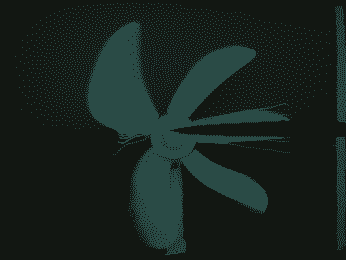}\hfill
			\includegraphics[width=\cimwid\linewidth,trim={0 0 0 0},clip]{Figs/img/REAL_sequence/RSSR_10.png}\hfill
			\includegraphics[width=\cimwid\linewidth,trim={0 0 0 0},clip]{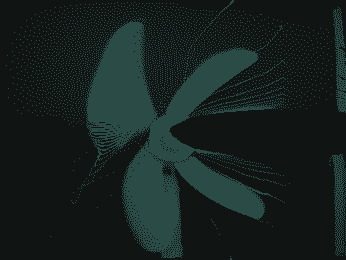}\hfill
    \end{minipage}
 \\
  \vspace{.2mm}
\begin{minipage}[t]{\linewidth}
    		\centering
      \begin{tikzpicture}[inner sep=0]
            \node[
            label= {[label distance=-0.20cm,text depth=2.3ex,rotate=90,align=center] left:\textcolor{black}{\footnotesize {(b)}}}]{\includegraphics[width=\cimwid\linewidth,trim={0 0 0 0},clip]{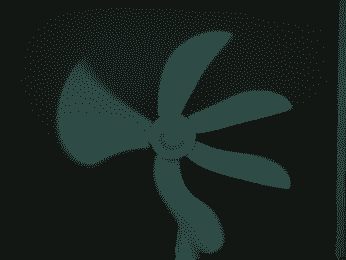}};
            \end{tikzpicture}\hfill
			\includegraphics[width=\cimwid\linewidth,trim={0 0 0 0},clip]{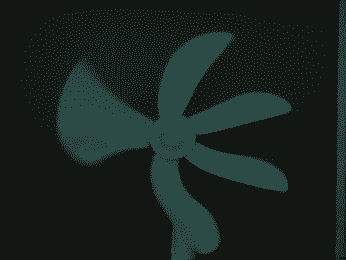}\hfill
			\includegraphics[width=\cimwid\linewidth,trim={0 0 0 0},clip]{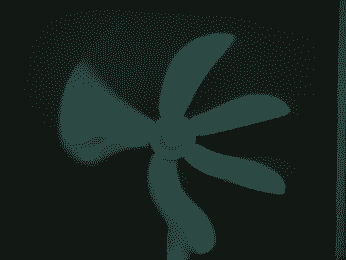}\hfill
			\includegraphics[width=\cimwid\linewidth,trim={0 0 0 0},clip]{Figs/img/REAL_sequence/CVR_6.png}\hfill
			\includegraphics[width=\cimwid\linewidth,trim={0 0 0 0},clip]{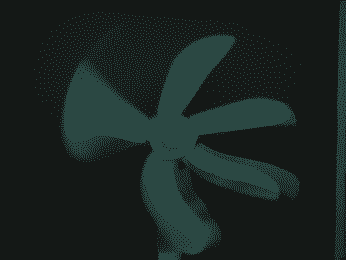}\hfill
			\includegraphics[width=\cimwid\linewidth,trim={0 0 0 0},clip]{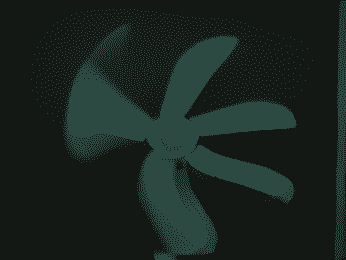}\hfill
			\includegraphics[width=\cimwid\linewidth,trim={0 0 0 0},clip]{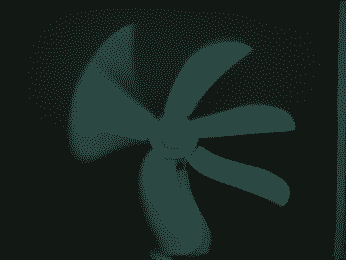}\hfill
    \end{minipage}
    \\
     \vspace{.2mm}
\begin{minipage}[t]{\linewidth}
    		\centering
      \begin{tikzpicture}[inner sep=0]
            \node[
            label= {[label distance=-0.20cm,text depth=2.3ex,rotate=90,align=center] left:\textcolor{black}{\footnotesize {(c)}}}]{\includegraphics[width=\cimwid\linewidth,trim={0 0 0 0},clip]{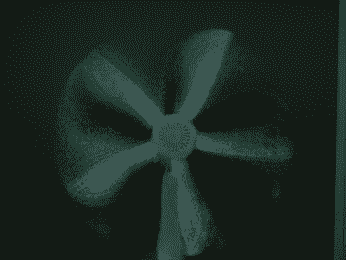}};
            \end{tikzpicture}\hfill
			\includegraphics[width=\cimwid\linewidth,trim={0 0 0 0},clip]{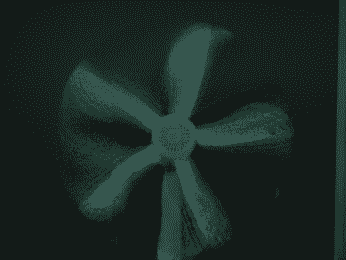}\hfill
			\includegraphics[width=\cimwid\linewidth,trim={0 0 0 0},clip]{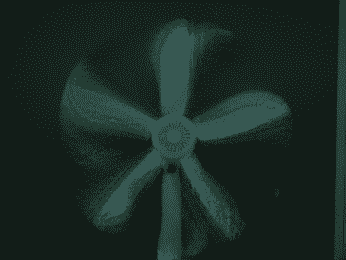}\hfill
			\includegraphics[width=\cimwid\linewidth,trim={0 0 0 0},clip]{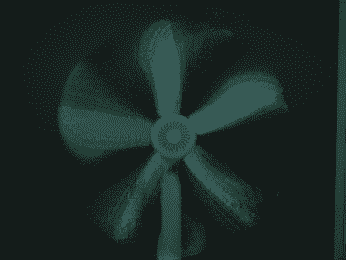}\hfill
			\includegraphics[width=\cimwid\linewidth,trim={0 0 0 0},clip]{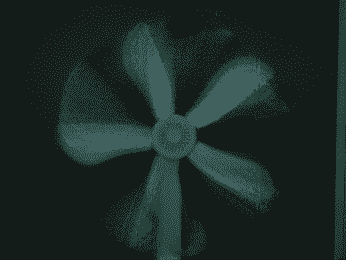}\hfill
			\includegraphics[width=\cimwid\linewidth,trim={0 0 0 0},clip]{Figs/img/REAL_sequence/00031_result_12.png}\hfill
			\includegraphics[width=\cimwid\linewidth,trim={0 0 0 0},clip]{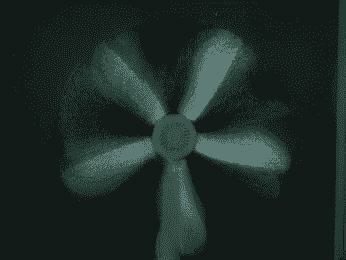}\hfill
    \end{minipage}
     \\
      \vspace{.2mm}
\begin{minipage}[t]{\linewidth}
    		\centering
			\begin{tikzpicture}[inner sep=0]
            \node[
            label= {[label distance=-0.20cm,text depth=2.3ex,rotate=90,align=center] left:\textcolor{black}{\footnotesize {(d)}}}]{\includegraphics[width=\cimwid\linewidth,trim={0 0 0 0},clip]{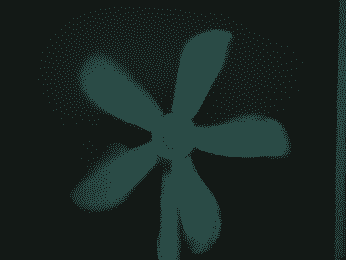}};
            \end{tikzpicture}\hfill
			\includegraphics[width=\cimwid\linewidth,trim={0 0 0 0},clip]{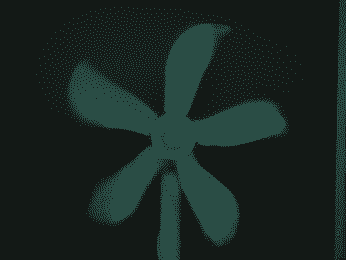}\hfill
			\includegraphics[width=\cimwid\linewidth,trim={0 0 0 0},clip]{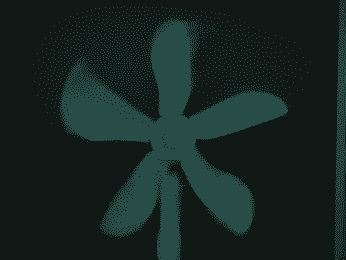}\hfill
			\includegraphics[width=\cimwid\linewidth,trim={0 0 0 0},clip]{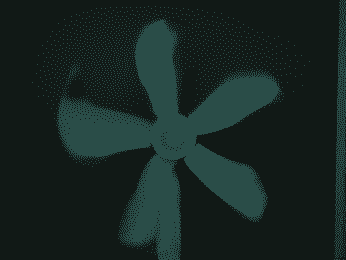}\hfill
			\includegraphics[width=\cimwid\linewidth,trim={0 0 0 0},clip]{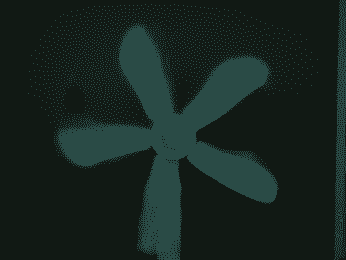}\hfill
			\includegraphics[width=\cimwid\linewidth,trim={0 0 0 0},clip]{Figs/img/REAL_sequence/06_001435_032300_00031-res_75.png}\hfill
            \includegraphics[width=\cimwid\linewidth,trim={0 0 0 0},clip]{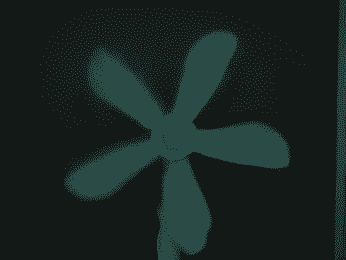}\hfill
    \end{minipage}
    \\
     \vspace{.4mm}
    \begin{minipage}[t]{\linewidth}
    		\centering
            \begin{tikzpicture}[inner sep=0]
            \node[
            label= {[label distance=0.05cm,text depth=-1ex,align=left] below:\textcolor{black}{\footnotesize {t = 1/8}}},
            label= {[label distance=-0.33cm,text depth=2.3ex,rotate=90,align=center] left:\textcolor{black}{\footnotesize {Event}}}]{\includegraphics[width=\ccimwid\linewidth,trim={0 0 0 0},clip,cframe=black .005mm]{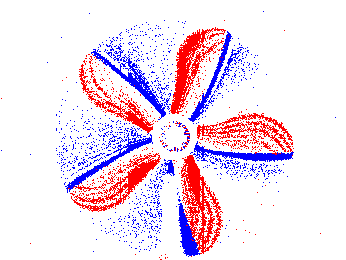}};
            \end{tikzpicture}
			\hfill
            \begin{tikzpicture}[inner sep=0]
            \node [label={[label distance=0.05cm,text depth=-1ex]below: \textcolor{black}{\footnotesize {t = 2/8}}}]  {
			\includegraphics[width=\ccimwid\linewidth,trim={0 0 0 0},clip,cframe=black .005mm]{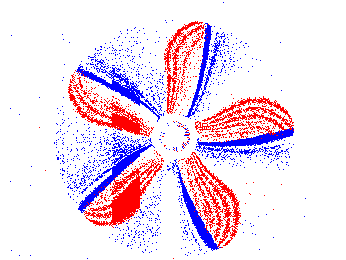}};\end{tikzpicture}
			\hfill
            \begin{tikzpicture}[inner sep=0]
            \node [label={[label distance=0.05cm,text depth=-1ex]below: \textcolor{black}{\footnotesize {t = 3/8}}}]  {
			\includegraphics[width=\ccimwid\linewidth,trim={0 0 0 0},clip,cframe=black .005mm]{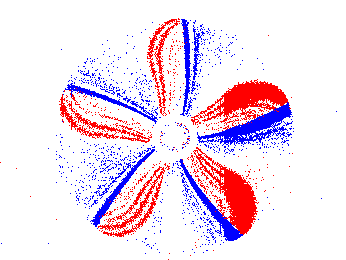}};\end{tikzpicture}
			\hfill
            \begin{tikzpicture}[inner sep=0]
            \node [label={[label distance=0.05cm,text depth=-1ex]below: \textcolor{black}{\footnotesize {t = 4/8}}}]  {
			\includegraphics[width=\ccimwid\linewidth,trim={0 0 0 0},clip,cframe=black .005mm]{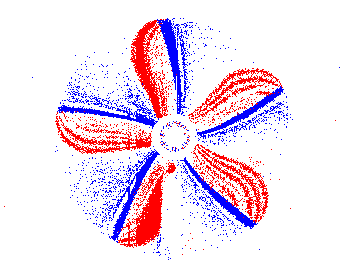}};\end{tikzpicture}
			\hfill
            \begin{tikzpicture}[inner sep=0]
            \node [label={[label distance=0.05cm,text depth=-1ex]below: \textcolor{black}{\footnotesize {t = 5/8}}}]  {
			\includegraphics[width=\ccimwid\linewidth,trim={0 0 0 0},clip,cframe=black .005mm]{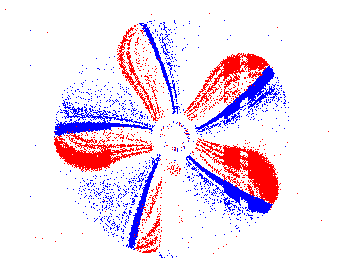}};\end{tikzpicture}
			\hfill
            \begin{tikzpicture}[inner sep=0]
            \node [label={[label distance=0.05cm,text depth=-1ex]below: \textcolor{black}{\footnotesize {t = 6/8}}}]  {
			\includegraphics[width=\ccimwid\linewidth,trim={0 0 0 0},clip,cframe=black .005mm]{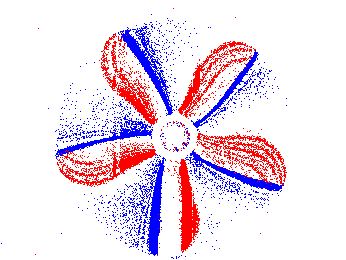}};\end{tikzpicture}
			\hfill
            \begin{tikzpicture}[inner sep=0]
            \node [label={[label distance=0.05cm,text depth=-1ex]below: \textcolor{black}{\footnotesize {t = 7/8}}}]  {
            \includegraphics[width=\ccimwid\linewidth,trim={0 0 0 0},clip,cframe=black .005mm]{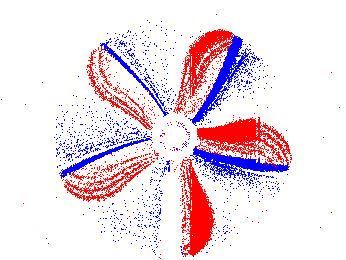}};\end{tikzpicture}
			\hfill
    \end{minipage}
	\caption{A challenging example on real-world Gev-RS-Real dataset \cite{zhou2022evunroll} of unrolling RS images to GS video sequences. {Frame-based methods, RSSR~\cite{fan2021inverting} and CVR~\cite{fan2022context}, face difficulties in handling the typical non-linear motion. Event-based method EvUnroll~\cite{zhou2022evunroll}} suffers from "halo" artifacts and blurry details due to the distribution gap between synthetic training and real testing data. By contrast, our \myname\ achieves visually pleasant results by directly fitting real-world data distribution in a self-supervised manner.
 }
	\label{mul_image_real}
\end{figure}

Recovering GS frame from RS observations by removing the spatial distortion is highly ill-posed, especially when there is no supplementary information available on camera or object movements. \wang{Current methods tackle this issue by relying on artificial assumptions about scenes~\cite{lao2018robust,purkait2018minimal,rengarajan2016bows} and motions~\cite{zhuang2019learning,liu2020deep,rengarajan2017unrolling} and thus fall short of accurate GS frame reconstruction. Moreover, these methods can only generate GS frames at a specific time due to the lack of {\it temporal dynamic information}.
While} the SDR task is further complicated by the need to recover undistorted GS frames for arbitrary timestamps, necessitating the recovery of lost intra-/inter-frame information in each moment. Recent methodologies tackle this requirement by leveraging motion linearity on prior assumptions~\cite{fan2021inverting,fan2022context} or data-specific characteristics learned from specific datasets~\cite{zhou2022evunroll}.
However, these artificial assumptions and data-specific characteristics are not always held in real-world scenarios, leading to significant performance degradation as illustrated in~\cref{mul_image_real}.
 %
 
 %
%
%
%
%
%

%

%
%
%

%
%
%
%

%

%
%

%

%


%
%

%


 %

%

%
%
%
%
%
%
%

%


Therefore, \lei{the task of SDR by inverting RS images to continuous-time GS videos} in real-world scenarios still remains an ongoing and challenging research area. In this paper, we propose to leverage inter-/intra-frame information with the aid of event cameras, a neuromorphic sensor that asynchronously emits events in response to the brightness change at an extremely high temporal resolution~\cite{brandli2014240,gallego2020event}.
While preliminary results have shown the feasibility of using events \wang{at the intra-frame time} to correct RS distortions through EvUnroll~\cite{zhou2022evunroll}, \wang{the failure to utilize inter-frame information and the use of} the linear motion assumption in the RS2GS flow module still fall short of achieving accurate SDR. Furthermore, the model is trained solely on synthetic RS datasets, consisting of manually synthesized events and RS images, which limits its performance in real-world scenarios due to the ``synthetic-to-real'' gap caused by discrepancies in {\it data distribution} and {\it modality correspondence}. On the one hand, the distribution gap between the synthetic and real datasets exists in either events or RS images due to the imperfection of physical cameras, \eg, intrinsic thermal noise and event threshold variations~\cite{brandli2014240}. In addition, the emission rate of events is confined to the read-out bandwidth~\cite{inivaiton2020}, resulting in temporal disorders of events that are difficult to simulate using event simulators~\cite{rebecq2018esim,delbruck2020v2e}. On the other hand, the per-pixel correspondence between the events and images is relatively simple and can be established in network training~\cite{zhangdata,pan2019bringing,zhang2022unifying}, but the learned correspondence in pre-trained models is fragile and can be disrupted by variations in the response functions of RS cameras in real-world scenes~\cite{zhang2022unifying}.

%
%

%
%
%
%

%

%
    


%

Therefore, training with real events and RS images is essential to ensure optimal SDR performance in real-world scenarios.
However, collecting RS images with high frame-rate GS references for training purposes is sophisticated and requires the use of an additional expensive high-speed camera, such as Phantom VEO 640 (\$ 67,500) used in Gev-RS~\cite{zhou2022evunroll}. And this highlights the need for cost-effective alternatives: {\it Can we learn SDR from the real-world RS images without the ground-truth GS references?} 

%



%
%
 %
%
The answer is {\bf YES}. We propose a novel self-supervised framework that can achieve accurate SDR from distorted RS observations without the need for ground-truth GS references. The proposed framework can perform RS to GS (RS2GS) conversion of arbitrary timestamps using real events and RS images, which is a significant advancement over existing methods. {\it To the best of our knowledge, this is the first attempt to apply self-supervised learning to RS2GS conversion.}
Specifically, we first develop an Event-based Inter/intra-frame Compensator (E-IC) to establish a unified spatial and temporal connection between RS and GS image domains via flow-based and synthesis-based techniques. 
Our E-IC takes flexible segments of events as inputs and thus can restore latent GS images at arbitrary timestamps without re-training, benefiting GS video extraction from a single RS frame.
By exploring the cross-domain constraints from spatial and temporal perspectives, we train the E-IC network with real events and RS frames in a fully self-supervised manner and we call it {\bf \myname-S}.
%
%
%
Taking into account the motion consistency and occlusion disturbance, we further propose a Motion and Occlusion Aware (MOA) module and extend {\bf SelfUnroll-S} to {\bf SelfUnroll-M}, which produces smoother GS videos even in the presence of occlusions. Finally, we construct a new dataset composed of real events and RS frames to facilitate the research of RS2GS conversion in real-world scenarios.

\par
Overall, our contributions are mainly three-fold:
\begin{itemize}
    \item We unify the spatial and temporal connection between RS and GS image domains through E-IC, which is able to restore high frame-rate GS videos from a single RS image and the concurrent events.

    \item We propose a self-supervised learning framework for 
    event-based RS2GS conversion, which enables model training with real-world events and RS images.

    \item We build a new real-world dataset containing both RS frames and event streams for a variety of indoor and outdoor scenes, benefiting event-based RS correction research. The dataset and code are available at \url{\website}.

\end{itemize}

Our proposed \myname\ outperforms state-of-the-art RS2GS approaches on publicly available benchmarks. Experiments on real-world datasets also reveal the effectiveness and superiority of our self-supervised approach.
\section{Related Work}\label{Sec:RelatedWork}
\noindent\textbf{Rolling Shutter Correction.}
    Existing frame-based rolling shutter correction methods can be roughly categorized into two classes: {\it model-driven approaches} and {\it learning-based approaches}.
%
For model-driven methods, Baker \etal\cite{baker2010removing} present a constant affine or translational distortion model to estimate the per-pixel motion vector from consecutive RS frames and correct the rolling shutter distortion. Rengarajan \etal\cite{rengarajan2016bows} propose the rule ``straight line must remain straight'' to estimate camera motion by extracting curves. Purkait \etal\cite{Purkait_2017_ICCV} leverage geometric properties of the 3D scene to correct the distortion by estimating the orthogonal vanishing direction. Zhuang \etal\cite{zhuang2017rolling} propose an RS-aware differential SfM algorithm, where the camera motion and dense depth map are utilized in an RS-aware warping for image rectification. Albl \etal\cite{albl2020two} use a novel and effective dual-scanning (bottom-to-top and top-to-bottom) RS camera setup for RS correction. RS camera motion estimation problem\cite{grundmann2012calibration,liu2013bundled,lao2018robust,purkait2018minimal} can also be addressed with the aid of RANSAC\cite{fischler1981random}. 
\par 
Recently, learning-based approaches are developed to achieve better RS2GS conversion performance.
Rengarajan \etal\cite{rengarajan2017unrolling} first propose a CNN model to estimate camera motion parameters from a single RS image. 
Zhuang \etal\cite{zhuang2019learning} extend their previous work \cite{zhuang2017rolling} and develop depth- and motion-aware models to predict dense depth maps and camera motions from a single RS image. 
Liu \etal\cite{liu2020deep} design a differentiable forward warping module that enables learning RS correction in an end-to-end manner. Fan \etal\cite{fan2021sunet} leverage symmetric consistency constraint to aggregate the contextual cues. 
Zhong \etal\cite{zhong2021towards} build the Joint Correction and Deblurring (JCD) network using a deformable attention module to simultaneously achieve RS correction and motion deblurring. \lin{However, most RS correction methods are designed to restore one GS image at a specific moment, and thus fail to extract and leverage the latent scene dynamic.}
%

%
\par
\noindent\textbf{Video Frame Interpolation.}
Existing Video Frame Interpolation (VFI) approaches\cite{bao2019depth,bao2019memc,liu2017video} can be categorized into flow-based and kernel-based approaches. Flow-based approaches generally predict intermediate images by bidirectional optical flow \cite{jiang2018super,sun2018pwc}. Nevertheless, most of them assume uniform motion and linear optical flow between consecutive frames \cite{reda2019unsupervised,bao2019memc}, which may be violated in real scenes with complex and nonlinear motions. Kernel-based methods usually model the frame interpolation as the local convolution with reference frames \cite{niklaus2017video,niklaus2017videoadapt}, which is more robust to brightness changes, but the scalability of kernel-based approaches is often limited by the fixed sizes of convolution kernels. 
Although VFI approaches are able to extract scene dynamics, existing methods generally assume GS references and cannot be directly applied to RS inputs.

\par

\begin{figure*}[!htb]
  \centering
  \begin{tabular}{c c c}
  \includegraphics[width=0.32\textwidth]{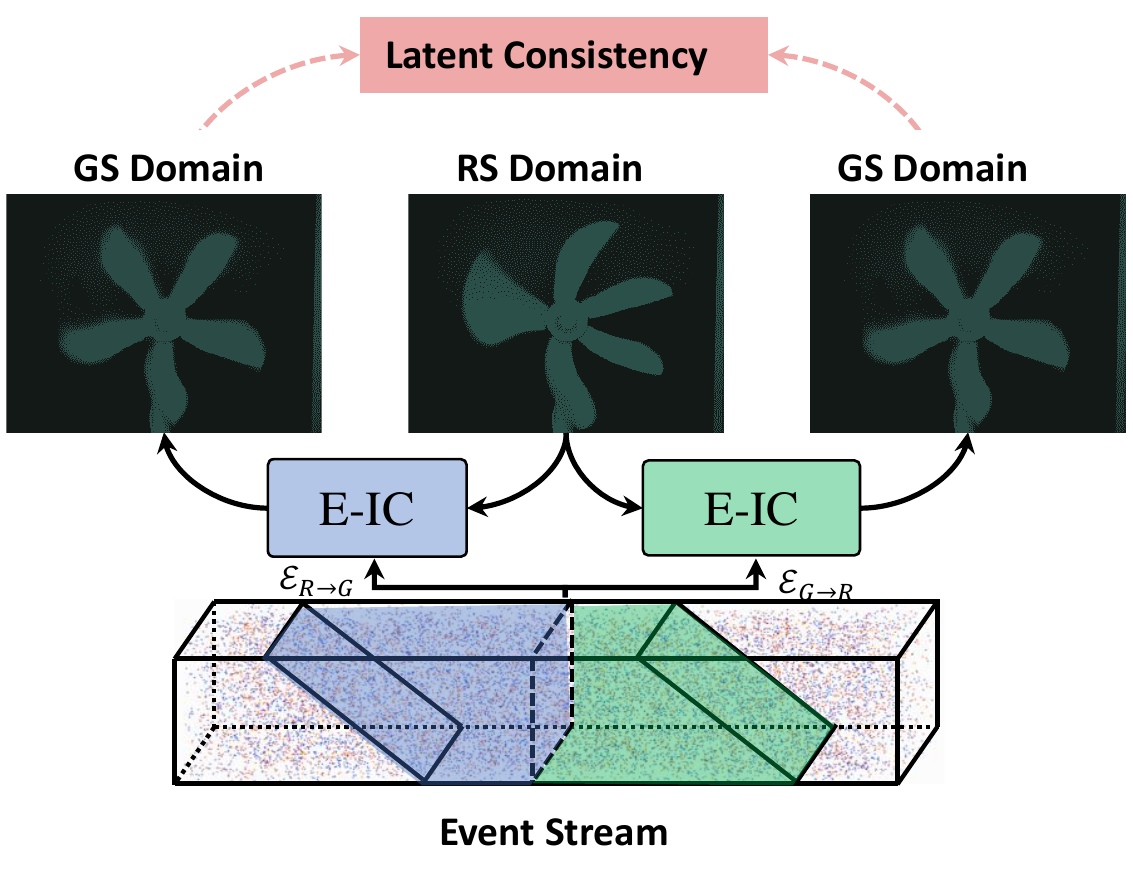}&
  \includegraphics[width=0.32\textwidth]{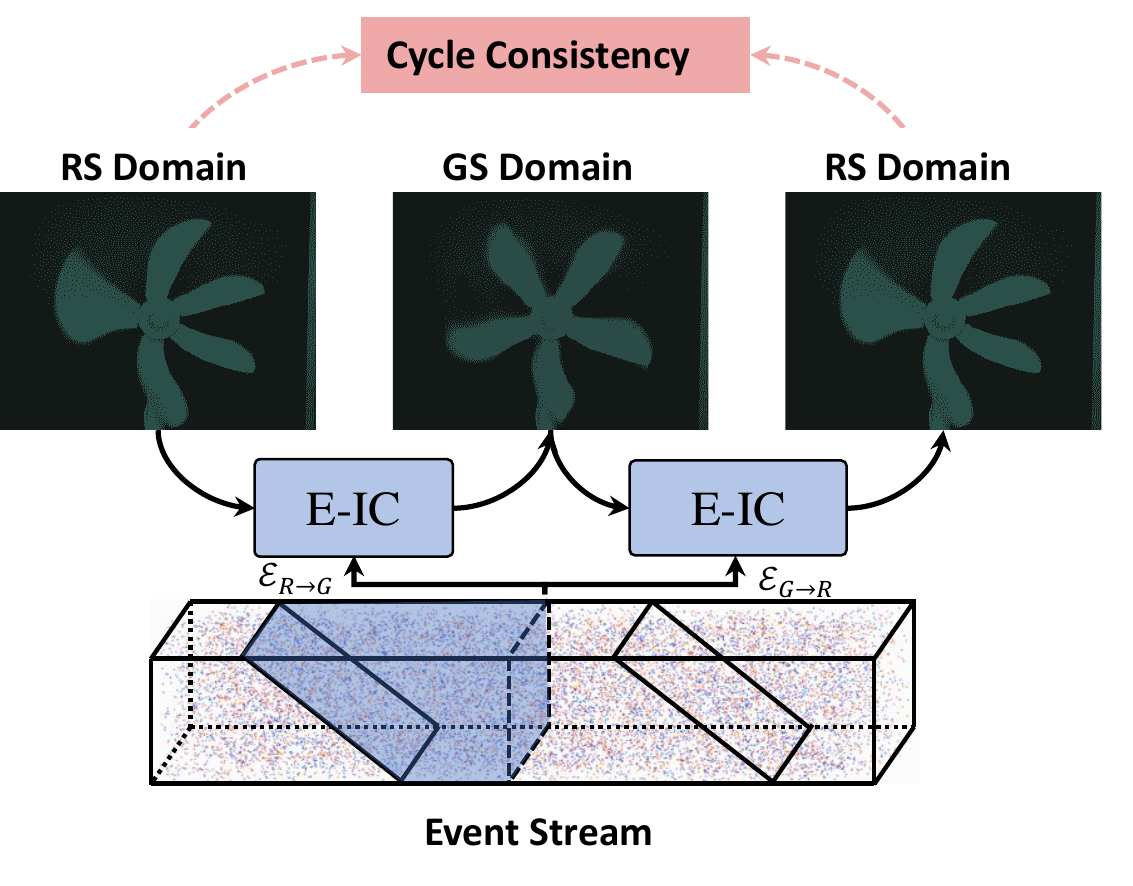} & \includegraphics[width=0.32\textwidth]{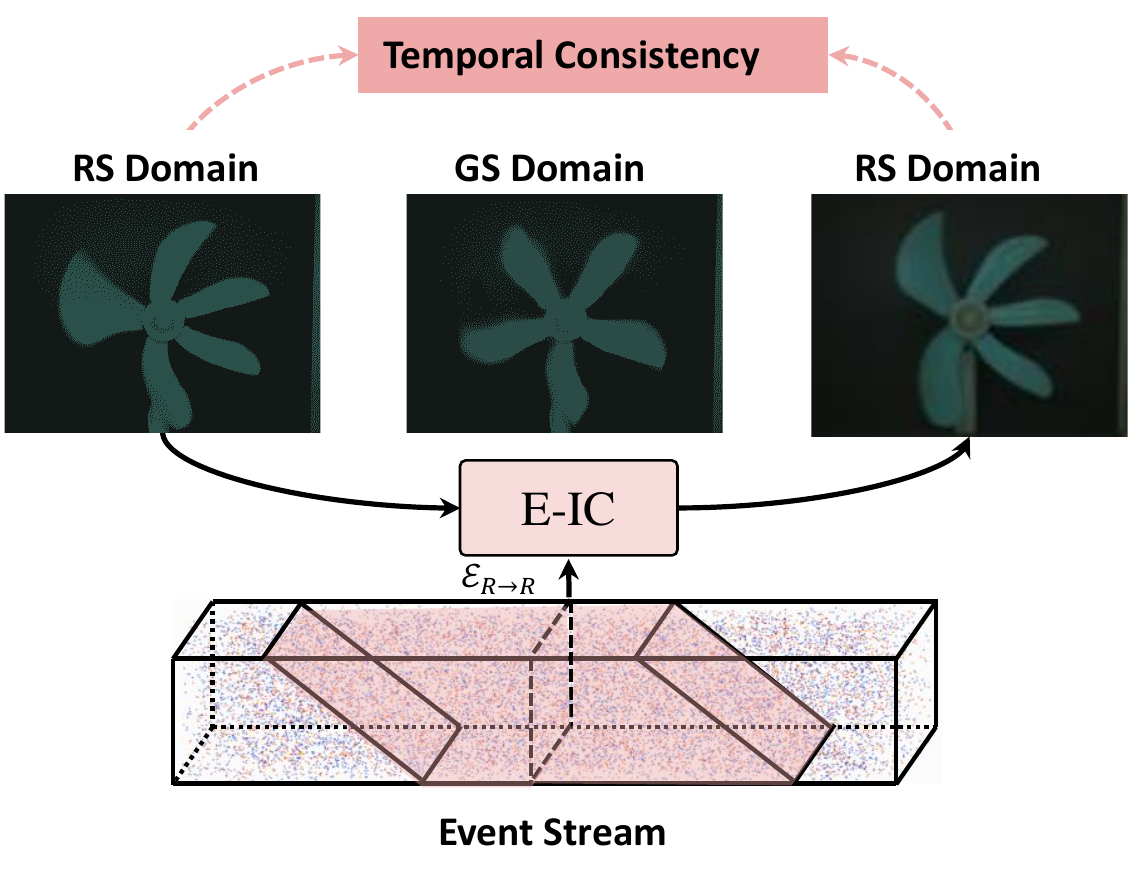}  \\ 
  \small{(a) Latent Consistency.} &\small{(b) Cycle Consistency.} & \small{(c) Temporal Consistency.}  
  \end{tabular}
  \caption{Illustration of the proposed self-supervised learning framework and the domain transformation by E-IC between RS and GS parameterized with event segmentation, \ie, $\mathcal{E}_{R\rightarrow G}$, $\mathcal{E}_{G\rightarrow R}$, and $\mathcal{E}_{R\rightarrow R}$.}
   \label{rsgsev}
\end{figure*}
\par
\noindent\textbf{Event-based RS Correction and VFI.} 
Event cameras are neuromorphic sensors that report asynchronous event streams in response to brightness changes~\cite{lichtsteiner128Times1282008}, which poses a paradigm shift in visual perception and enables almost continuous observation of dynamic scenes.
Thanks to the extremely low latency, events implicitly encode inter/intra-frame information in terms of motions and textures~\cite{gallego2020event,wang2020event,lin2020learning,xu2021motion}, benefiting both VFI~\cite{tulyakov2021time,tulyakov2022time,he2022timereplayer,zhang2022unifying} and RS correction~\cite{zhou2022evunroll}.

For VFI tasks, Timelens \cite{tulyakov2021time} and Timelens++\cite{tulyakov2022time} are pioneer works that marry the advantages of warping-based and synthesis-based interpolation approaches, which have the ability to handle illumination changes and the sudden appearance of new objects between reference frames.  Inspired by \cite{zhu2017unpaired}, He \etal\cite{he2022timereplayer} design an unsupervised learning framework for video interpolation with event streams by utilizing cycle consistency. Zhang \etal\cite{zhang2022unifying} jointly solve the deblurring and interpolation problem by a Learnable Double Integral (LDI) network, which is able to generate high frame-rate sharp videos from consecutive blurry inputs. For RS correction, EvUnroll \cite{zhou2022evunroll} is the first attempt to recover GS frames during the intra-frame time, \wang{which partially achieves scene dynamic recovery}. However, the aforementioned event-based approaches cannot fully achieve scene dynamic recovery from RS images.
\par
\noindent\textbf{Scene Dynamic Recovery from Rolling Shutter Images.} \label{subSec:Joint RS Correction and Video Reconstruction.}
{Scene Dynamic Recovery (SDR) from rolling shutter images which is capable of generating
the undistorted GS frames at any given
timestamp can be seen as a combination of RS correction and VFI, and few works have considered the challenging situation.} Fan \etal\cite{fan2021inverting} build the first network RSSR to extract a latent GS video sequence from two consecutive RS images. Under the assumption of a constant velocity, they convert the predicted optical flow to RS undistortion flow corresponding to scanlines. Furthermore, a context-aware architecture CVR\cite{fan2022context} is proposed by leveraging a contextual aggregation procedure to alleviate the holes and artifacts caused by occlusions. Zhong \etal\cite{albl2020two} propose IFED\cite{zhong2022bringing} that can merge the symmetric information of dual reversed rolling shutter distortion images, and reconstruct GS video sequences. However, due to the limited inter-/intra-frame information, the aforementioned methods often struggle to restore accurate GS results and extract scene dynamics in real-world scenarios with complex motions. 
Overall, the existing SDR methods encounter challenges in achieving accurate GS results, particularly when dealing with complex non-linear motion. Besides most methods are trained over synthetic datasets and often suffer from performance degradation in real-world scenarios due to the ``synthetic-to-real'' gap. To address these issues, we propose the E-IC to recover high frame-rate GS videos from RS frames and events, and develop a fully self-supervised method to directly fit real-world data distribution.

\section{Method}\label{Sec:Methods}
\begin{figure*}[t]
    \centering
    \includegraphics[width=1\linewidth]{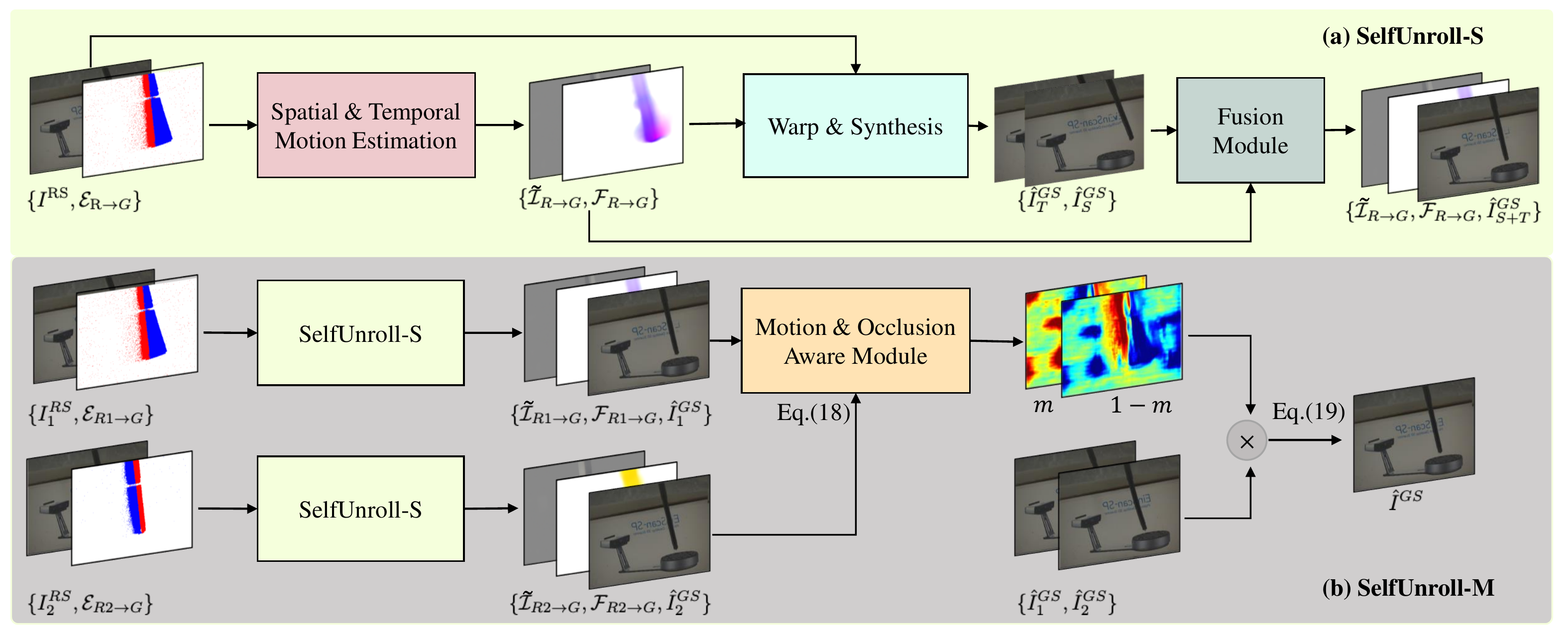}
    \caption{ The overall pipelines of SelfUnroll-S and SelfUnroll-M. \textbf{(a) SelfUnroll-S.} The RS2GS flow $F_{R\rightarrow{G}}$ and residual $\hat{\mathcal{I}}_{R\rightarrow{G}}$ are estimated from the input pair $\{I^{RS},\mathcal{E}_{R \rightarrow {G}}\}$, and the connection between RS domain and GS domain is established by the spatial and temporal transition. A fusion module is applied to take advantage of spatial and temporal compensation. \textbf{(b) SelfUnroll-M.} The input pairs $\{I^{RS}_{1},\mathcal{E}_{R1 \rightarrow {G}}\}$ and $\{I^{RS}_{2},\mathcal{E}_{R2 \rightarrow {G}}\}$ are fed to the SelfUnroll-S to generate GS frame \ie $\hat{I}^{GS}_{1},\hat{I}^{GS}_{2}$, and a motion and occlusion-aware module output the attention map to generate the final reconstruction $\hat{I}^{GS}$.}
    \label{fig:net-arch}
\end{figure*}

We first formulate our task in \cref{sec:problemform} and then design the Event-based Inter/intra-frame Compensator (E-IC) in \cref{sec:eic} based on a unified spatio-temporal connection between RS and GS images. In \cref{sec:SelfUnroll-S}, we introduce SelfUnroll-S built upon E-IC and present a self-supervised learning framework by utilizing the consistencies between RS and GS domains. We further develop a Motion and Occlusion Aware (MOA) module to handle occlusions by exploiting temporal information which extends SelfUnroll-S to SelfUnroll-M in \cref{sec:SelfUnroll-M}. Finally, we present the network details and event representation in \cref{sec:net_detail_event_repre}.


\par 

\subsection{Problem Formulation}\label{sec:problemform}
Generally, we can define a scene dynamic $\mathcal{I}: \mathbb{R}\times \mathbb{Z}^2\to \mathbb{R}_+$, and $\mathcal{I}(t,\mathbf{x})$ determines a continuous video parameterized with time $t$ and pixel location $\mathbf{x}\triangleq (x,y)$. Then a GS image $I^{GS}$ is exposed globally for all pixels at $t_s$,
\begin{equation}
    I^{GS}(\mathbf{x}) \triangleq \mathcal{I}(t_s,\mathbf{x}),
\end{equation}
while an RS image $I^{RS}$ is exposed line-by-line,
\begin{equation}\label{eq:gs2rs}
    I^{RS}(\mathbf{x}) \triangleq \mathcal{I}(\mathcal{T}^{RS}(\mathbf{x}),\mathbf{x}),
\end{equation}
where $\mathcal{T}(\mathbf{x})$ is the location-aware time shifting operator and the superscription of $RS$ denotes the line-by-line time shifting for rolling shutter cameras, \ie,  $\mathcal{T}^{RS}(\mathbf{x}) = {t}_{0}+y \frac{|T|}{H}$ with $t_0$ the start time of exposure, $H$ the image height and $T$ the exposure time interval. 

\wang{Most existing RS2GS works\cite{zhong2021towards,liu2020deep} aims to restore a GS image $I^{GS}$ at a specific time, such as $t_s = {t}_{0}+\frac{T}{2}$, }
\begin{equation}\label{rs2gs}
\mathcal{I}(t_s,\mathbf{x}) \triangleq I^{GS}(\mathbf{x}) = \operatorname{RS2GS} \left(I^{RS}(\mathbf{x})\right).
\end{equation}
\wang{Instead of only restoring the GS image of a single timestamp, this work is dedicated to SDR, which can restore the latent scene dynamic $\mathcal{I}$ of any given timestamp $t\in\mathbb{R}$, }
\begin{equation}\label{sdr}
    \mathcal{I}(t,\mathbf{x}) \triangleq \mathcal{H}\left(I^{RS}(\mathbf{x}),t \right) = \mathcal{H}\left(\mathcal{I}(\mathcal{T}^{RS}(\mathbf{x}),\mathbf{x}),{t} \right)
\end{equation}
which is actually an inversion of \cref{eq:gs2rs} but severely ill-posed because of the missing inter/intra-frame information. 

\lei{To address this problem, artificial assumptions on motion linearity\cite{fan2021inverting,fan2022context} and data-specific characteristics\cite{zhou2022evunroll} are commonly utilized but often suffer from sub-optimal solutions in real-world scenarios.}
\wang{To improve the SDR, this paper proposes utilizing events to learn $\mathcal{H}$ without motion assumptions and building a self-supervised learning framework to learn $\mathcal{H}$ from the real-world RS images and events.}

\par 
Event cameras report asynchronous events whenever the brightness exceeds the event threshold $\eta>0$ in the logarithmic domain\cite{brandli2014240}, \ie,
\begin{equation}\label{eq:event_basic}
 \operatorname{log}(\mathcal{I}(t,\mathbf{x})) -  \operatorname{log}(\mathcal{I}(\tau,\mathbf{x})) = p \cdot \eta,  
\end{equation}
where $\operatorname{log}(\mathcal{I}(t,\mathbf{x}))$ and $\operatorname{log}(\mathcal{I}(\tau,\mathbf{x}))$ denote the log-scale pixel brightness of position $\mathbf{x}$ at time $t$ and $\tau$, and $p\in\{+1,-1\}$ is the 
polarity indicating brightness increase ($+1$) and decrease ($-1$). 
With the aid of events, the scene dynamic transition \cref{sdr} can be reformulated as
\begin{equation}\label{sdr_trans}
    \mathcal{I}(\mathcal{T}_d,{\mathbf{x}})  = \mathcal{H} \left(\mathcal{I}(\mathcal{T}_s,{\mathbf{x}}), \mathcal{E}_{[\mathcal{T}_s,\mathcal{T}_d]}  \right),
\end{equation}
where we drop $\mathbf{x}$ in $\mathcal{T}$ for simplicity and $\mathcal{E}_{[\mathcal{T}_s,\mathcal{T}_d]}$ denotes the set of events triggered during $[\mathcal{T}_s,\mathcal{T}_d]$. 

\lei{Note that \cref{sdr_trans} defines a flexible transition operator $\mathcal{H}$ for the scene dynamic $\mathcal{I}$. By defining different location-aware time shifting operators $\mathcal{T}_s(\mathbf{x})$ and $\mathcal{T}_d(\mathbf{x})$, one can achieve arbitrary exposure mode transitions. For instance, $\mathcal{H}$ with $\mathcal{T}_d(\mathbf{x}) \triangleq t$ and $\mathcal{T}_s(\mathbf{x}) \triangleq t_0+y\frac{|T|}{H}$ defines an RS2GS inversion operator, while $\mathcal{H}$ becomes a GS2RS inversion operator if $\mathcal{T}_s(\mathbf{x}) \triangleq t$ and $\mathcal{T}_d(\mathbf{x}) \triangleq t_0+y\frac{|T|}{H}$, where $t_0$ is the starting timestamp of the RS exposure.} 

\begin{figure*}[!t]
	\centering
\begin{minipage}{0.19\linewidth}
\centering
    \includegraphics[width=\linewidth]{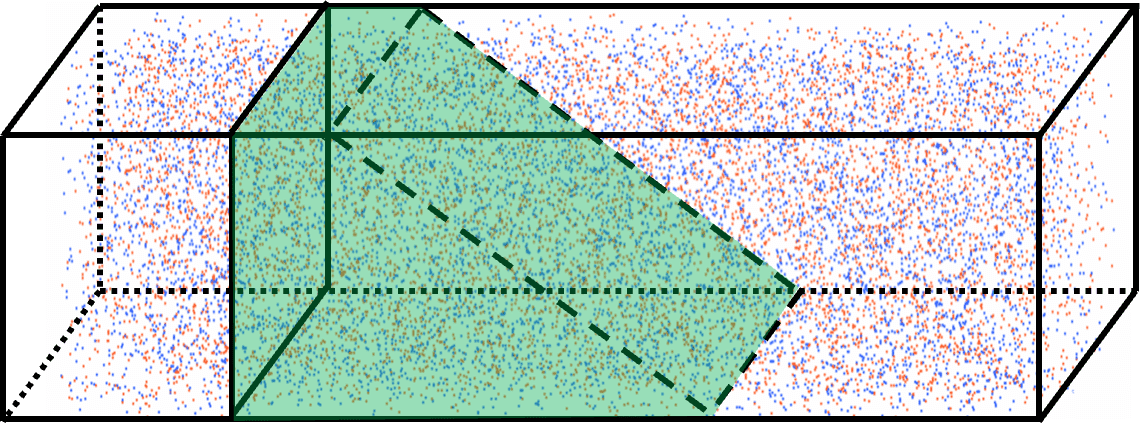}\\
    \footnotesize $\mathcal{E}_{R\to G}= \mathcal{E}_{[\mathcal{T}^{RS}_{\bf x},-0.25T]}$\vspace{.5em}\\
    \includegraphics[width=\linewidth,cframe=black .05mm]{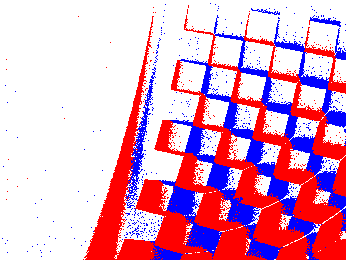} \\
    \includegraphics[width=\linewidth,cframe=black .05mm]{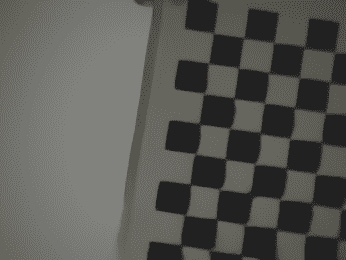}\\
    {\footnotesize  $t=-0.25T$}
\end{minipage}
\begin{minipage}{0.19\linewidth}
\centering
    \includegraphics[width=\linewidth]{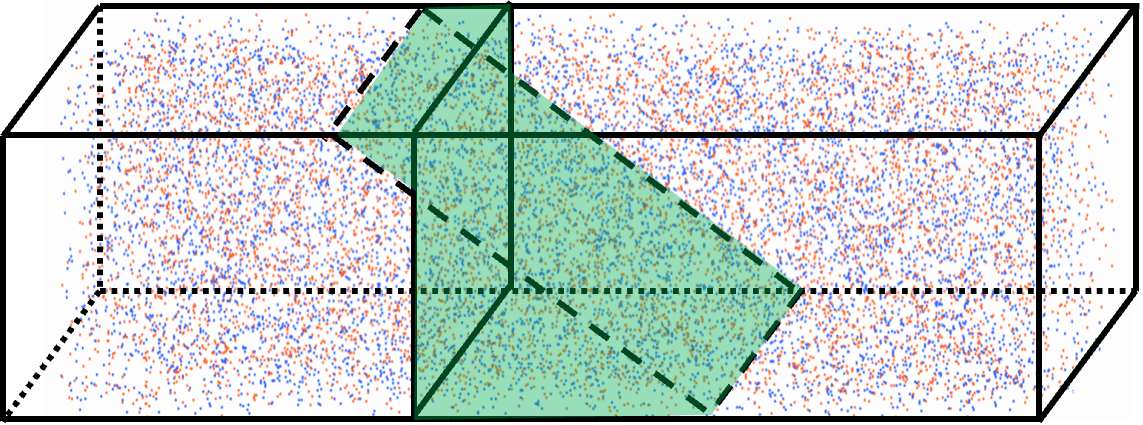}\\
    \footnotesize $\mathcal{E}_{R\to G}= \mathcal{E}_{[\mathcal{T}^{RS}_{\bf x},0.25T]}$\vspace{.5em}\\
    \includegraphics[width=\linewidth,cframe=black .05mm]{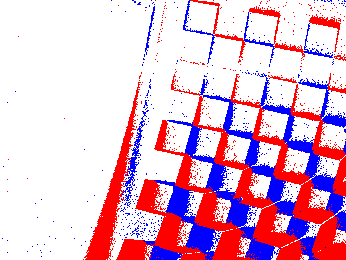}\\
    \includegraphics[width=\linewidth,cframe=black .05mm]{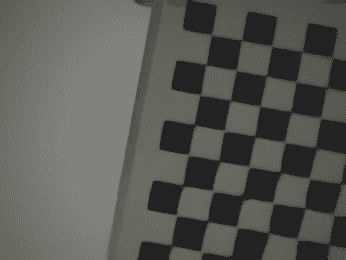}\\
    {\footnotesize $t=0.25T$}
 \end{minipage}
\begin{minipage}{0.19\linewidth}
\centering
    \includegraphics[width=\linewidth ]{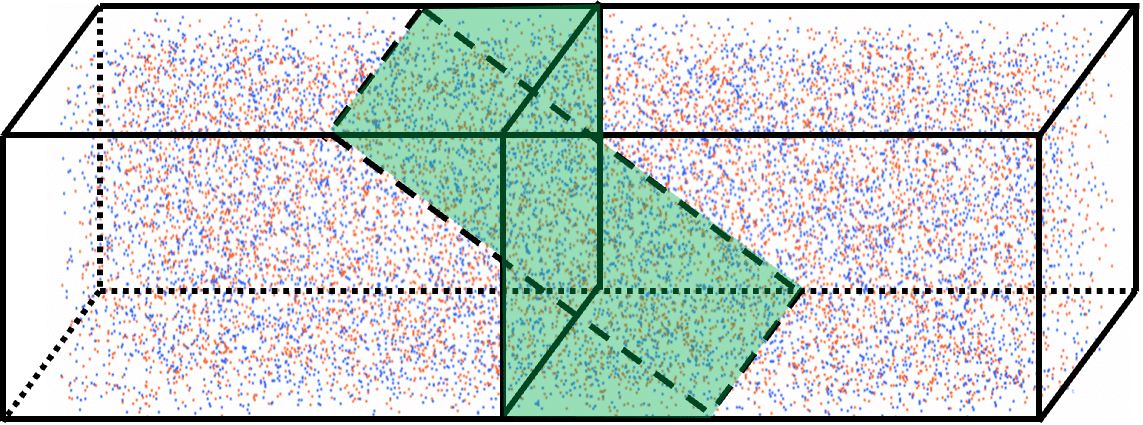}\\
    \footnotesize $\mathcal{E}_{R\to G}= \mathcal{E}_{[\mathcal{T}^{RS}_{\bf x},0.5T]}$\vspace{.5em}\\
    \includegraphics[width=\linewidth,cframe=black .05mm]{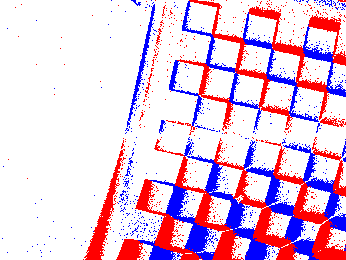}\\
    \includegraphics[width=\linewidth,cframe=black .05mm]{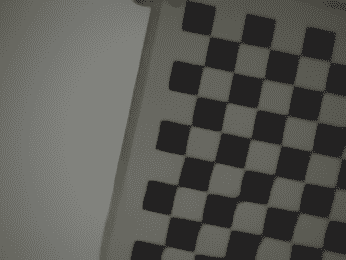}\\
    {\footnotesize $t=0.5T$}
\end{minipage}
\begin{minipage}{0.19\linewidth}
\centering
   \includegraphics[width=\linewidth ]{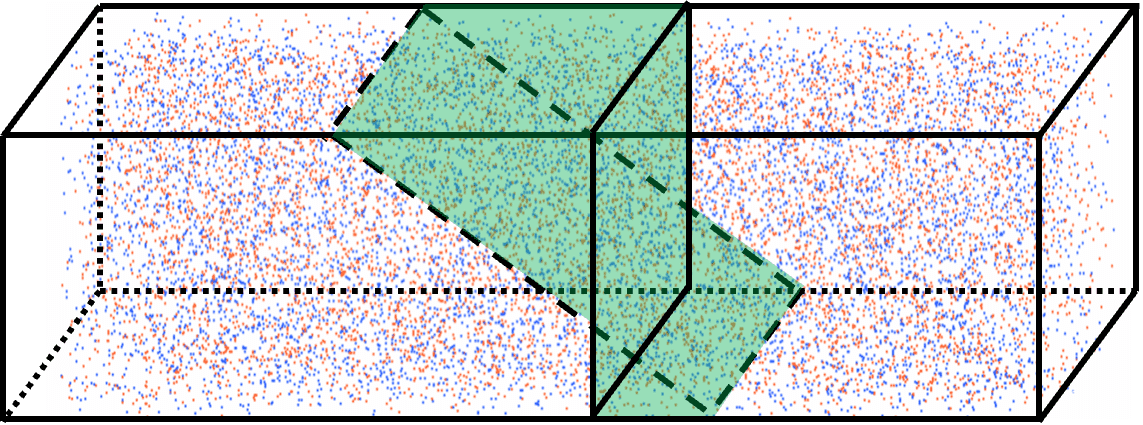}\\
   \footnotesize $\mathcal{E}_{R\to G}= \mathcal{E}_{[\mathcal{T}^{RS}_{\bf x},0.75T]}$\vspace{.5em}\\
   \includegraphics[width=\linewidth,cframe=black .05mm]{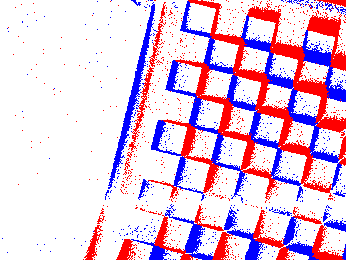}\\
   \includegraphics[width=\linewidth,cframe=black .05mm]{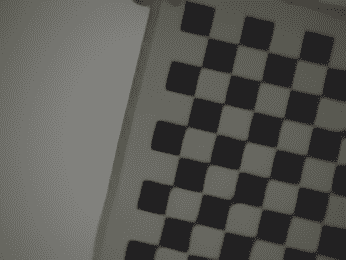}\\
    {\footnotesize $t=0.75T$}
 \end{minipage}
\begin{minipage}{0.19\linewidth}
    \centering
    \includegraphics[width=\linewidth]{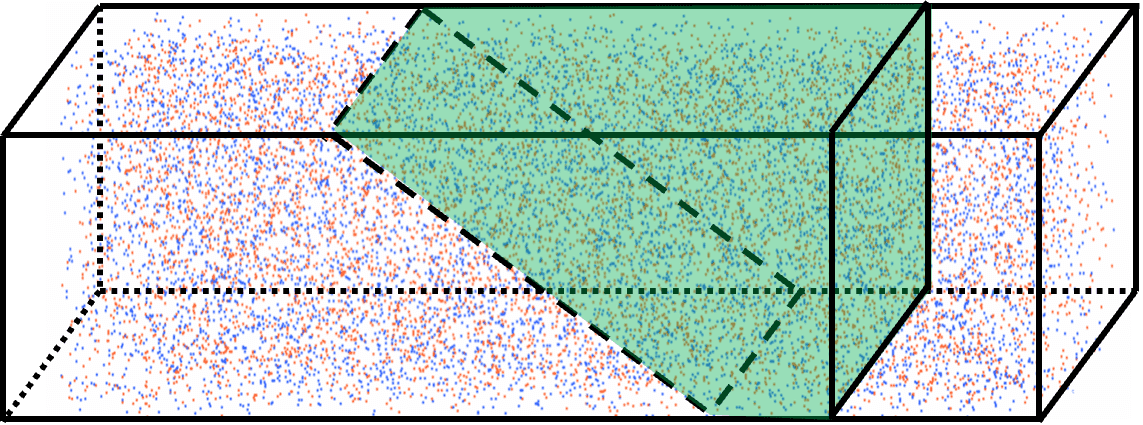}\\
    \footnotesize $\mathcal{E}_{R\to G}= \mathcal{E}_{[\mathcal{T}^{RS}_{\bf x},1.25T]}$\vspace{.5em}\\
    \includegraphics[width=\linewidth,cframe=black .05mm]{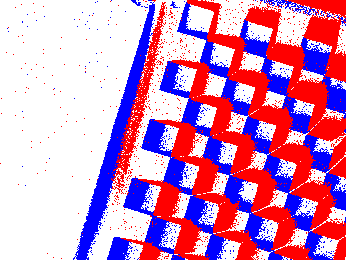}\\
    \includegraphics[width=\linewidth,cframe=black .05mm]{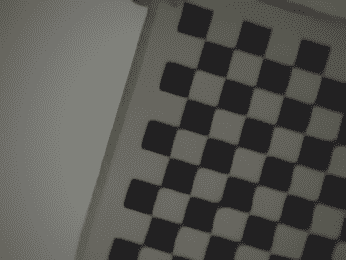}\\
    {\footnotesize $t=1.25T$}
 \end{minipage}

\caption{
Illustration of event segments $\mathcal{E}_{R\to G}$ (1$^{st}$ row) and corresponding event frames (2$^{nd}$ row) for RS2GS inversions (3$^{rd}$ row) of different latent GS timestamps $t\in \{-0.25T,0.25T,0.5T,0.75T,1.25T\}$. Note that $t=-0.25T$ or $1.25T$ represents the reconstructed GS frames not in the RS exposure time interval $[0, T]$.
}
\label{fig:event_seg}
\vspace{1em}
\end{figure*}

For a given transition operator $\mathcal{H}$, we can finally achieve the SDR from RS observations $I^{RS} \triangleq \mathcal{I}(\mathcal{T}_s,{\mathbf{x}})$ with $\mathcal{T}_s(\mathbf{x}) = t_0+y\frac{|T|}{H}$, \ie,
\begin{equation}
    \mathcal{I}(t,\mathbf{x})  =  \mathcal{H} \left(\mathcal{I}(\mathcal{T}_s,{\mathbf{x}}), \mathcal{E}_{[\mathcal{T}_s,t]}  \right)
\end{equation}
where the event segment $\mathcal{E}_{[\mathcal{T}_s,t]}$ is only parameterized by timestamp $t$, illustrated in \cref{fig:event_seg}.
Thus, it is essential to determine the accurate transition operator $\mathcal{H}$ between different exposure modes.

\subsection{Event-based Inter/intra-frame Compensator}\label{sec:eic}
The transition functional $\mathcal{H}$ of scene dynamic $\mathcal{I}$ can be considered in both temporal and spatial dimensions.

\noindent{\bf Temporal Transition.} For a given coordinate $\mathbf{x}$, its temporal intensity transition can be formulated by integrating events over time,
\begin{equation}\label{eq:temporal_trans}
    \mathcal{I}(\mathcal{T}_d,{\mathbf{x}}) =  \mathcal{I}(\mathcal{T}_s,{\mathbf{x}}) \cdot \tilde{\mathcal{I}}\left(\mathcal{E}_{[\mathcal{T}_s,\mathcal{T}_d]},{\mathbf{x}}\right),
\end{equation}
with $\tilde{\mathcal{I}}$ denoting the brightness change between $\mathcal{T}_s,\mathcal{T}_d$, 
\begin{equation}
\tilde{\mathcal{I}}\left(\mathcal{E}_{[\mathcal{T}_s,\mathcal{T}_d]},{\mathbf{x}}\right) \triangleq \operatorname{exp}\left(\eta \int_{\mathcal{T}_s}^{\mathcal{T}_d} e_{\mathbf{x}}(s) ds\right),
\end{equation}
where $e_{\mathbf{x}}(t)\triangleq p \cdot \delta(t-t_0)$ is the continuous event representation of event $(\mathbf{x},t_0,p)$ with $\delta(\cdot)$ denoting a Dirac function. 

\noindent{\bf Spatial Translation.} Besides, benefiting from the low latency of event cameras, events naturally encode the motion information of dynamic scenes. Thus, given the source image, it is also feasible to achieve pixel translation in the spatial domain by utilizing the motion embedded in events, \ie,
\begin{equation}\label{eq:spatial_trans}
    \mathcal{I}(\mathcal{T}_d,\mathbf{x}) = \mathcal{I}(\mathcal{T}_s,\mathbf{x} + \mathcal{F}(\mathcal{E}_{[\mathcal{T}_s, \mathcal{T}_d]})),
\end{equation}
where $\mathcal{F}(\mathcal{E}_{[\mathcal{T}_s, \mathcal{T}_d]})$ denotes the optical flow estimated from events $\mathcal{E}_{[\mathcal{T}_s, \mathcal{T}_d]}$ triggered in the interval $[\mathcal{T}_s,\mathcal{T}_d]$. 

\par 
However, directly computing the target image by Eq.~\eqref{eq:temporal_trans} or \eqref{eq:spatial_trans} often suffers from the instability of the event threshold and the disturbance from noise events, and thus we propose to employ deep learning-based methods for better pixel transition performance. 
In our approach, multiple Event-based Inter/intra-frame Compensators (E-ICs) are designed to implement the function $\mathcal{H}$ and take advantage of spatial and temporal compensations, \ie,
\begin{equation}\label{eq:eic}
   \operatorname{E-IC}_{{S+T}} = \operatorname{Fusion}(\operatorname{E-IC}_{{S}}, \operatorname{E-IC}_{{T}}),
\end{equation}
where $\operatorname{E-IC}_{{T}}$ and $\operatorname{E-IC}_{{S}}$ are networks that respectively approximate the functions of Eqs.~\eqref{eq:temporal_trans} and \eqref{eq:spatial_trans}, and $\operatorname{E-IC}_{{S+T}}$ fuses the results output from $\operatorname{E-IC}_{{S}}$ and $\operatorname{E-IC}_{{T}}$. Define $\operatorname{E-IC}_{*}$ as any one of $\operatorname{E-IC}_{{S}}$, $\operatorname{E-IC}_{{T}}$, and $\operatorname{E-IC}_{{S+T}}$, we can achieve flexible RS-GS conversion by feeding different events and images into our E-IC, as depicted in {\cref{rsgsev}}. 
Specifically, we define the RS2GS, GS2RS, and RS2RS conversion as 
\begin{equation}\label{eq:e-ic}
\begin{aligned}
&\textbf{RS2GS:}\quad    \hat{I}^{GS}= \operatorname{E-IC}_{*}(I^{RS}, \mathcal{E}_{R\rightarrow G}),
\\
&\textbf{GS2RS:}\quad    \hat{I}^{RS} = \operatorname{E-IC}_{*}(I^{GS}, \mathcal{E}_{G\rightarrow R}),
\\
&\textbf{RS2RS:}\quad    \hat{I}^{RS} = \operatorname{E-IC}_{*}(I^{RS}, \mathcal{E}_{R\rightarrow R}),
\end{aligned}
\end{equation}
where $I$ and $\hat{I}$ denote the observed and restored $\mathcal{I}$ in the form of either RS or GS images, $\mathcal{E}_{R\rightarrow G}$, $\mathcal{E}_{G\rightarrow R}$, and $\mathcal{E}_{R\rightarrow R}$ indicate the events segmented according to the timestamps of input and output images, as illustrated in {\cref{rsgsev}}.

%
%
%
%
%
%
%
%
%
%
%

%

%
\subsection{SelfUnroll with a Single RS Frame}\label{sec:SelfUnroll-S}

The E-IC described in \cref{sec:eic} establishes a unified transformation between images in RS and GS domains based on events.
We implement E-IC for spatial pixel translation with a flow-based network \cite{tulyakov2021time} and for temporal intensity transition with a synthesis-based network \cite{xu2021motion}, and fuse two E-ICs to achieve RS correction from both spatial and temporal perspectives. We refer to this network as SelfUnroll-S (SelfUnroll with single RS frame) as shown in {\cref{fig:net-arch}}, and propose a self-supervised learning framework for training.
\subsubsection{Self-supervised Learning Framework} \label{subSec:Self-supervised Learning Framework}

Based on E-ICs that achieve flexible image transition between RS and GS domains, we devise the following three constraints and form a self-supervised learning framework.

\noindent\textbf{Latent Consistency.} 
Given two consecutive RS images $I^{RS}_1$ and $I^{RS}_2$ captured with exposure time $\mathcal{T}^{RS}_1,\mathcal{T}^{RS}_2$, we define the target GS image with exposure time $\mathcal{T}$. 
Then with event segments $\mathcal{E}_{[\mathcal{T}^{RS}_1,\mathcal{T}]}, \mathcal{E}_{[\mathcal{T}^{RS}_2,\mathcal{T}]}$, one can invert two RS images to the same latent GS image, leading to the latent consistency $\mathcal{L}_{lc}$ in the GS domain (\cref{rsgsev} \textcolor{red}{(a)}), \ie, 
\begin{equation}\label{eq:lcloss}
        \mathcal{L}_{lc} = \left\| \operatorname{E-IC}_{*}(I^{RS}_1, \mathcal{E}_{[\mathcal{T}^{RS}_1,\mathcal{T}]}) - \operatorname{E-IC}_{*}(I^{RS}_2, \mathcal{E}_{[\mathcal{T}^{RS}_2,\mathcal{T}]}) \right\|_1 .
\end{equation}

\noindent\textbf{Cycle Consistency.} We can formulate a cycle consistency $\mathcal{L}_{cc}$ in the RS domain by conducting the RS2GS process followed by a GS2RS one (\cref{rsgsev} \textcolor{red}{(b)}), \ie, 
\begin{equation}\label{eq:ccloss}
    \mathcal{L}_{cc}=\left\| \operatorname{E-IC}_{*}\left(\operatorname{E-IC}_{*}(I^{RS}, \mathcal{E}_{[\mathcal{T}^{RS},\mathcal{T}]}), \mathcal{E}_{[\mathcal{T},\mathcal{T}^{RS}]})\right) -  I^{RS} \right\|_1
\end{equation}

\noindent\textbf{Temporal Consistency.} Using the events between two consecutive RS frames can establish a temporal consistency in the RS domain (\cref{rsgsev} \textcolor{red}{(c)}), \ie, 
\begin{equation}\label{eq:tcloss}
    \begin{aligned}
        \mathcal{L}_{tc}=&\| \operatorname{E-IC}_{*}(I^{RS}_1, \mathcal{E}_{[\mathcal{T}^{RS}_1,\mathcal{T}^{RS}_2]}) - I^{RS}_2 \|_1
        \\
        +&\| \operatorname{E-IC}_{*}(I^{RS}_2, \mathcal{E}_{[\mathcal{T}^{RS}_2,\mathcal{T}^{RS}_1]}) - I^{RS}_1 \|_1.
    \end{aligned}
\end{equation}

\par 
In the self-supervised learning framework, $\mathcal{L}_{lc}$ supervises the structure of reconstruction by constraining the same latent GS image restored from different RS inputs. By exploiting both RS2GS and GS2RS conversions, $\mathcal{L}_{cc}$ ensures the stable brightness for image transitions between the RS and GS domains. Lastly, $\mathcal{L}_{tc}$ leverages information from adjacent RS frames, providing strong supervision for learning inter/intra-frame relationships from events.

\subsubsection{Optimization} \label{subSec:Optimization}
\par
In addition to the three constraints described in \cref{subSec:Self-supervised Learning Framework}, we also employ the Total Variation (TV) loss to smooth the flow map predicted in $\operatorname{E-IC}_{{S}}$. Finally, the total self-supervisions can be summarized as follows,
\begin{equation}\label{eq:total_loss}
  \mathcal{L} = \lambda_1 \mathcal{L}_{lc} + \lambda_2 \mathcal{L}_{cc}+ \lambda_3\mathcal{L}_{tc}+ \lambda_4 \mathcal{L}_{tv},
\end{equation}
with $\lambda_1,\lambda_2$, $\lambda_3$, and $\lambda_4$ are balancing parameters and set as $\left \{1,1,1,0.01 \right \}$ for network training.
\par
All E-ICs in our \myname\ network need to be trained with the above self-supervision consistencies, including $\operatorname{E-IC}_{{S}}$, $\operatorname{E-IC}_{{T}}$, and $\operatorname{E-IC}_{{S+T}}$. And the whole network is trained in an end-to-end manner.
%


\subsection{SelfUnroll with Multiple RS Frames} \label{sec:SelfUnroll-M}
By exploiting the proposed E-ICs, our SelfUnroll-S is able to extract GS images at arbitrary timestamps from a single RS frame. However, real-world disturbances, \eg, \textit{foreground occlusions} and \textit{noise events}, might pose challenges to SelfUnroll-S. On the one hand, foreground occlusions often violate the brightness constancy assumption in $\operatorname{E-IC}_{S}$. Although $\operatorname{E-IC}_{T}$ can predict new objects in the scene, it tends to produce distorted colors due to the gap between monochrome events and color RS images.
On the other hand, real-world events are noisy due to the non-ideality of event cameras, and the accumulation of noise events often leads to performance degradation when inferring the GS images far from RS inputs.
We develop a Motion and Occlusion Aware (MOA) module to address the above limitations by leveraging two consecutive RS images and their in-between events. Then we extend SelfUnroll-S to SelfUnroll-M, which produces smoother and more accurate GS results.

 \subsubsection{Motion and Occlusion Aware Module} \label{arh SelfUnroll-M}
The goal of the proposed MOA module is to utilize the temporal motion information from multiple inputs and improve the robustness of SelfUnroll against foreground occlusions and noise events. Specifically, given two consecutive RS images $I^{RS}_1,I^{RS}_2$ captured during the exposure time $\mathcal{T}^{RS}_1,\mathcal{T}^{RS}_2$ and the target GS image $I^{GS}$ with exposure time $\mathcal{T}$, we first employ SelfUnroll-S to restore two separate GS results $\hat{I}_{1}^{GS},\hat{I}_{2}^{GS}$ from $I^{RS}_1,I^{RS}_2$, \ie,
 \begin{equation}\label{eq:ts}
\begin{aligned}
    \{\hat{I}_{1}^{GS},\mathcal{F}_{1},&\tilde{\mathcal{I}}_{1}\}= \operatorname{SelfUnroll-S}(I^{RS}_1, \mathcal{E}_{[\mathcal{T}^{RS}_1,\mathcal{T}]}),
\\
  \{\hat{I}^{GS}_{2},\mathcal{F}_{2},&\tilde{\mathcal{I}}_{2}\}= \operatorname{SelfUnroll-S}(I^{RS}_2, \mathcal{E}_{[\mathcal{T}^{RS}_2,\mathcal{T}]}),
\
\end{aligned}
\end{equation}
where $\mathcal{F}_{1} \triangleq \mathcal{F}(\mathcal{E}_{[\mathcal{T}^{RS}_1,\mathcal{T}]}), \mathcal{F}_{2} \triangleq \mathcal{F}(\mathcal{E}_{[\mathcal{T}^{RS}_2,\mathcal{T}]})$ and 
$\tilde{\mathcal{I}}_{1} \triangleq \tilde{\mathcal{I}}(\mathcal{E}_{[\mathcal{T}^{RS}_1,\mathcal{T}]}), \tilde{\mathcal{I}}_{2} \triangleq \tilde{\mathcal{I}}(\mathcal{E}_{[\mathcal{T}^{RS}_2,\mathcal{T}]})$, and the coordinate $\mathbf{x}$ is omitted for readability. Then we combine the spatial and temporal transition information output from SelfUnroll-S with events,
 \begin{equation}\label{eq:occlusion attention}
\begin{aligned}
 S_{1}=\{\mathcal{E}_{[\mathcal{T}^{RS}_1,\mathcal{T}]}&,\mathcal{F}_{1},\tilde{\mathcal{I}}_{1}\},
\\
 S_{2}=\{\mathcal{E}_{[\mathcal{T}^{RS}_2,\mathcal{T}]}&,\mathcal{F}_{2},\tilde{\mathcal{I}}_{2}\},
\end{aligned}
\end{equation}
which are fed to the MOA module to predict the confidence map $m$ for temporal information fusion,
\begin{equation}\label{eq:temporalinfofusion}
\begin{aligned}
&\hat{I}^{GS}= m\hat{I}^{GS}_{1}+(1-m)\hat{I}^{GS}_{2}, 
\\
&\text{where}\quad m = \operatorname{MOA}(S_{1},S_{2}).
\end{aligned}
\end{equation}
Here $\hat{I^{GS}}$ indicates the final reconstruction result.

\par 
Overall, SelfUnroll-M can be regarded as the combination of a SelfUnroll-S (or two weight-sharing SelfUnroll-S) and the MOA module.
The SelfUnroll-M is fed with a pair of inputs, \ie, two consecutive RS frames $I^{RS}_1, I^{RS}_2$ and the events between the target GS frame and input RS frames $\mathcal{E}_{[\mathcal{T}^{RS}_1,\mathcal{T}]},\mathcal{E}_{[\mathcal{T}^{RS}_2,\mathcal{T}]}$, and outputs the GS frame $\hat{I}^{GS}$. For clarity, we summarize the function of SelfUnroll-M as
  \begin{equation}\label{eq:SelfUnroll-M}
\begin{aligned}
\hat{I}^{GS} =\operatorname{SelfUnroll-M}(I^{RS}_1, I^{RS}_2, \mathcal{E}_{[\mathcal{T}^{RS}_1,\mathcal{T}]}, \mathcal{E}_{[\mathcal{T}^{RS}_2,\mathcal{T}]}).
\end{aligned}
\end{equation}

\subsubsection{Optimization} \label{loss SelfUnroll-M}
Since SelfUnroll-M is built upon SelfUnroll-S, training SelfUnroll-M is equivalent to training the additional MOA module. 
Provided with the SelfUnroll-S pre-trained using \cref{eq:total_loss}, we fix the weights of SelfUnroll-S and formulate a dual cycle consistency $\mathcal{L}_{dcc}$ to train the MOA module in a self-supervised manner. Based on \cref{eq:SelfUnroll-M}, we first employ SelfUnroll-M to restore an intermediate GS image $\hat{I}^{GS}$ at exposure time $\mathcal{T}$ from two RS frames $I^{RS}_1,I^{RS}_2$ and the corresponding events $\mathcal{E}_{[\mathcal{T}^{RS}_1,\mathcal{T}]},\mathcal{E}_{[\mathcal{T}^{RS}_2,\mathcal{T}]}$.  By exploiting the flexible conversion ability of SelfUnroll-M between RS and GS domains, we treat $\hat{I}^{GS}$ as one reference image and estimate the original RS inputs,
\begin{equation}
\begin{aligned}
    \label{eq:newccloss_inputs}
     \tilde{I}^{RS}_{1}= &\operatorname{SelfUnroll-M}\left(\hat{I}^{GS}, {I}^{RS}_2,\mathcal{E}_{[\mathcal{T},\mathcal{T}^{RS}_2]},\mathcal{E}_{[\mathcal{T}^{RS}_1,\mathcal{T}^{RS}_2]}\right),
    \\
    \tilde{I}^{RS}_{1}=&\operatorname{SelfUnroll-M}\left(\hat{I}^{GS}, {I}^{RS}_2,\mathcal{E}_{[\mathcal{T},\mathcal{T}^{RS}_1]},\mathcal{E}_{[\mathcal{T}^{RS}_2,\mathcal{T}^{RS}_1]}\right).
\end{aligned}
\end{equation}
Then one can formulate the dual cycle consistency loss as 
\begin{equation}\label{eq:newccloss}
    \mathcal{L}_{dcc}=\left\| \tilde{I}^{RS}_{1}  -  I^{RS}_1 \right\|_1 +\left\| \tilde{I}^{RS}_{2} -  I^{RS}_2 \right\|_1.
\end{equation}


%

\subsection{Network Details and Event Representation}\label{sec:net_detail_event_repre}

\noindent\textbf{Network Details.}
We adopt the residual dense network~\cite{Jin_2019_CVPR} as  $\operatorname{E-IC}_{{T}}$. 
The $\operatorname{E-IC}_{{S}}$ , fusion module, and MOA module are implemented based on UNet\cite{ronneberger2015u}. Note that all the modules in our network are trained from scratch.

\noindent\textbf{Event Segments.} As described in \cref{eq:eic}, our proposed $\operatorname{E-IC}$s enable flexible RS2RS, RS2GS, and GS2RS transformations based on the corresponding event segment $\mathcal{E}_{R\rightarrow R}$, $\mathcal{E}_{R\rightarrow G}$, $\mathcal{E}_{G\rightarrow R}$. 
Here we explain the principle of event segments using the RS2GS case.
Since the RS image is exposed row by row during exposure time interval $[0,T]$, each row has a row-specific exposure timestamp determined by the location-aware time shifting operator $\mathcal{T}_{\bf x}^{RS}$. For the latent GS image, all pixels are exposed simultaneously at a single and fixed timestamp $t$. Thus, we define the event segment for the RS2GS transformation by $\mathcal{E}_{R\to G}\triangleq \mathcal{E}_{[\mathcal{T}_{\bf x}^{RS},t]}$, corresponding to the events located at the green faded regions in \cref{fig:event_seg}. The event segment varies with respect to the latent GS timestamps and we give five illustrative examples including three GS images inside the RS exposure time ($t=0.25, 0.5, 0.75T$ in \cref{fig:event_seg}) and two outside the RS exposure time ($t=-0.25, 1.25T$ in \cref{fig:event_seg}).

\noindent\textbf{Event Representation.} According to $\mathcal{E}_{[\mathcal{T}_{\bf x}^{RS},t]}$, events in $\mathcal{E}_{R\to G}$ of the different rows have different time intervals, therefore we stack events in a row-aware manner. For row $y$, we evenly divide $N$ temporal bins ($N=16$ in this work) between the latent GS timestamp $t$ and the RS timestamp  $\mathcal{T}^{RS}_{\bf x}=y\frac{|T|}{H}$ with $H$ the image height and $T$ the RS exposure time interval. To guarantee the performance of forward (\ie, $\mathcal{T}^{RS}_{\bf x}<t$) and backward (\ie, $\mathcal{T}^{RS}_{\bf x}>t$) conversion, we apply time flip and polarity reversal to events using the event pre-processing operator in \cite{zhang2022unifying}.
We then accumulate the events inside each temporal bin and form a $2N\times 1 \times W$ tensor. The above operation is repeated for each row, and we finally concatenate them to form a $2N\times H \times W$ event tensor with $2, H, W $ indicating event polarity, image height, and width, respectively.

\begin{table}[t]
\caption{Overview of datasets for event-based RS correction. Data\textsuperscript{*}: including RS frames and events. {\textdagger: the Gev-RS-Real dataset is built by two different event cameras with resolutions of 1280$\times$720 and 346$\times$260 respectively.}}
\centering
\begin{tabular}{lcccc}

\hline
Datasets  & Data\textsuperscript{*}  & Resolution & \# Train & \# Test  \\ \hline
Fastec-RS\cite{liu2020deep}  & synthetic & 640$\times$480  & 54 & 21\\ 
Gev-RS\cite{zhou2022evunroll}  & synthetic  & 640$\times$360 & 20 & 9\\ 
Gev-RS-Real\cite{zhou2022evunroll}  & real & 346$\times$260\textsuperscript{\textdagger} & 9 & 7 \\
DRE  & real & 346$\times$260 & 75 & 25\\
\hline
\end{tabular}
\vspace{-1em}
\label{dataset}
\end{table}

\def\imwidth{0.485}
\def\subimwidth{0.245}

\begin{figure}[t]
\footnotesize
	\centering
        \begin{minipage}[t]{\linewidth}
    		\centering
			\includegraphics[width=\subimwidth\linewidth,trim={0 0 0 0},clip]{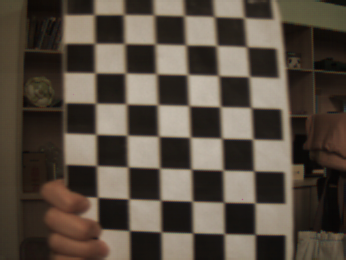}\hfill
			\includegraphics[width=\subimwidth\linewidth,trim={0 0 0 0},clip]{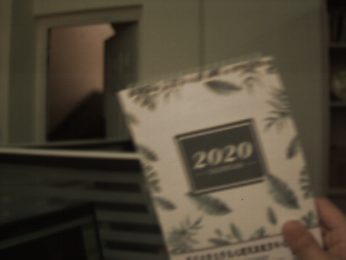}\hfill
			\includegraphics[width=\subimwidth\linewidth,trim={0 0 0 0},clip]{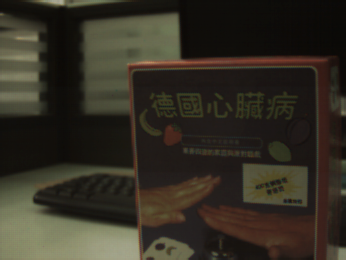}\hfill
			\includegraphics[width=\subimwidth\linewidth,trim={0 0 0 0},clip]{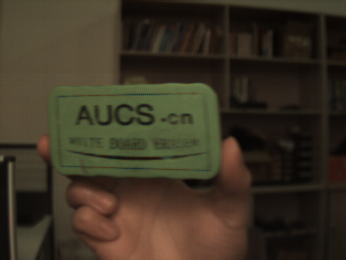}\hfill

    \end{minipage}
    \vspace{0.01 pt}
    \\
    \begin{minipage}[t]{\linewidth}
    		\centering
			\includegraphics[width=\subimwidth\linewidth,trim={0 0 0 0},clip]{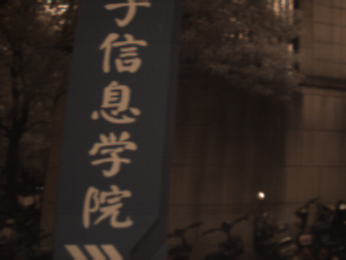}\hfill
			\includegraphics[width=\subimwidth\linewidth,trim={0 0 0 0},clip]{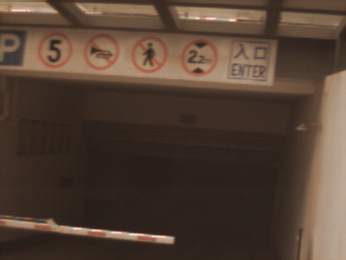}\hfill
			\includegraphics[width=\subimwidth\linewidth,trim={0 0 0 0},clip]{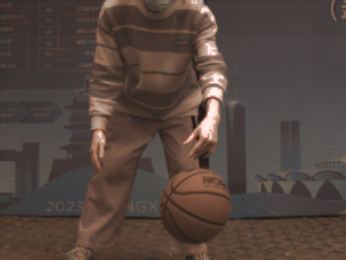}\hfill
			\includegraphics[width=\subimwidth\linewidth,trim={0 0 0 0},clip]{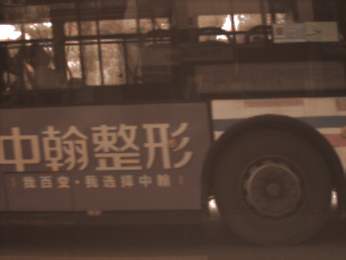}\hfill
     (a) Examples of scenes on the DRE dataset \vspace{1.5mm}
    \end{minipage}
    \\
    	\begin{minipage}[t]{\imwidth\linewidth}
    		\centering
			\begin{tikzpicture}[spy using outlines={rectangle,green,magnification=\ssmag,size=\ssizz},inner sep=0]
				\node {\includegraphics[width=\linewidth]{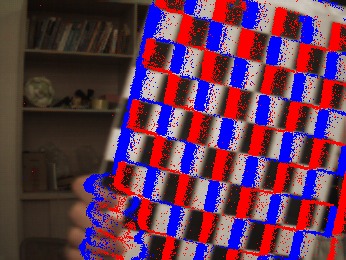}};
				\end{tikzpicture}
        (b) Before line delay calibration \\
        (events in GS exposure)\vspace{0.5em}
    	\end{minipage}%
    \hfill
    	\begin{minipage}[t]{\imwidth\linewidth}
    		\centering
    		\begin{tikzpicture}[spy using outlines={rectangle,green,magnification=\ssmag,size=\ssizz},inner sep=0]
				\node {\includegraphics[width=\linewidth]{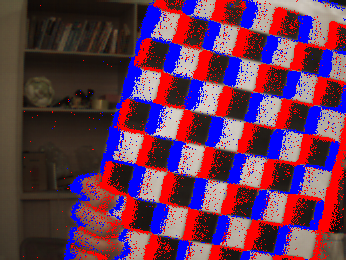}};
				\end{tikzpicture}
			(c) After line delay calibration \\
        (events in RS exposure)\vspace{0.5em}
    	\end{minipage}%

	\caption{
 \zhang{Examples of the proposed DAVIS-RS-Event (DRE) dataset. (a) Indoor (top) and outdoor (bottom) samples captured under different camera motions and dynamic scenes. (b) Before calibration, the line delay is set to zero, resulting in a misalignment between the RS image and events. (c) After calibration, the line delay is adjusted to align the RS image and events.}
 }\vspace{-1em}
	\label{fig:calibration}
\end{figure}

\section{Experiment}\label{Sec:Experiment}
\subsection{Datasets}\label{sec:datasets}
We evaluate the proposed algorithm on four datasets, including \textbf{two synthetic datasets}, \ie, Fastec-RS\cite{liu2020deep} and  Gev-RS\cite{zhou2022evunroll}, with simulated events and RS images, and \textbf{two real-world datasets}, \zhang{\ie, Gev-RS-Real\cite{zhou2022evunroll} and our proposed DAVIS-RS-Event (DRE), where DRE provides more sequences (100 \vs 16) with a richer diversity of scenes than Gev-RS-Real. We summarize the above datasets in \cref{dataset}.
}

%


%
%

\par
\noindent\textbf{Fastec-RS.} The Fastec-RS\cite{liu2020deep} dataset captures multiple 2,400 FPS GS videos by a Fastec TS5 high-speed GS camera at the resolution of 640$\times$480, and then synthesizes RS images. The corresponding event streams are synthesized with V2E \cite{delbruck2020v2e} using the high frame-rate GS video. We follow the same data splitting strategy as DSUN \cite{liu2020deep} and EvUnroll \cite{zhou2022evunroll}, \ie, 54 sequences for training and 21 for testing.


\par
\noindent\textbf{Gev-RS.} The Gev-RS\cite{zhou2022evunroll} dataset uses a high-speed Phantom VEO 640 camera to collect 5,700 FPS GS videos with the $640\times360$ image resolution. Furthermore, the event streams are simulated with DVS-Voltmeter \cite{lin2022dvsvoltmeter}, and RS effects are generated by the same method as Fastec-RS\cite{liu2020deep}. The Gev-RS dataset has 29 paired sequences composed of 3,700 GS-event-RS triplet clips. Similar to EvUnroll\cite{zhou2022evunroll}, we divide them into the training and testing datasets at a ratio of 7:3.

\par
\noindent\textbf{Gev-RS-Real.} The Gev-RS-Real dataset \cite{zhou2022evunroll} is built with the real-world events and RS images captured by a hybrid camera system consisting of an RS camera and an event camera. The event streams are collected by two different event cameras, \ie, DAVIS346 and PROPHESEE GEN4.0, for data diversity, and the ground-truth GS references are not available. In total, Gev-RS-Real has 16 event-RS paired sequences, where we select 9 of them for training and the other 7 for testing.
\par

\begin{figure}[t]
    \centering
    \includegraphics[width=1\linewidth]{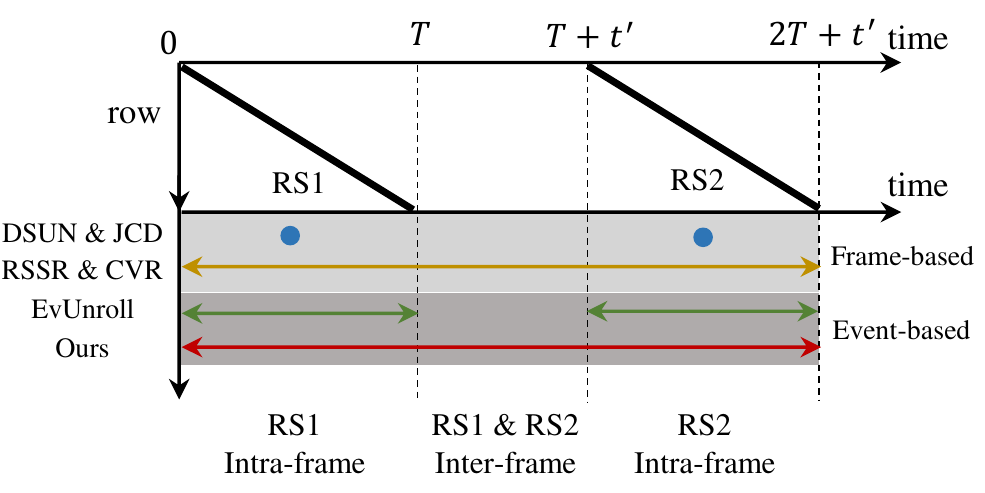}
    \caption{\zhang{Illustration of inter-frame and intra-frame time of RS video, where $T$ represents the exposure length of RS frames and $t^{\prime}$ denotes the interval time between consecutive RS frames. The output settings of RS2GS approaches are demonstrated in the shaded regions. DSUN\cite{liu2020deep} and JCD\cite{zhong2021towards} only recover GS frame at a pre-defined timestamp, and EvUnroll\cite{zhou2022evunroll} is dedicated to restoring the intra-frame GS results. By contrast, RSSR\cite{fan2021inverting}, CVR\cite{fan2022context}, and our proposed SelfUnroll can achieve SDR and restore GS frames in both intra-frame and inter-frame time.
    }
    }
    \label{fig:inter_intra_time}
\end{figure}
\noindent\textbf{DAVIS-RS-Event (DRE).}
\zhang{We construct a real-world dataset named DAVIS-RS-Event (DRE) with a variety of indoor and outdoor scenes to foster RS correction research. We use a DAVIS346 camera to capture real-world RS images and the corresponding event streams, and further apply the calibration tool Kalibr\cite{oth2013rolling} to determine the line delay between RS frames and events, as demonstrated in \cref{fig:calibration}.
In total, our DRE contains more sequences (\cref{dataset}) and richer diversity of scenes than the abovementioned datasets, which provides a more comprehensive dataset for RS2GS evaluation in real-world scenarios.
}

\input{Figs/singel-frame}
\begin{table}[t]
\centering
\caption{Overview of RS inversion methods.}
\resizebox{1\linewidth}{!}{
\begin{tabular}{lccccc}
\hline
\multirow{2}{*}{Methods} & \multicolumn{2}{c}{Input} & \multirow{2}{*}{Output} & \multirow{2}{*}{SDR} & \multirow{2}{*}{Unsupervised} \\ \cline{2-3}
 & Frame & Event &       &     &                   \\ \hline
DSUN\cite{liu2020deep} & multiple     &     $\times$  & frame &     $\times$ & $\times$           \\
 JCD\cite{zhong2021towards}& multiple     &   $\times$    & frame  &     $\times$& $\times$          \\
RSSR\cite{fan2021inverting} & multiple     &  $\times$     & video  &     \checkmark& $\times$            \\
CVR\cite{fan2022context}   & multiple     &   $\times$    & video &     \checkmark & $\times$           \\
 EvUnroll\cite{zhou2022evunroll}& single     &  \checkmark     & video  &     $\times$& $\times$            \\
 \hline
 SelfUnroll-S & single    &   \checkmark    & video &     \checkmark & \checkmark \\
 SelfUnroll-M & multiple     &    \checkmark   & video &     \checkmark & \checkmark \\ \hline
\end{tabular}
}
\label{tab:methods}
\end{table}

\begin{table}[!t]
\caption{Quantitative comparisons with respect to the single GS frame reconstruction on the {Fastec-RS} dataset. \textcolor{red}{\textbf{Bold}} and \textcolor{blue}{\underline{underlined}} numbers represent the best and the second-best performance.}
\centering
\begin{tabular}{lccc}
\hline
Methods  & PSNR$\uparrow$   & SSIM$\uparrow$  & LPIPS$\downarrow$   \\ \hline
DSUN\cite{liu2020deep}     & 26.52  & 0.792 & 0.1222  \\
JCD \cite{zhong2021towards}     & 24.84  & 0.780  & 0.1070   \\
RSSR \cite{fan2021inverting}    & 21.23  & 0.776 & 0.1659 \\
CVR \cite{fan2022context}     & 28.72  & 0.847 & 0.1107  \\
EvUnroll \cite{zhou2022evunroll} & {31.32}  & {0.880}  & {0.0840}   \\ 
\myname-S\ (Ours)     & \textcolor{blue}{\underline{31.85}} & \textcolor{blue}{\underline{0.894}} & \textcolor{blue}{\underline{0.0723}} \\ 
\myname-M\ (Ours)     & \textcolor{red}{\bf{32.34}} & \textcolor{red}{\bf{0.901}} & \textcolor{red}{\bf{0.0722}} \\ \hline
\end{tabular}
\label{table_fastec}
\end{table}




\begin{table*}[!t]
    \centering
    \caption{Quantitative comparisons of the proposed SelfUnrolls to the state-of-the-art methods on the {Gev-RS} dataset. Given an RS frame with the exposure time $[0,T]$, all methods output 9 GS frames at timestamps $t\in \{0.1T,0.2T,0.3T,0.4T,0.5T,0.6T,0.7T,0.8T,0.9T\}$ in the GS video sequence reconstruction task. In the single-frame reconstruction task, we evaluate the middle frame at the timestamp $t=0.5T$ of the whole video. \textcolor{red}{\textbf{Bold}} and \textcolor{blue}{\underline{underlined}} numbers represent the best and the second-best performance. The symbol / denotes infeasible to reconstruct GS sequences.}
    \begin{tabular}{l p{1.5cm}<{\centering} p{1.5cm}<{\centering} p{1.5cm}<{\centering} c p{1.5cm}<{\centering} p{1.5cm}<{\centering} p{1.5cm}<{\centering}}
        \hline
        \multirow{2}{*}{Method} & \multicolumn{3}{c}{Single GS frame} & & \multicolumn{3}{c}{GS video sequence} \\
        \cline{2-4} \cline{6-8}
        & PSNR$\uparrow$ & SSIM$\uparrow$ & LPIPS$\downarrow$ & & PSNR$\uparrow$ & SSIM$\uparrow$ & LPIPS$\downarrow$ \\
        \hline
        DSUN \cite{liu2020deep} & 20.13 & 0.654 & 0.0864 && / & / & / \\
        JCD \cite{zhong2021towards} & 20.04 & 0.661 & 0.1796 && / & /& / \\
        RSSR \cite{fan2021inverting} & 22.21 & 0.749 & 0.1011 && 20.39 & 0.680 & 0.1195 \\
        CVR \cite{fan2022context} & 23.18 & 0.766 & 0.1022 && 23.52 & 0.771 & 0.1076 \\
        EvUnroll \cite{zhou2022evunroll} & {31.29} & {0.914} & {0.0230} && {29.41} & {0.896} & {0.0383} \\
        \myname-S\ (Ours) & \textcolor{blue}{\underline{32.26}} & \textcolor{blue}{\underline{0.926}} & \textcolor{blue}{\underline{0.0204}} && \textcolor{blue}{\underline{31.95}} & \textcolor{blue}{\underline{0.923}} & \textcolor{blue}{\underline{0.0231}} \\
        \myname-M\ (Ours) & \textcolor{red}{\textbf{32.62}} & \textcolor{red}{\textbf{0.932}} & \textcolor{red}{\textbf{0.0200}} && \textcolor{red}{\textbf{32.71}} & \textcolor{red}{\textbf{0.934}} & \textcolor{red}{\textbf{0.0194}} \\
        \hline
    \end{tabular}
    \label{tab:gev-comparison}
\end{table*}
\def\imwidth{0.498}
\def\cimwid{0.14}

\def\zuoxia{(-0.3,-0.9)}
\def\youshang{(0.2,0.45)}

\def\ssyy{(-0.8,-0.85)}
\def\ssizz{0.5cm}
\def\sswidth{0.245\textwidth}
\def\ssmag{3}
\def\scc{(2.12,1.4)}

\begin{figure}[!htp]
\footnotesize
	\centering

    	\begin{minipage}[t]{\imwidth\linewidth}
    		\centering
			\begin{tikzpicture}[spy using outlines={rectangle,green,magnification=\ssmag,size=\ssizz},inner sep=0]
				\node {\includegraphics[width=\linewidth]{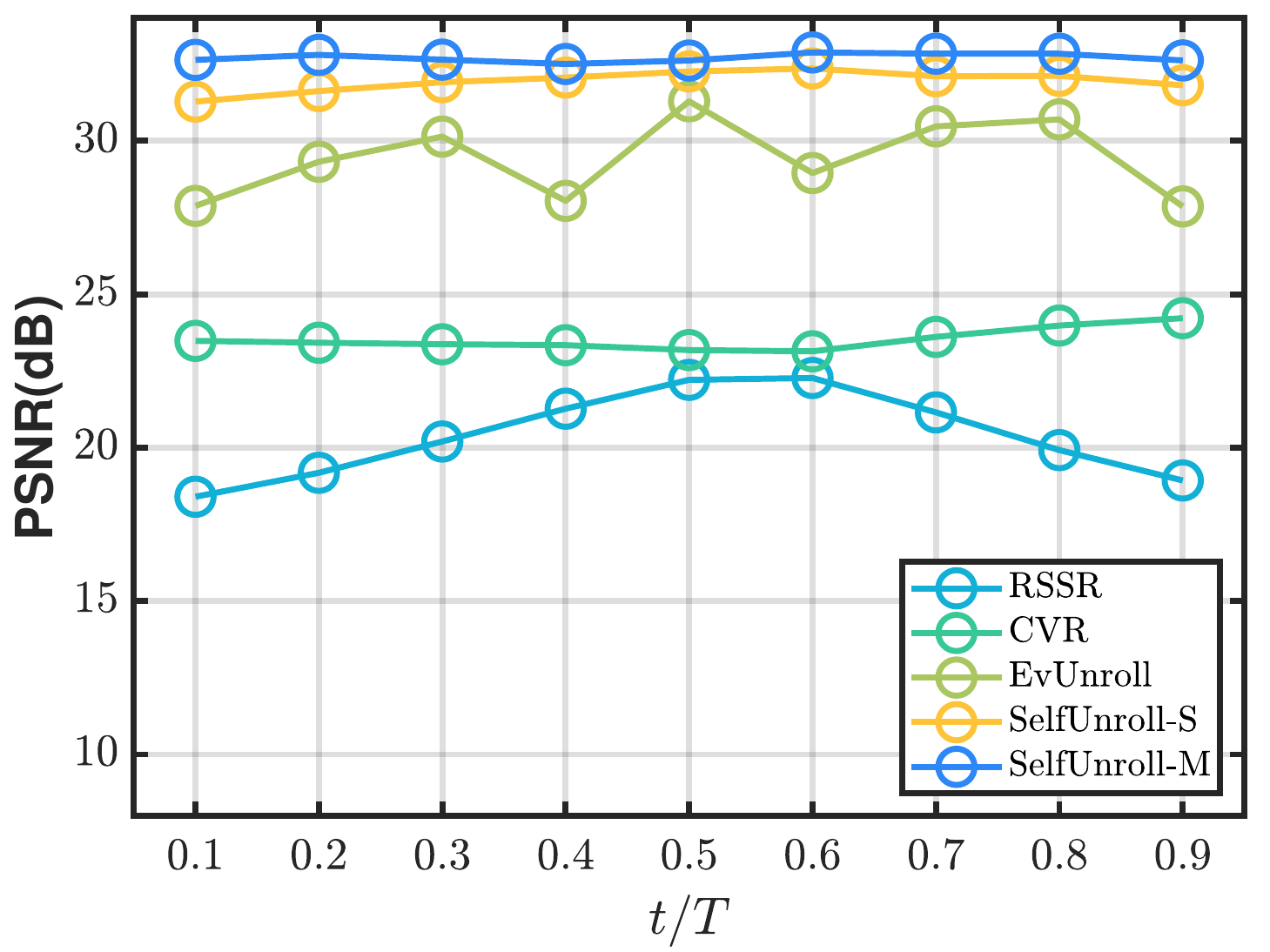}};
				\end{tikzpicture}
    (a) All methods \vspace{0.5em}
    	\end{minipage}%
    \hfill
    	\begin{minipage}[t]{\imwidth\linewidth}
    		\centering
    		\begin{tikzpicture}[spy using outlines={rectangle,green,magnification=\ssmag,size=\ssizz},inner sep=0]
				\node {\includegraphics[width=\linewidth]{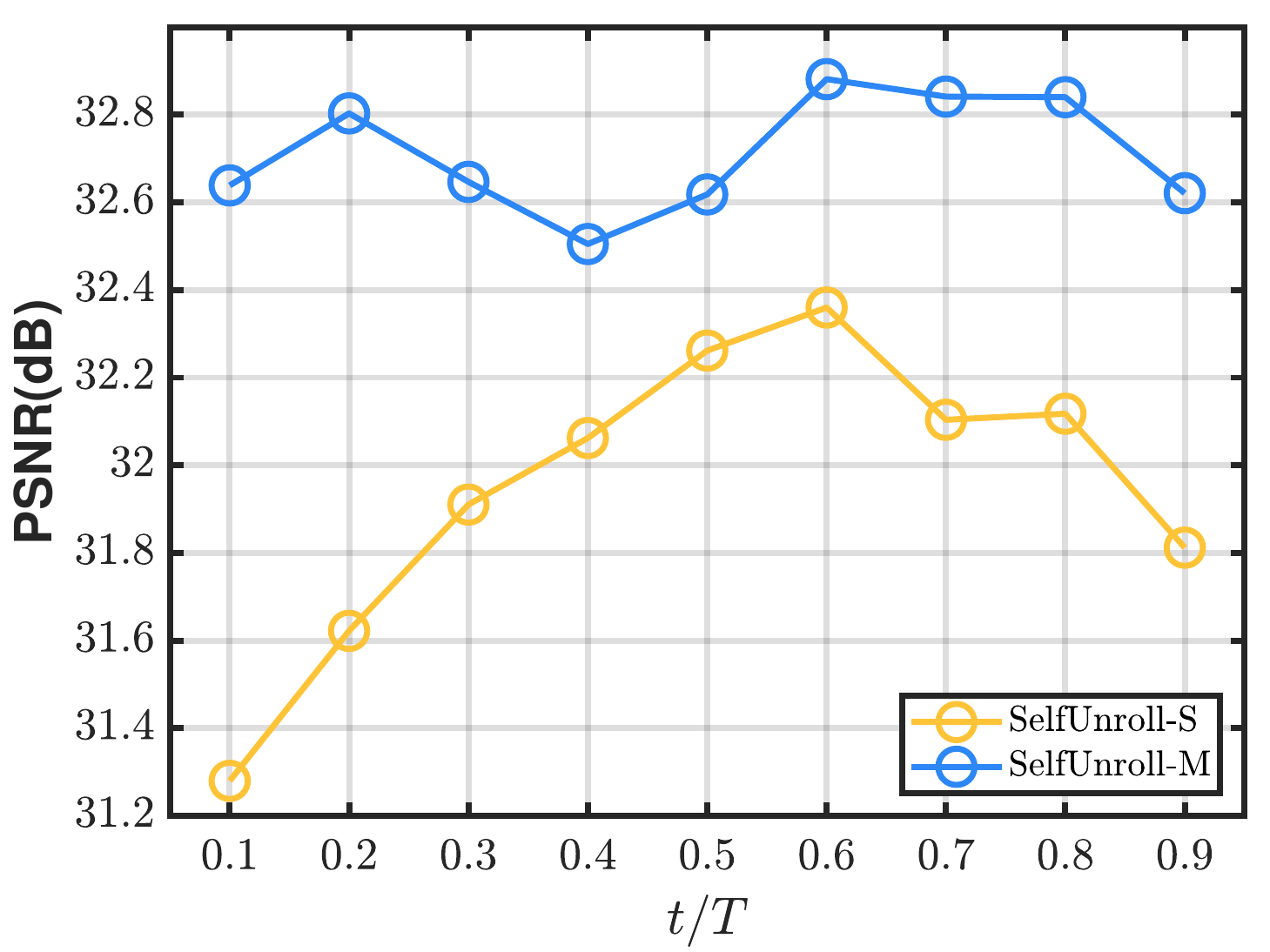}};
				\end{tikzpicture}
			(b) SelfUnroll-S \vs SelfUnroll-M \vspace{0.5em}
    	\end{minipage}%

	\caption{Evaluation of Intra-frame GS video reconstruction at different target exposure timestamps on the Gev-RS dataset.}
	\label{fig:sequence-graph}
\end{figure}

\par
\subsection{Implementation Details}
 \noindent\textbf{Training.} We use PyTorch \cite{paszke2019pytorch} to implement the proposed network with an NVIDIA GeForce RTX 3090 GPU. 
\zhang{We randomly crop the RS images to $128\times128$ patches for training. Adam optimizer \cite{kingma2014adam} and SGDR scheduler \cite{loshchilov2016sgdr} are employed with an initial learning rate set to $1\times 10^{-4}$. Our models are trained in a two-stage manner: we first train the SelfUnroll-S using \cref{eq:total_loss} for 100 epochs and then train SelfUnroll-M via \cref{eq:newccloss} for another 100 epochs. Both SelfUnroll-S and SelfUnroll-M are trained with events and RS frames, and the ground-truth GS images are only used for performance evaluation.
}

\par
\noindent{\bf Evaluation Metrics.} For quantitative evaluation, we take the Peak Signal-to-Noise Ratio (PSNR), Structural SIMilarity (SSIM), and Learned Perceptual Image Patch Similarity (LPIPS). \zhang{Better reconstruction results are indicated by high PSNR and SSIM scores, as well as low LPIPS scores, \ie, PSNR $\uparrow$, SSIM $\uparrow$, and LPIPS $\downarrow$.}

\par

\subsection{Comparison on the Synthetic Datasets}\label{Comparison on the Synthetic Datasets}
Quantitative experiments are conducted on two synthetic datasets with GS references, \ie, Gev-RS\cite{zhou2022evunroll} and Fastec-RS\cite{liu2020deep}. Our proposed SelfUnrolls are compared against state-of-the-art RS2GS methods, including four frame-based approaches, \ie, DSUN\cite{fan2021sunet}, RSSR\cite{fan2021inverting}, JCD\cite{zhong2021towards} and CVR \cite{fan2022context}, and an event-based approach, EvUnroll\cite{zhou2022evunroll}. 
\zhang{We summarize the details of the abovementioned approaches in \cref{tab:methods,fig:inter_intra_time}. The experimental settings for all methods are the same, and we additionally retrain DSUN, RSSR, JCD, and CVR on the Gev-RS dataset for a fair comparison. In the following, we divide SDR into two subtasks, \ie, intra-frame and inter-frame reconstruction as depicted in \cref{fig:inter_intra_time}, and conduct comparisons in \cref{Comparison on Intra-frame Reconstruction,Comparison on Inter-frame Reconstruction}, respectively.
}

\def\imwidth{0.187}
\def\cimwid{0.13}
\def\rswidth{0.187}
\def\zuoxia{(-1.2,-0.9)}
\def\youshang{(-0.4,-0.45)}
\def\seqwidth{0.80}
\def\ssyy{(-0.8,-0.85)}
\def\ssizz{0.5cm}
\def\sswidth{0.245\textwidth}
\def\ssmag{3}
\def\scc{(2.12,1.4)}

\begin{figure*}[!t]
\footnotesize
	\centering

    	\begin{minipage}[t]{\imwidth\linewidth}
    		\centering
    		\begin{tikzpicture}[spy using outlines={rectangle,green,magnification=\ssmag,size=\ssizz},inner sep=0]
				\node {\includegraphics[width=\linewidth]{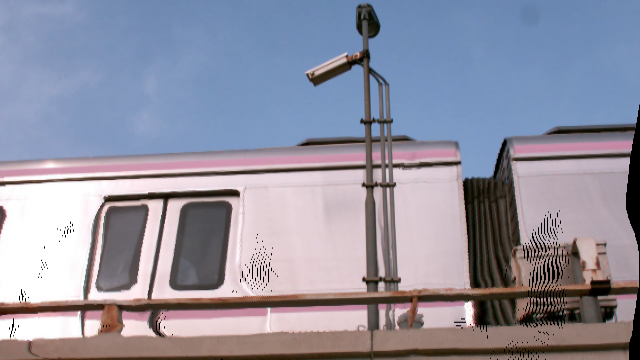}};
				\draw[green] \zuoxia rectangle \youshang;
				\end{tikzpicture}
			
			(a) RSSR\vspace{0.5em}
    	\end{minipage}%
    \hfill
    	\begin{minipage}[t]{\imwidth\linewidth}
    		\centering
    	    \begin{tikzpicture}[spy using outlines={green,magnification=\ssmag,size=\ssizz},inner sep=0]
				\node {\includegraphics[width=\linewidth]{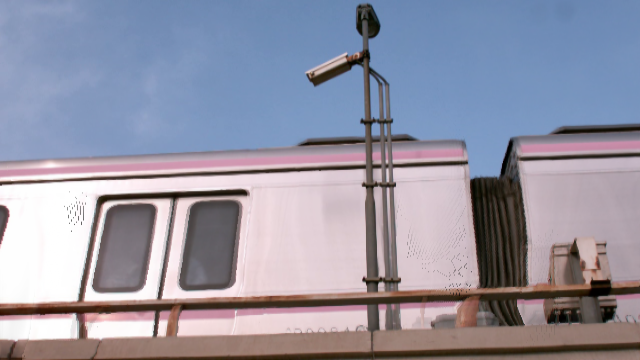}};
				\draw[green] \zuoxia rectangle \youshang;
				\end{tikzpicture}
			(b) CVR\vspace{0.5em}
    	\end{minipage}%
    \hfill
    	\begin{minipage}[t]{\imwidth\linewidth}
    		\centering
    	    \begin{tikzpicture}[spy using outlines={green,magnification=\ssmag,size=\ssizz},inner sep=0]
				\node {\includegraphics[width=\linewidth]{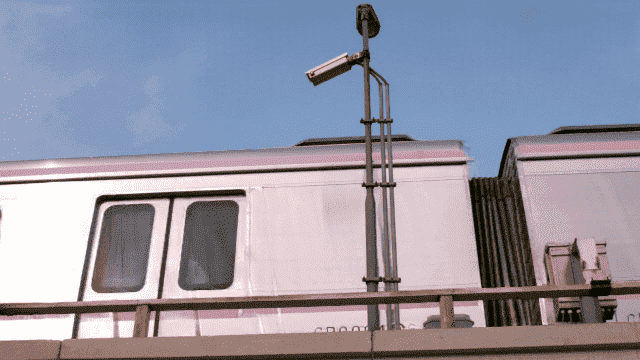}};
				\draw[green] \zuoxia rectangle \youshang;
				\end{tikzpicture}
			(c) EvUnroll\vspace{0.5em}
    	\end{minipage}%
    \hfill
		\begin{minipage}[t]{\imwidth\linewidth}
			\centering
			\begin{tikzpicture}[spy using outlines={green,magnification=\ssmag,size=\ssizz},inner sep=0]
				\node {\includegraphics[width=\linewidth]{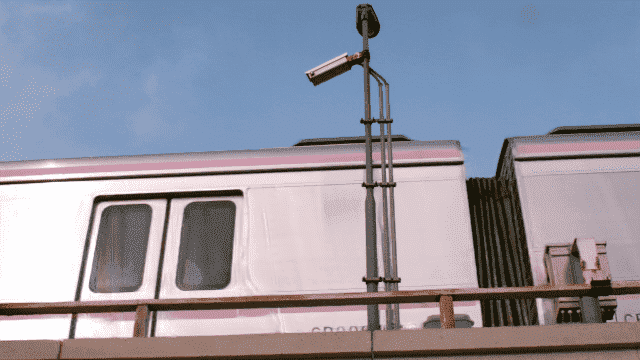}};
				\draw[green] \zuoxia rectangle \youshang;
				\end{tikzpicture}
			(d) SelfUnroll-S\vspace{.5em}
			
			
		\end{minipage}%
\hfill
\begin{minipage}[t]{\imwidth\linewidth}
    		\centering
    		\begin{tikzpicture}[spy using outlines={rectangle,green,magnification=\ssmag,size=\ssizz},inner sep=0]
				\node {\includegraphics[width=\linewidth]{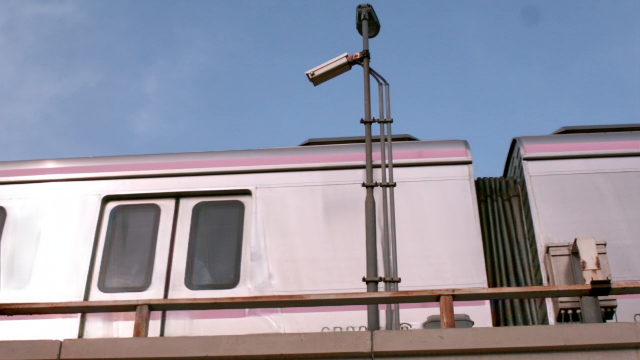}};
				\draw[green] \zuoxia rectangle \youshang;
				\end{tikzpicture}
			
			(e) SelfUnroll-M\vspace{0.5em}
    	\end{minipage}%
 \hfill
    
    
     

  \begin{minipage}[t][3.9cm][t]{\rswidth\textwidth}
  \vspace{0pt}
    		\centering
              
              \begin{tikzpicture}[spy using outlines={rectangle,green,magnification=\ssmag,size=\ssizz},inner sep=0]
        				\node {\includegraphics[width=\linewidth]{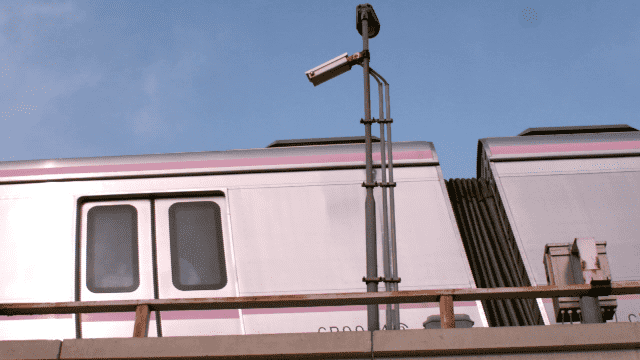}};
        				\draw[green] \zuoxia rectangle \youshang;
        				\end{tikzpicture}
            
            RS frame (previous)\vspace{0.3em}
             \vfill
			\begin{tikzpicture}[spy using outlines={rectangle,green,magnification=\ssmag,size=\ssizz},inner sep=0]
				\node {\includegraphics[width=\linewidth]{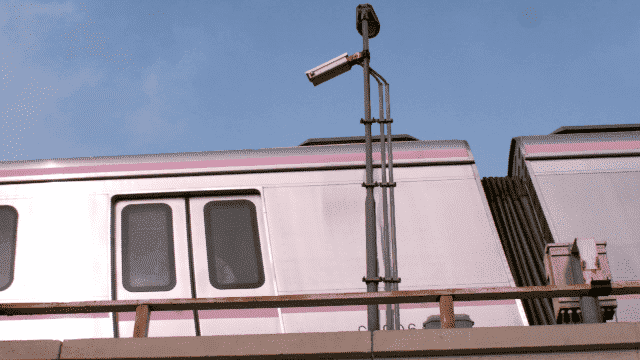}};
				\draw[green] \zuoxia rectangle \youshang;
				\end{tikzpicture}
            
           RS frame (next)\vspace{0.3em}
           \begin{tikzpicture}[spy using outlines={rectangle,green,magnification=\ssmag,size=\ssizz},inner sep=0]
        				\node {\includegraphics[width=\linewidth]{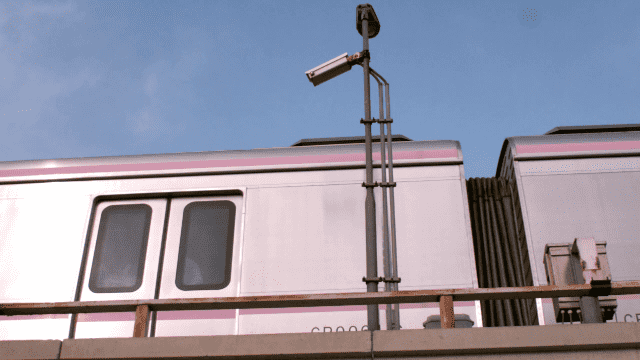}};
        				\draw[green] \zuoxia rectangle \youshang;
        				\end{tikzpicture}
            
            GT\vspace{0.3em}

    	\end{minipage}%
\hspace*{1mm}
\begin{minipage}[t]{\seqwidth\textwidth}
            \vspace{0pt}
    		\centering
      \begin{tikzpicture}[inner sep=0]
            \node [label={[label distance=0.25cm,text depth=-1ex,rotate=90]right: \textcolor{black}{\scriptsize {(a)}}}] at (0,8.7) {};
            \end{tikzpicture}
			\includegraphics[width=\cimwid\linewidth,trim={100 10 400 275},clip]{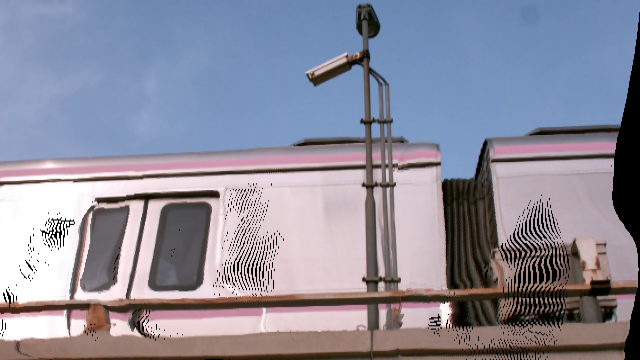}\hfill
			\includegraphics[width=\cimwid\linewidth,trim={100 10 400 275},clip]{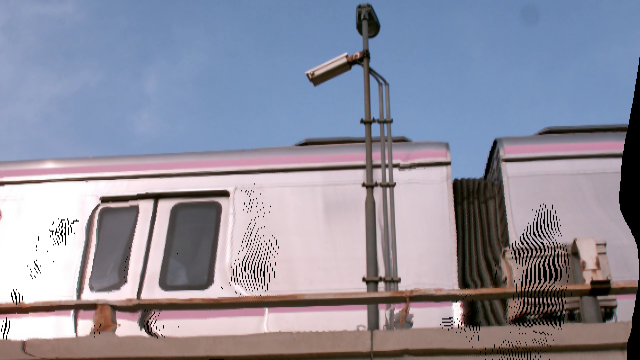}\hfill
			\includegraphics[width=\cimwid\linewidth,trim={100 10 400 275},clip]{Figs/img/NEW_sequence_gev/newrssr_4.png}\hfill
			\includegraphics[width=\cimwid\linewidth,trim={100 10 400 275},clip]{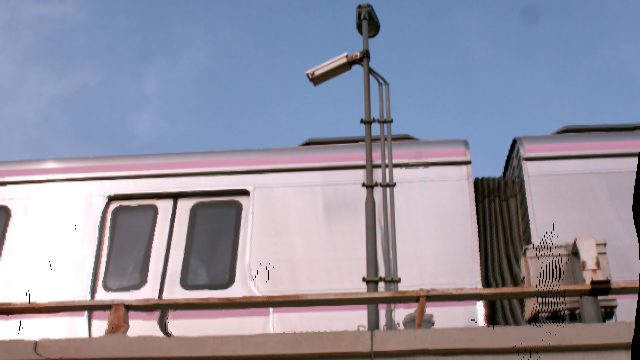}\hfill
			\includegraphics[width=\cimwid\linewidth,trim={100 10 400 275},clip]{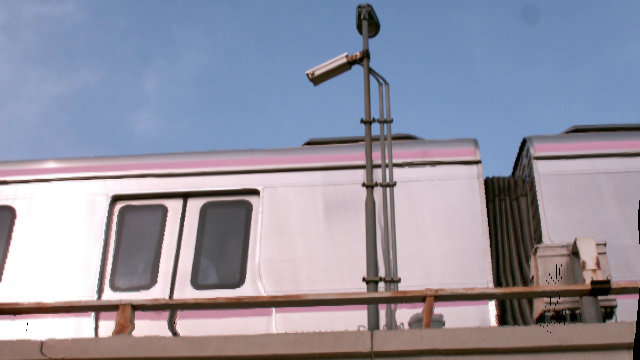}\hfill
			\includegraphics[width=\cimwid\linewidth,trim={100 10 400 275},clip]{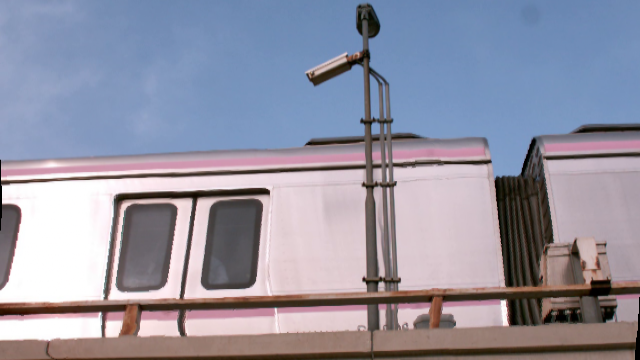}\hfill
			\includegraphics[width=\cimwid\linewidth,trim={100 10 400 275},clip]{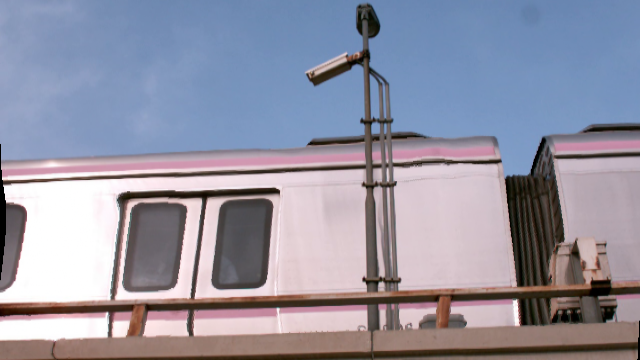}\hfill
   \\
   \vspace{0.4em}
            \begin{tikzpicture}[inner sep=0]
            \node [label={[label distance=0.25cm,text depth=-1ex,rotate=90]right: \textcolor{black}{\scriptsize {(b)}}}] at (0,8.7) {};
            \end{tikzpicture}
			\includegraphics[width=\cimwid\linewidth,trim={100 10 400 275},clip]{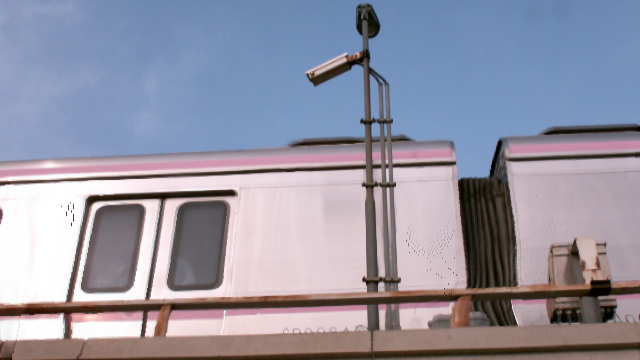}\hfill
			\includegraphics[width=\cimwid\linewidth,trim={100 10 400 275},clip]{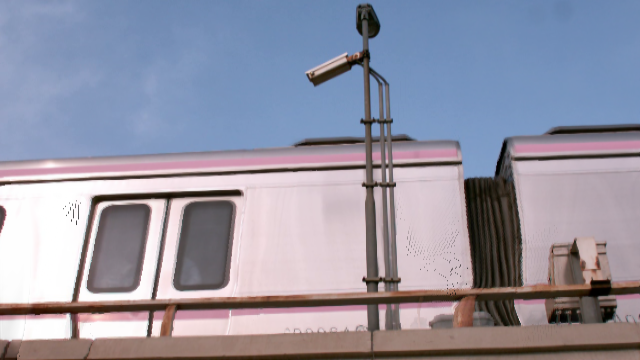}\hfill
			\includegraphics[width=\cimwid\linewidth,trim={100 10 400 275},clip]{Figs/img/NEW_sequence_gev/newcvr_4.png}\hfill
			\includegraphics[width=\cimwid\linewidth,trim={100 10 400 275},clip]{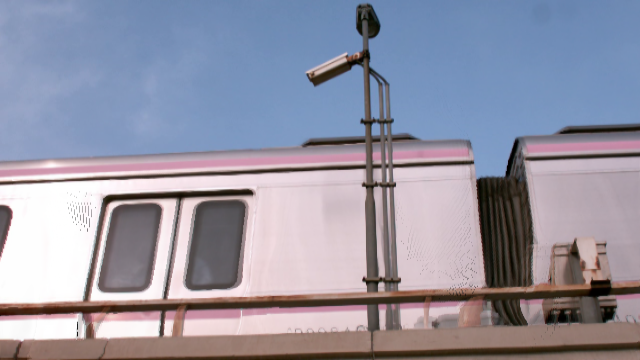}\hfill
			\includegraphics[width=\cimwid\linewidth,trim={100 10 400 275},clip]{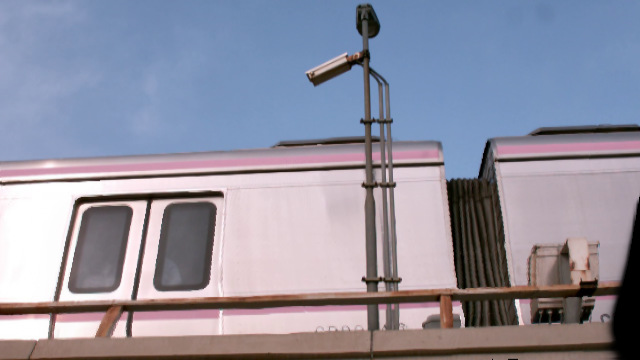}\hfill
			\includegraphics[width=\cimwid\linewidth,trim={100 10 400 275},clip]{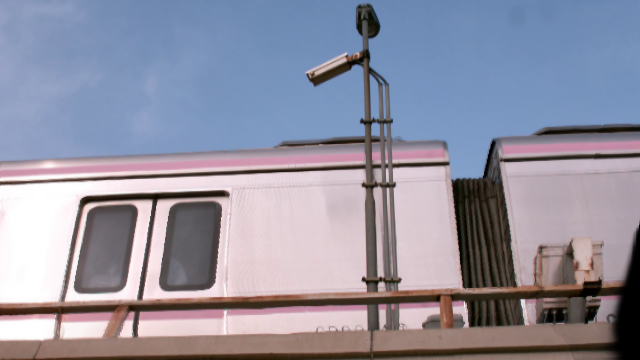}\hfill
			\includegraphics[width=\cimwid\linewidth,trim={100 10 400 275},clip]{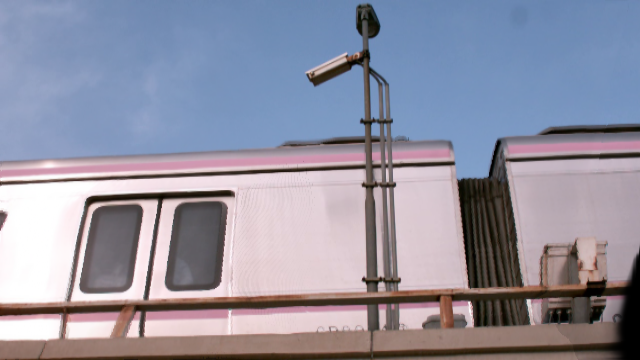}\hfill
   \\
   \vspace{0.4em}
            \begin{tikzpicture}[inner sep=0]
            \node [label={[label distance=0.25cm,text depth=-1ex,rotate=90]right: \textcolor{black}{\scriptsize {(c)}}}] at (0,8.7) {};
            \end{tikzpicture}
            \includegraphics[width=\cimwid\linewidth,trim={100 10 400 275},clip]{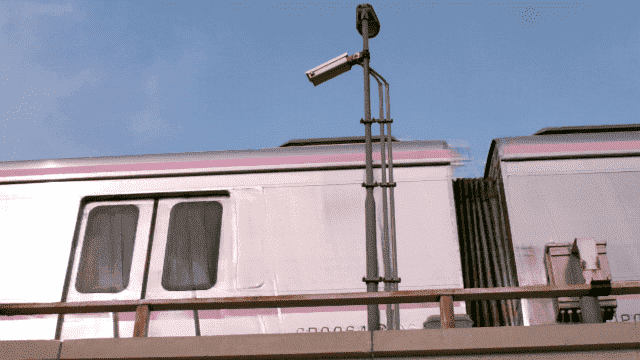}\hfill
			\includegraphics[width=\cimwid\linewidth,trim={100 10 400 275},clip]{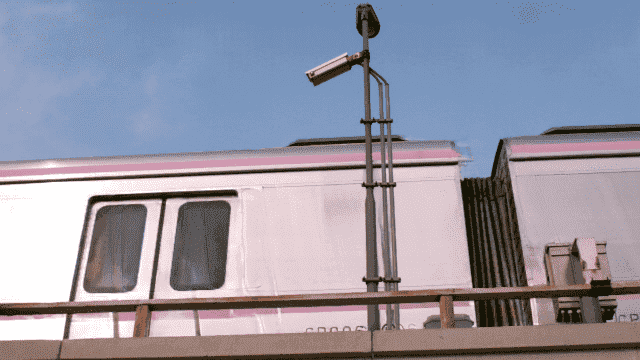}\hfill
			\includegraphics[width=\cimwid\linewidth,trim={100 10 400 275},clip]{Figs/img/NEW_sequence_gev/evunroll_4.png}\hfill
			\includegraphics[width=\cimwid\linewidth,trim={100 10 400 275},clip]{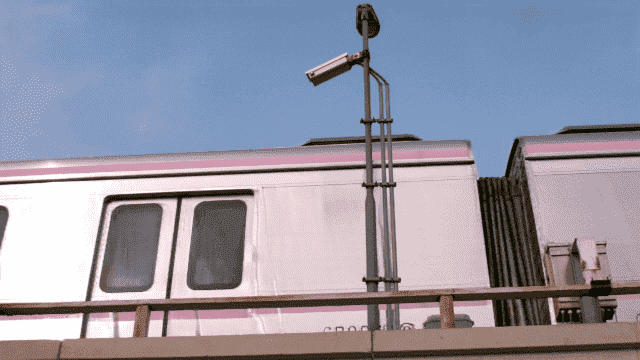}\hfill
			\includegraphics[width=\cimwid\linewidth,trim={100 10 400 275},clip]{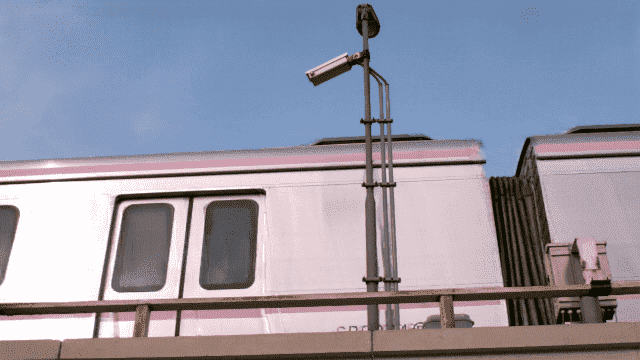}\hfill
			\includegraphics[width=\cimwid\linewidth,trim={100 10 400 275},clip]{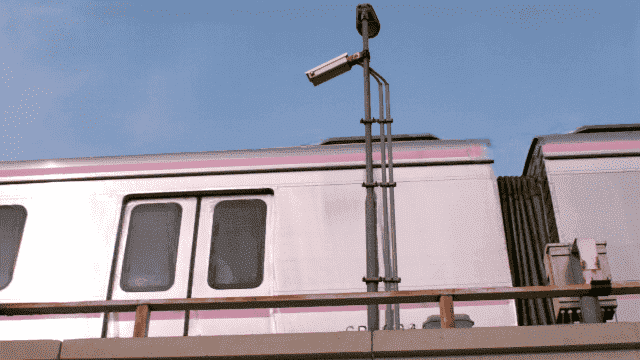}\hfill
			\includegraphics[width=\cimwid\linewidth,trim={100 10 400 275},clip]{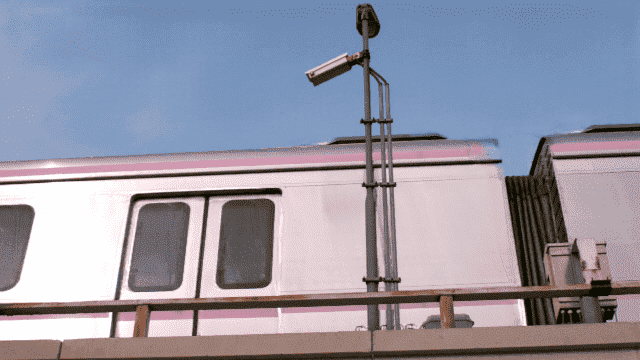}\hfill
    \\
    \vspace{0.3em}
            \begin{tikzpicture}[inner sep=0]
            \node [label={[label distance=0.25cm,text depth=-1ex,rotate=90]right: \textcolor{black}{\scriptsize {(d)}}}] at (0,8.7) {};
            \end{tikzpicture}
            \includegraphics[width=\cimwid\linewidth,trim={100 10 400 275},clip]{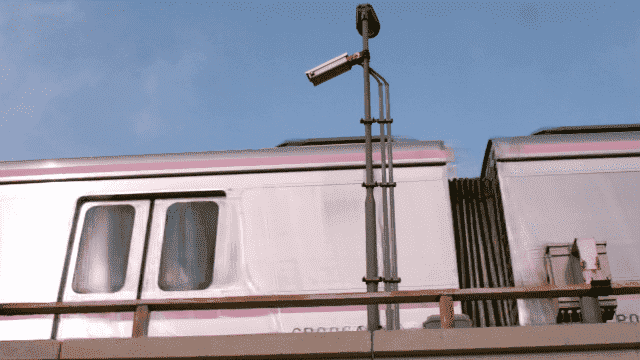}\hfill
			\includegraphics[width=\cimwid\linewidth,trim={100 10 400 275},clip]{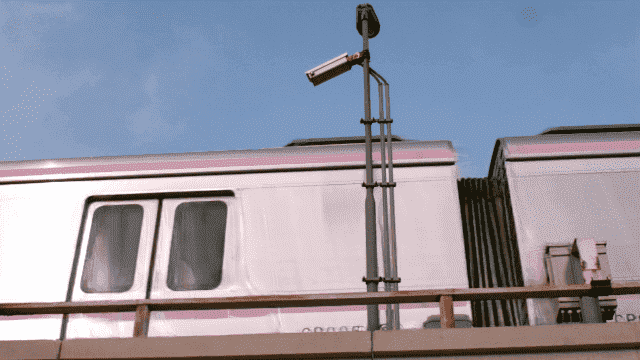}\hfill
			\includegraphics[width=\cimwid\linewidth,trim={100 10 400 275},clip]{Figs/img/NEW_sequence_gev/res_04.png}\hfill
			\includegraphics[width=\cimwid\linewidth,trim={100 10 400 275},clip]{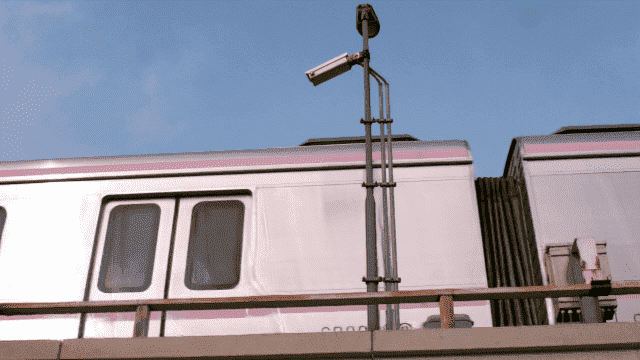}\hfill
			\includegraphics[width=\cimwid\linewidth,trim={100 10 400 275},clip]{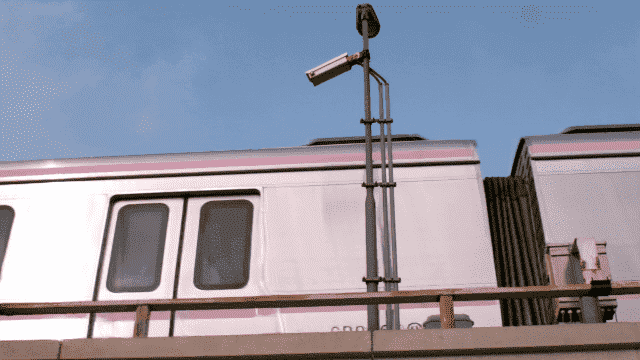}\hfill
			\includegraphics[width=\cimwid\linewidth,trim={100 10 400 275},clip]{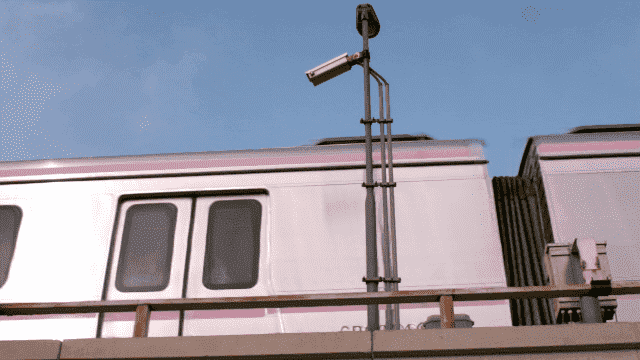}\hfill
			\includegraphics[width=\cimwid\linewidth,trim={100 10 400 275},clip]{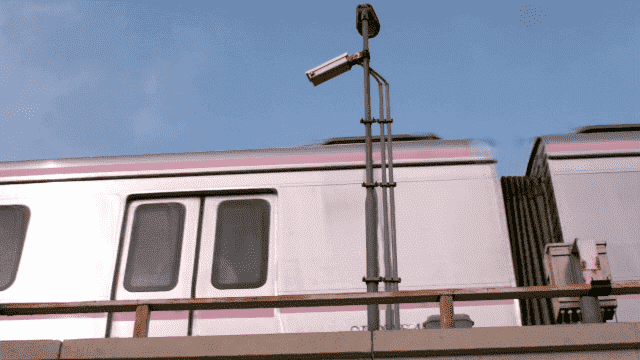}\hfill
   \\
   \vspace{0.4em}
            \begin{tikzpicture}[inner sep=0]
            \node [label={[label distance=0.25cm,text depth=-1ex,rotate=90]right: \textcolor{black}{\scriptsize {(e)}}}] at (0,8.7) {};
            \end{tikzpicture}
            \includegraphics[width=\cimwid\linewidth,trim={100 10 400 275},clip]{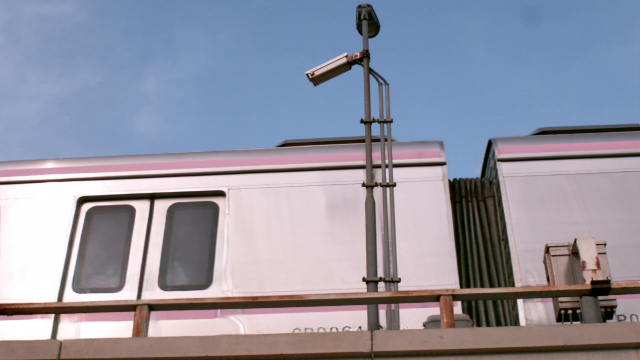}\hfill
			\includegraphics[width=\cimwid\linewidth,trim={100 10 400 275},clip]{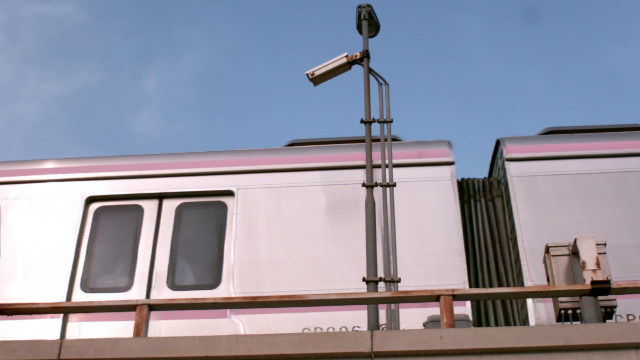}\hfill
			\includegraphics[width=\cimwid\linewidth,trim={100 10 400 275},clip]{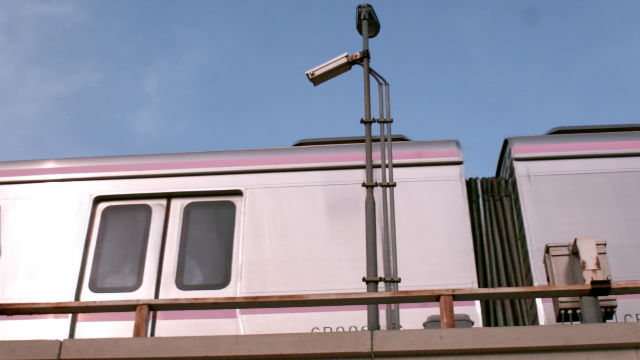}\hfill
			\includegraphics[width=\cimwid\linewidth,trim={100 10 400 275},clip]{Figs/img/NEW_sequence_gev/24209_1_33_00000019-res_400.png}\hfill
			\includegraphics[width=\cimwid\linewidth,trim={100 10 400 275},clip]{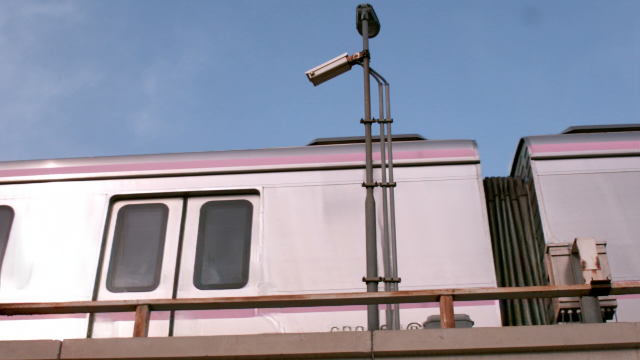}\hfill
			\includegraphics[width=\cimwid\linewidth,trim={100 10 400 275},clip]{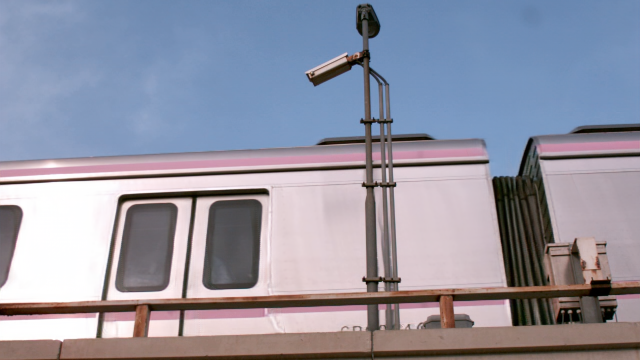}\hfill
			\includegraphics[width=\cimwid\linewidth,trim={100 10 400 275},clip]{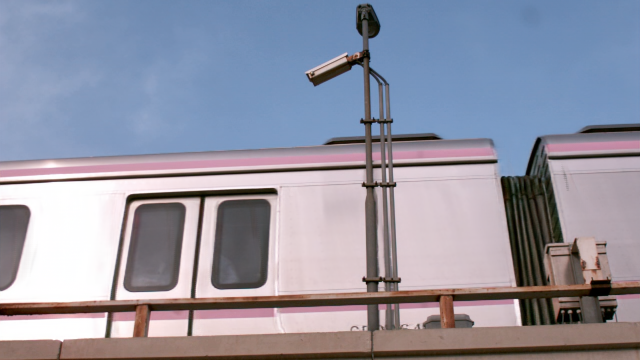}\hfill
   \\
   \vspace{0.4em}
            \begin{tikzpicture}[inner sep=0]
            \node [label={[label distance=0.25cm,text depth=-1ex,rotate=90]right: \textcolor{black}{\scriptsize {GT}}}] at (0,8.7) {};
            \end{tikzpicture}
			\includegraphics[width=\cimwid\linewidth,trim={100 10 400 275},clip]{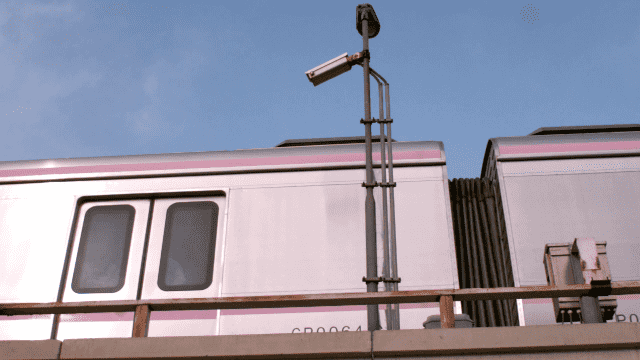}\hfill
			\includegraphics[width=\cimwid\linewidth,trim={100 10 400 275},clip]{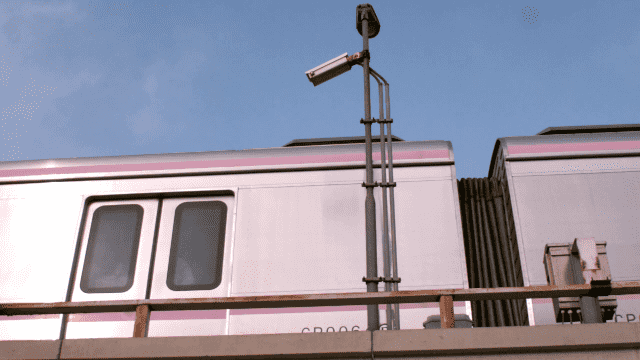}\hfill
			\includegraphics[width=\cimwid\linewidth,trim={100 10 400 275},clip]{Figs/img/NEW_sequence_gev/gt_04.png}\hfill
			\includegraphics[width=\cimwid\linewidth,trim={100 10 400 275},clip]{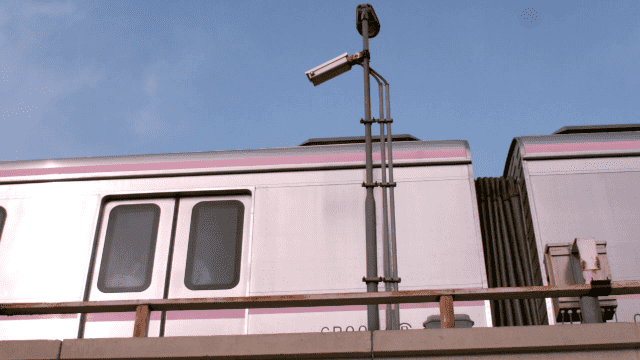}\hfill
			\includegraphics[width=\cimwid\linewidth,trim={100 10 400 275},clip]{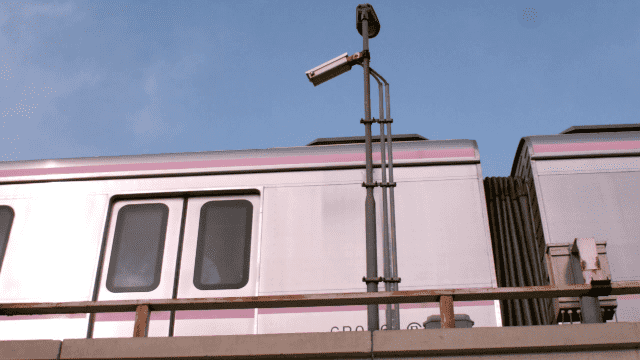}\hfill
			\includegraphics[width=\cimwid\linewidth,trim={100 10 400 275},clip]{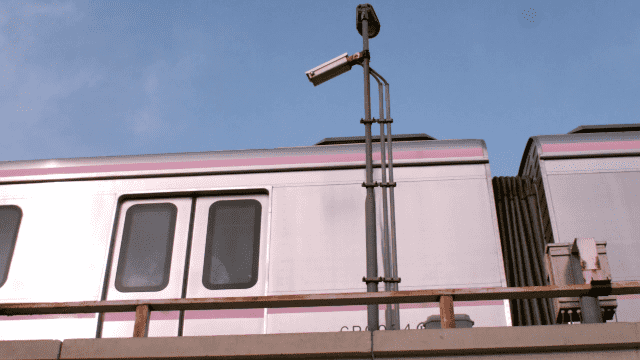}\hfill
			\includegraphics[width=\cimwid\linewidth,trim={100 10 400 275},clip]{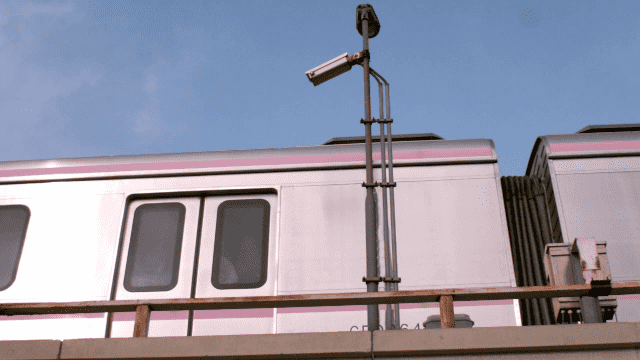}\hfill
   
   
   \vspace{0.4em}
   GS Video Sequence (From top to bottom correspond to RSSR, CVR, EvUnroll, SelfUnroll-S, SelfUnroll-M and GT)
    \end{minipage}
	\caption{Qualitative comparisons of GS video sequence restorations on the {Gev-RS}  dataset.}
	\label{mul_image}
 \vspace{.1em}
\end{figure*}

\subsubsection{Comparison on Intra-frame Reconstruction}\label{Comparison on Intra-frame Reconstruction}
%
%
%
%
%

Quantitative comparisons on the Fastec-RS and Gev-RS datasets are presented in \cref{table_fastec,tab:gev-comparison}, demonstrating that event-based RS2GS approaches, \ie, EvUnroll\cite{zhou2022evunroll} and the proposed SelfUnrolls, outperform frame-based methods by a large margin. \lin{This validates the assistance of the inter-/intra-frame information provided by events for RS correction.}
Among the event-based methods, our SelfUnrolls show 
favorable performance compared to the state-of-the-art method EvUnroll.

{The first row of \cref{suppfig-gev1} presents qualitative comparisons on the Fastec-RS dataset. It is observed that both CVR and EvUnroll can restore the foreground tree, but at expense of distorting the background building. Regarding the reconstruction on the Gev-RS dataset in the last row of \cref{suppfig-gev1}, it should be noted that CVR fails to accurately estimate the underlying RS geometry, as demonstrated by the numbers on the car. On the other hand, although EvUnroll can correct the distortion, it also introduces artifacts on the edges of restored objects, as evident in the electric bicycle.} \lin{In contrast, the proposed SelfUnrolls, \ie, SelfUnroll-S and SelfUnroll-M can rectify the edges of distorted objects and recover more realistic results, highlighting the effectiveness of the proposed algorithms.}

Furthermore, we draw attention to the intra-frame GS video reconstruction task. Quantitative and qualitative comparisons are presented in \cref{fig:sequence-graph,mul_image,tab:gev-comparison}. As shown in \cref{mul_image} (a) and (b), frame-based methods, such as RSSR and CVR, are unable to handle the dynamic scene with fast motion and fail to restore the shape. With the help of the motion and texture information that events provide, EvUnroll avoids the distortion of lines but suffers from severe artifacts (\cref{mul_image} (c)). \lin{In contrast, SelfUnrolls are able to reconstruct intra-frame GS video with high stability and outperform other methods. Specifically, SelfUnroll-M can generate more reliable textures without color distortion, such as the train windows and the brown railing in \cref{mul_image} (e)}, thanks to the effectiveness of the MOA in the information fusion of consecutive frames. 
%
%

%
%
%
 %

%
%

\begin{table*}[t]
    \centering
    \caption{Quantitative comparisons with respect to the inter-frame GS video reconstruction on the {Gev-RS} dataset. Given two RS frames with the exposure time $[0,T]$, $[T+t^{\prime},2T+t^{\prime}]$ ($t^{\prime}$ is the interval between two consecutive RS frames and is set to $\frac{2}{3}T$) respectively. All methods output 3 GS frames at timestamps $t\in \{T+0.25t^{\prime},T+0.50t^{\prime},T+0.75t^{\prime}\}$ in the task. \textcolor{red}{\textbf{Bold}} and \textcolor{blue}{\underline{underlined}} numbers represent the best and the second-best performance.}
    \begin{tabular}{l p{1.5cm}<{\centering} p{1.5cm}<{\centering} p{1.5cm}<{\centering} c p{1.5cm}<{\centering} p{1.5cm}<{\centering} p{1.5cm}<{\centering}}
\hline
         \multirow{2}{*}{Methods} & \multicolumn{3}{c}{1 frame skip} && \multicolumn{3}{c}{3 frames skip} \\ \cline{2-4} \cline{6-8}
                   & PSNR$\uparrow$ & SSIM$\uparrow$ & LPIPS$\downarrow$ && PSNR$\uparrow$ & SSIM$\uparrow$ & LPIPS$\downarrow$ \\ \hline
                   RSSR\cite{fan2021inverting} &    16.02   &   0.532   &    0.2088  &&    16.07  &    0.555   & 0.5333  \\
                   CVR\cite{fan2022context} &    23.84   &  0.760    &    0.1072  &&   23.90   &  0.763     & 0.1064  \\
                   EvUnroll\cite{zhou2022evunroll}+Timelens\cite{tulyakov2021time} &   19.17    &  0.726    &  0.1473    &&  19.29    &  0.724    &  0.1480    \\ 
               EvUnroll\cite{zhou2022evunroll}+DAIN\cite{bao2019depth}  &    19.29   &  0.688    &   0.1341   &&    19.18  &  0.680     &   0.1479   \\ 
                \myname-S\ (Ours)     &   \textcolor{blue}{\underline{30.66}}     &  \textcolor{blue}{\underline{0.903}}    &  \textcolor{blue}{\underline{0.0323}}    &&       \textcolor{blue}{\underline{30.68}}     &  \textcolor{blue}{\underline{0.903}}    &  \textcolor{blue}{\underline{0.0323}}     \\
                \myname-M\ (Ours)  &  \textcolor{red}{\textbf{31.73}}     &  \textcolor{red}{\textbf{0.919}}     &  \textcolor{red}{\textbf{0.0315}}    &&       \textcolor{red}{\textbf{31.77}}     &  \textcolor{red}{\textbf{0.920}}     &  \textcolor{red}{\textbf{0.0311}}       \\ \hline
    \end{tabular}
    \label{table_interframe}
\end{table*}

\def\imwidth{0.165}
\def\ssxxsr{(-0.42,0.58)} 
\def\ssxxsz{(-0.3,-0.10)} 

\def\newssxxs{(1.00,0.1)} 
\def\newssxxsr{(0.63,-0.250)} 
\def\newssxxsz{(0.95,-0.75)}

\def\newssyys{(1.45,-1.6)}
\def\newssyysr{(-0.1,-1.6)}
\def\ssyys{(1.45,-1.6)}
\def\ssyysr{(-0.1,-1.6)}

\def\hqfred{(0.8,-0.5)}
\def\hqfgreen{(0.2,-0.8)}

\def\hqfredz{(0.725,-0.507)}
\def\hqfgreenz{(0.095,-0.82)}

\def\hqfredpos{(1.8,-2.35)}
\def\hqfgreenpos{(-0.05,-2.35)}

\def\ssizz{1.35cm}
\def\newssizz {1.35cm}
\def\sswidth{0.245\textwidth}
\def\ssmag{5}
\def\newssmag{7}

\begin{figure*}[!htp]
\footnotesize
	\centering

  \begin{minipage}[t]{\imwidth\linewidth}
    		\centering
    		\begin{tikzpicture}[spy using outlines={green,magnification=\ssmag,size=\ssizz},inner sep=0]
				\node {\includegraphics[width=\linewidth]{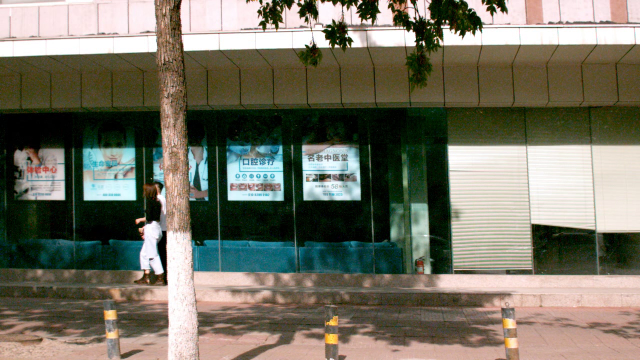}};
				\spy on \ssxxsr in node [left] at \ssyysr;
				\spy [red] on \ssxxsz in node [left,red] at \ssyys;
				\end{tikzpicture}
            RS image \\
			(PSNR, SSIM, LPIPS)\vspace{0.5em}
    	\end{minipage}%
    \hspace*{0mm}
    	\begin{minipage}[t]{\imwidth\linewidth}
    		\centering
    		\begin{tikzpicture}[spy using outlines={green,magnification=\ssmag,size=\ssizz},inner sep=0]
				\node {\includegraphics[width=\linewidth]{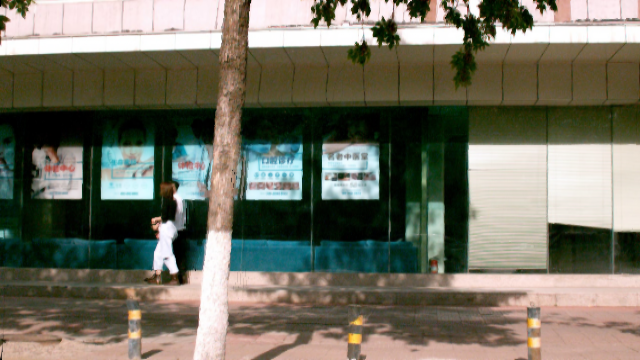}};
				\spy on \ssxxsr in node [left] at \ssyysr;
				\spy [red] on \ssxxsz in node [left,red] at \ssyys;
				\end{tikzpicture}
			CVR\\
			(16.21, 0.511, 0.1708)\vspace{0.5em}
    	\end{minipage}%
    \hspace*{0mm}
    \begin{minipage}[t]{\imwidth\linewidth}
    		\centering
    		\begin{tikzpicture}[spy using outlines={green,magnification=\ssmag,size=\ssizz},inner sep=0]
				\node {\includegraphics[width=\linewidth]{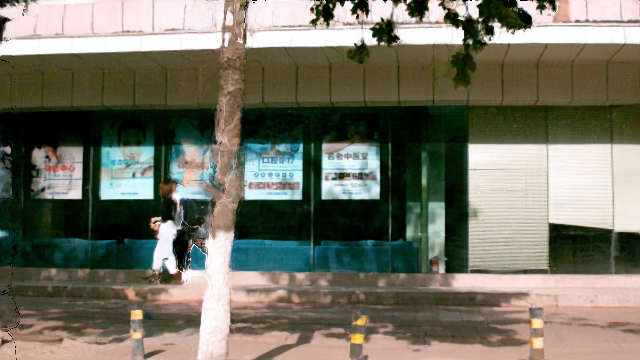}};
				\spy on \ssxxsr in node [left] at \ssyysr;
				\spy [red] on \ssxxsz in node [left,red] at \ssyys;
				\end{tikzpicture}
			
			EvUnroll + Timelens	\\
			(16.26, 0.500, 0.1503)\vspace{0.5em}
    	\end{minipage}%
   \hspace*{0mm}
\begin{minipage}[t]{\imwidth\linewidth}
    		\centering
			\begin{tikzpicture}[spy using outlines={green,magnification=\ssmag,size=\ssizz},inner sep=0]
				\node {\includegraphics[width=\linewidth]{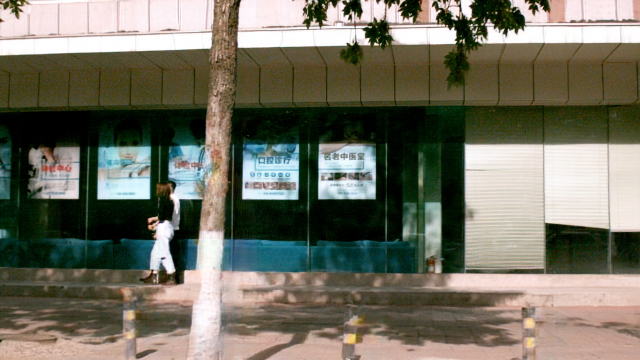}};
				\spy on \ssxxsr in node [left] at \ssyysr;
				\spy [red] on \ssxxsz in node [left,red] at \ssyys;
				\end{tikzpicture}
            SelfUnroll-S \\
			(\underline{27.99}, \underline{0.827}, \underline{0.0522})\vspace{0.5em}
    	\end{minipage}%
    \hspace*{0mm}
    	\begin{minipage}[t]{\imwidth\linewidth}
    		\centering
    		\begin{tikzpicture}[spy using outlines={green,magnification=\ssmag,size=\ssizz},inner sep=0]
				\node {\includegraphics[width=\linewidth]{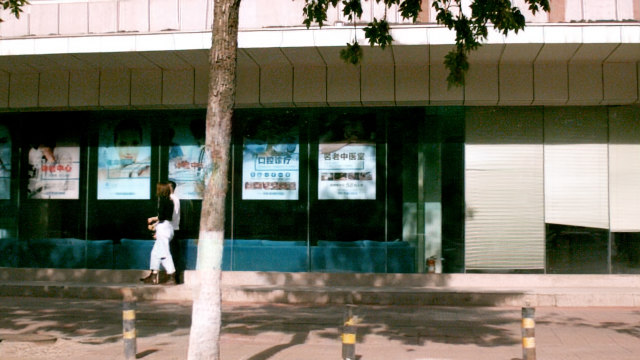}};
				\spy on \ssxxsr in node [left] at \ssyysr;
				\spy [red] on \ssxxsz in node [left,red] at \ssyys;
				\end{tikzpicture}
			
			SelfUnroll-M	\\
   (\textbf{29.11}, \textbf{0.869}, \textbf{0.0463})
			\vspace{0.5em}
    	\end{minipage}%
    \hspace*{0mm}
    \begin{minipage}[t]{\imwidth\linewidth}
    		\centering
    		\begin{tikzpicture}[spy using outlines={green,magnification=\ssmag,size=\ssizz},inner sep=0]
				\node {\includegraphics[width=\linewidth]{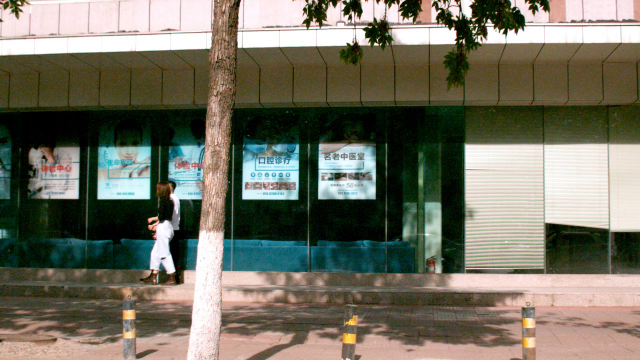}};
				\spy on \ssxxsr in node [left] at \ssyysr;
				\spy [red] on \ssxxsz in node [left,red] at \ssyys;
				\end{tikzpicture}
			
			GT	\\
			(Inf., 1.000, 0.0000)\vspace{0.5em}
    	\end{minipage}%

	
	\caption{Single GS frame reconstruction of inter-frame time on the Gev-RS dataset.}
	\label{fig:intervideo}
\end{figure*}
\input{Figs/Gev-RS-Real-img.tex}

 %
%
%

 %
%
%
%
%

\subsubsection{Comparison on Inter-frame Reconstruction}\label{Comparison on Inter-frame Reconstruction}

\def\imwidth{0.23}
\def\cimwid{0.135}

\def\zuoxia{(-0.77,-0.58)}
\def\youshang{(0.19,0.69)}

\def\ssyy{(-0.8,-0.85)}
\def\ssizz{0.5cm}
\def\sswidth{0.245\textwidth}
\def\ssmag{3}
\def\scc{(2.12,1.4)}
\def\leftarea{10}
\def\bottomarea{10}
\def\rightarea{10}
\def\toparea{10}

 \def\areaddd{{40px 10px 140px 30px}}
\begin{figure}[t]
\footnotesize
	\centering
        \hspace*{0.3cm}
    	\begin{minipage}[t]{\imwidth\linewidth}
    		\centering
			\begin{tikzpicture}[spy using outlines={rectangle,green,magnification=\ssmag,size=\ssizz},inner sep=0]
				\node {\includegraphics[width=\linewidth]{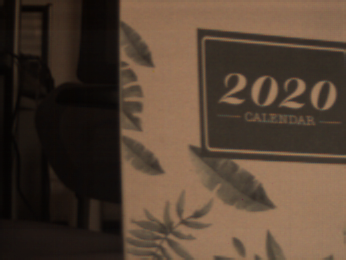}};
				\draw[green] \zuoxia rectangle \youshang;
				\end{tikzpicture}
            RS Image\vspace{0.3em}
    	\end{minipage}%
    \hfill
    	\begin{minipage}[t]{\imwidth\linewidth}
    		\centering
    		\begin{tikzpicture}[spy using outlines={rectangle,green,magnification=\ssmag,size=\ssizz},inner sep=0]
				\node {\includegraphics[width=\linewidth]{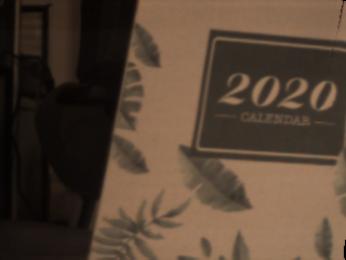}};
				\draw[green] \zuoxia rectangle \youshang;
				\end{tikzpicture}
			
			CVR\vspace{0.5em}
    	\end{minipage}%
    \hfill
    	\begin{minipage}[t]{\imwidth\linewidth}
    		\centering
    	    \begin{tikzpicture}[spy using outlines={green,magnification=\ssmag,size=\ssizz},inner sep=0]
				\node {\includegraphics[width=\linewidth]{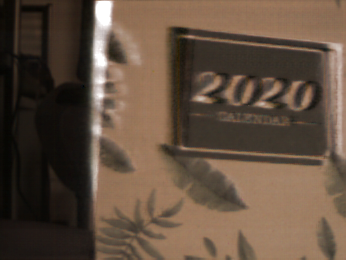}};
				\draw[green] \zuoxia rectangle \youshang;
				\end{tikzpicture}
			EvUnroll\vspace{0.5em}
    	\end{minipage}%
    \hfill
    	\begin{minipage}[t]{\imwidth\linewidth}
    		\centering
    	    \begin{tikzpicture}[spy using outlines={green,magnification=\ssmag,size=\ssizz},inner sep=0]
				\node {\includegraphics[width=\linewidth]{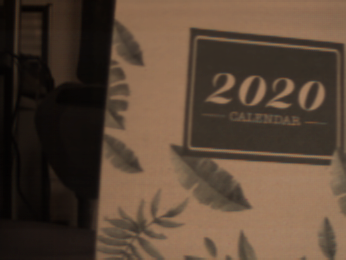}};
				\draw[green] \zuoxia rectangle \youshang;
				\end{tikzpicture}
			SelUnroll-M\vspace{0.5em}
    	\end{minipage}%
     \hfill
     \\
            \vspace*{1.mm}
            \begin{tikzpicture}[inner sep=0]
   \tikzstyle{every node}=[font=\scriptsize]
            \node[
            label= {[label distance=-1.3cm,text width=1cm,text depth=0.3ex,align=center,rotate = 90] left:\textcolor{black}{ {CVR}}}]{};
            \end{tikzpicture}
            \includegraphics[width=\cimwid\linewidth,trim={40px 10px 140px 30px},clip]{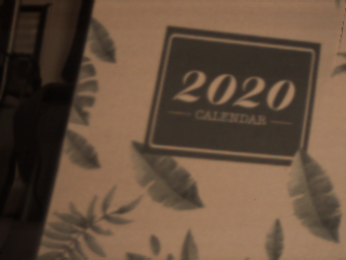}\hfill
			\includegraphics[width=\cimwid\linewidth,trim={40px 10px 140px 30px},clip]{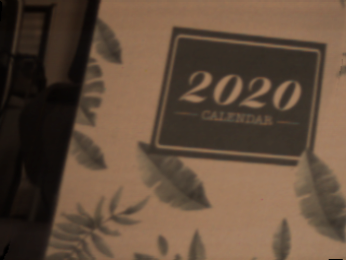}\hfill
			\includegraphics[width=\cimwid\linewidth,trim={40px 10px 140px 30px},clip]{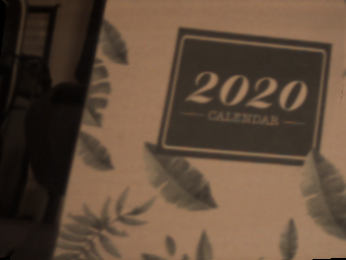}\hfill
			\includegraphics[width=\cimwid\linewidth,trim={40px 10px 140px 30px},clip]{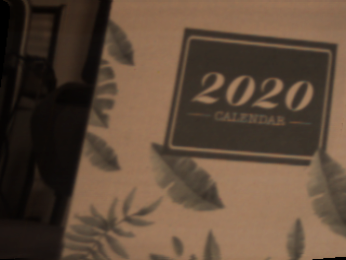}\hfill
       \includegraphics[width=\cimwid\linewidth,trim={40px 10px 140px 30px},clip]{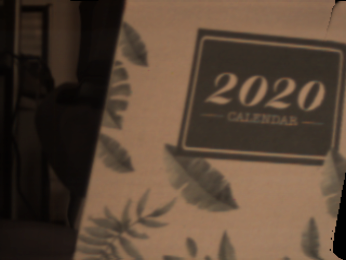}\hfill
			\includegraphics[width=\cimwid\linewidth,trim={40px 10px 140px 30px},clip]{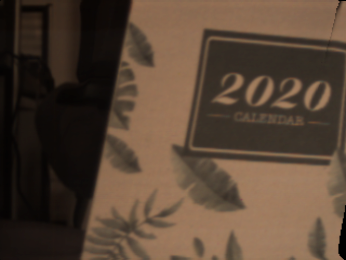}\hfill
                \includegraphics[width=\cimwid\linewidth,trim={40px 10px 140px 30px},clip]{Figs/img/dre-seq/cvr/ncvr7.png}\hfill
                \\
                \vspace*{1.mm}
			\begin{tikzpicture}[inner sep=0]
   \tikzstyle{every node}=[font=\scriptsize]
            \node[
            label= {[label distance=-1.3cm,text width=1cm,text depth=0.3ex,align=center,rotate = 90] left:\textcolor{black}{ {EvUnroll }}}]{};
            \end{tikzpicture}
            \includegraphics[width=\cimwid\linewidth,trim={40px 10px 140px 30px},clip]{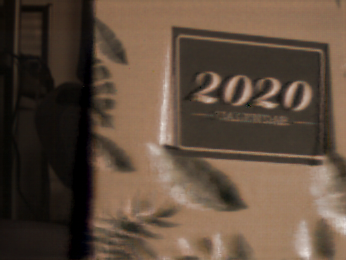}\hfill
			\includegraphics[width=\cimwid\linewidth,trim={40px 10px 140px 30px},clip]{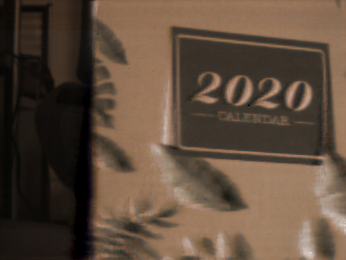}\hfill
			\includegraphics[width=\cimwid\linewidth,trim={40px 10px 140px 30px},clip]{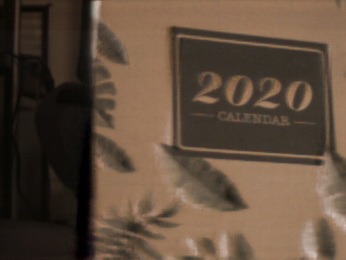}\hfill
			\includegraphics[width=\cimwid\linewidth,trim={40px 10px 140px 30px},clip]{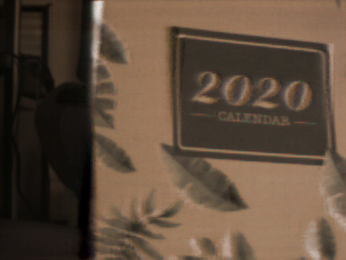}\hfill
			\includegraphics[width=\cimwid\linewidth,trim={40px 10px 140px 30px},clip]{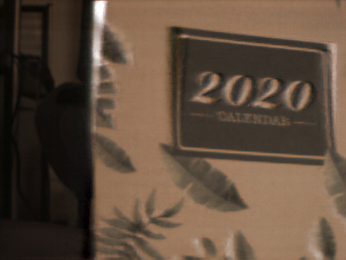}\hfill
            \includegraphics[width=\cimwid\linewidth,trim={40px 10px 140px 30px},clip]{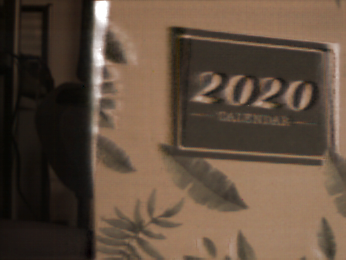}\hfill
                \includegraphics[width=\cimwid\linewidth,trim={40px 10px 140px 30px},clip]{Figs/img/dre-seq/evunroll/rs-2023_03_12_13_50_01_000094_0_875.0.png}\hfill
                \\
                \vspace*{1.mm}
            \begin{tikzpicture}[inner sep=0]
   \tikzstyle{every node}=[font=\scriptsize]
            \node[
            label= {[label distance=-0.3cm,text width=1.5cm,text depth=0.3ex,align=left,rotate = 90] left:\textcolor{black}{ {SelfUnroll-M}}}]{};
            \end{tikzpicture}
            \includegraphics[width=\cimwid\linewidth,trim={40px 10px 140px 30px},clip]{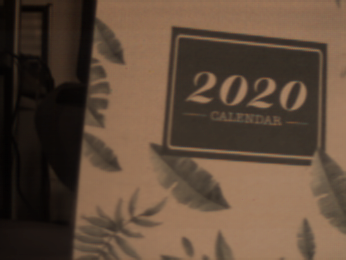}\hfill
			\includegraphics[width=\cimwid\linewidth,trim={40px 10px 140px 30px},clip]{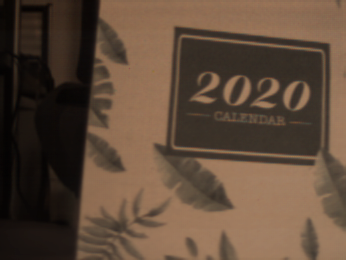}\hfill
			\includegraphics[width=\cimwid\linewidth,trim={40px 10px 140px 30px},clip]{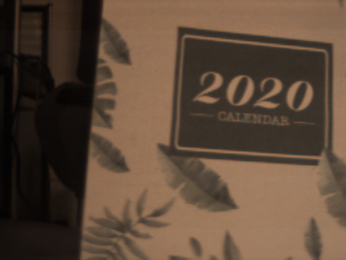}\hfill
			\includegraphics[width=\cimwid\linewidth,trim={40px 10px 140px 30px},clip]{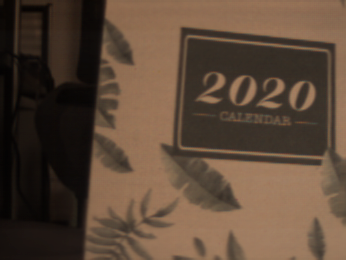}\hfill
			\includegraphics[width=\cimwid\linewidth,trim={40px 10px 140px 30px},clip]{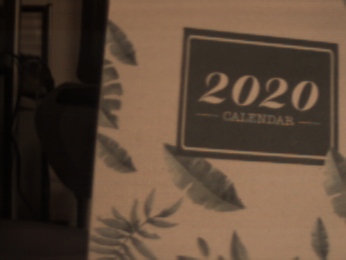}\hfill
            \includegraphics[width=\cimwid\linewidth,trim={40px 10px 140px 30px},clip]{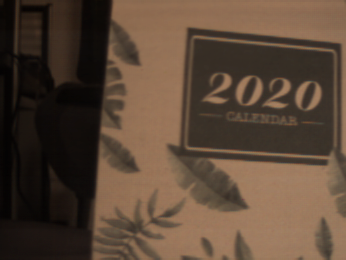}\hfill
                \includegraphics[width=\cimwid\linewidth,trim={40px 10px 140px 30px},clip]{Figs/img/dre-seq/our/rs-2023_03_12_13_50_01_000093-res_88.png}\hfill
                \\
                \vspace*{1.mm}
            \begin{tikzpicture}[inner sep=0]
   \tikzstyle{every node}=[font=\scriptsize]
            \node[
            label= {[label distance=-1.3cm,text width=1cm,text depth=0.3ex,align=center,rotate = 90] left:\textcolor{black}{ {Event}}}]{};
            \end{tikzpicture}
            \includegraphics[width=\cimwid\linewidth,trim={40px 10px 140px 30px},clip,cframe=black .01mm]{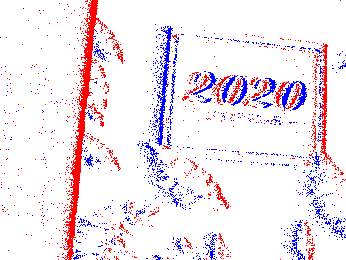}\hfill
			\includegraphics[width=\cimwid\linewidth,trim={40px 10px 140px 30px},clip,cframe=black .01mm]{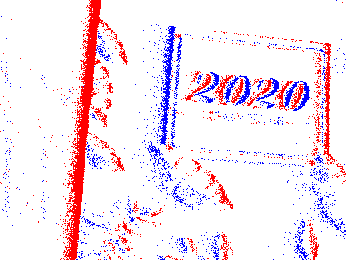}\hfill
			\includegraphics[width=\cimwid\linewidth,trim={40px 10px 140px 30px},clip,cframe=black .01mm]{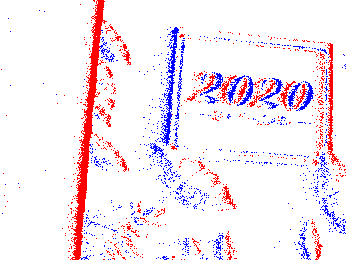}\hfill
			\includegraphics[width=\cimwid\linewidth,trim={40px 10px 140px 30px},clip,cframe=black .01mm]{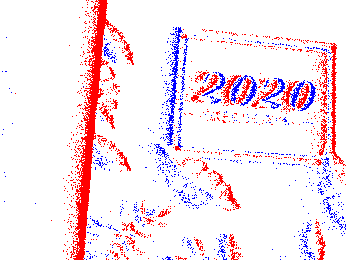}\hfill
			\includegraphics[width=\cimwid\linewidth,trim={40px 10px 140px 30px},clip,cframe=black .01mm]{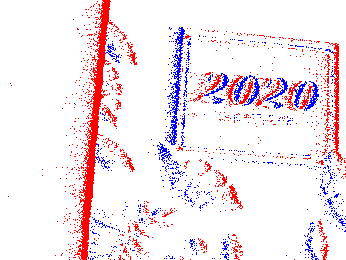}\hfill
            \includegraphics[width=\cimwid\linewidth,trim={40px 10px 140px 30px},clip,cframe=black .01mm]{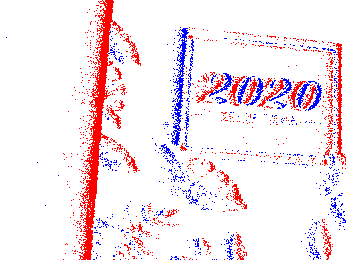}\hfill
                \includegraphics[width=\cimwid\linewidth,trim={40px 10px 140px 30px},clip,cframe=black .01mm]{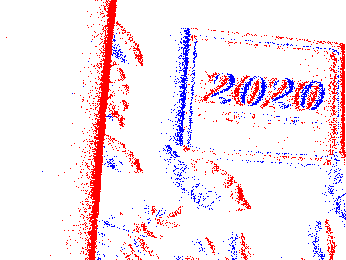}\hfill
 \vspace{.1em}
	\caption{Qualitative comparison on video sequence restoration over the DRE dataset. The first row indicates the RS image and the single GS frame recoveries respectively by CVR, EvUnroll, and SelfUnroll-M. Event frames are shown as the references of the correct GS edges.}
	\label{dre_mul_image}
\end{figure}

The quantitative comparisons on the inter-frame GS video reconstruction task are depicted in \cref{table_interframe} to evaluate the performance of SelfUnrolls. The results demonstrate that SelfUnroll-S and SelfUnroll-M outperform the frame-based methods, RSSR and CVR, by a large margin, highlighting the importance of events. \wang{In addition, we compare SelfUnrolls with cascade approaches that combine EvUnroll\cite{zhou2022evunroll} with state-of-the-art video frame interpolation methods, \ie, Timelens\cite{tulyakov2021time} and DAIN\cite{bao2019depth}, denoting as EvUnroll+Timelens and EvUnroll+DAIN respectively. The RS frames are first corrected to the GS frames corresponding to the middle scanline by EvUnroll and then interpolated by Timelens or DAIN. As shown in \cref{fig:intervideo}, CVR fails to achieve accurate correction due to the lack of inter-frame information. As for EvUnroll+Timelens, the cascade approach experiences a significant decline in performance as the RS2GS errors propagate to the frame interpolation stage.} In contrast, SelfUnrolls use E-ICs to transition between input and output images at any arbitrary moment rather than a specific time period, thereby maintaining high-quality RS2GS reconstruction during the inter-frame time. In consideration of the performance of SelfUnrolls, the advantage of SelfUnroll-M indicates the importance of consecutive frame information in inter-frame reconstruction.

\def\imwidth{0.182}
\def\cimwid{0.14}

\def\zuoxia{(-0.3,-0.9)}
\def\youshang{(0.2,0.45)}

\def\ssyy{(-0.8,-0.85)}
\def\ssizz{1.45cm}
\def\sswidth{0.245\textwidth}
\def\ssmag{5}
\def\scc{(2.12,1.4)}

\def\newssxxsr{(-0.3,-0.4)} 
\def\newssxxsz{(0.3,-0.2)} 
\def\ssyys{(1.60,-1.75)}
\def\ssyysr{(-0.13,-1.75)}

\begin{figure*}[t]
\footnotesize
	\centering
                
    \hspace*{.6cm}
    	\begin{minipage}[t]{\imwidth\linewidth}
    		\centering
			\begin{tikzpicture}[spy using outlines={rectangle,green,magnification=\ssmag,size=\ssizz},inner sep=0]
				\node {\includegraphics[width=\linewidth]{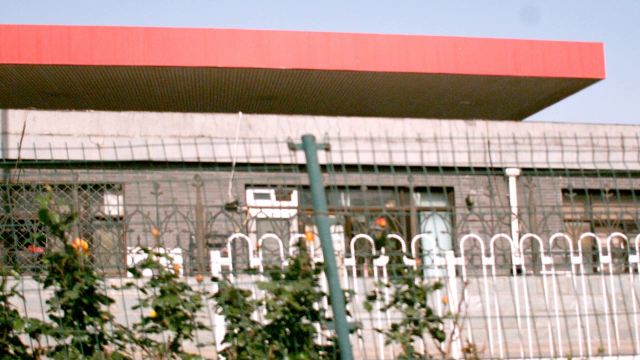}};
				\spy on \newssxxsr in node [left] at \ssyysr;
				\spy [red] on \newssxxsz in node [left,red] at \ssyys;
				\end{tikzpicture}
            RS image\\
			(PSNR, SSIM, LPIPS) \vspace{0.5em}
    	\end{minipage}%
    \hfill
    	\begin{minipage}[t]{\imwidth\linewidth}
    		\centering
    		\begin{tikzpicture}[spy using outlines={rectangle,green,magnification=\ssmag,size=\ssizz},inner sep=0]
				\node {\includegraphics[width=\linewidth]{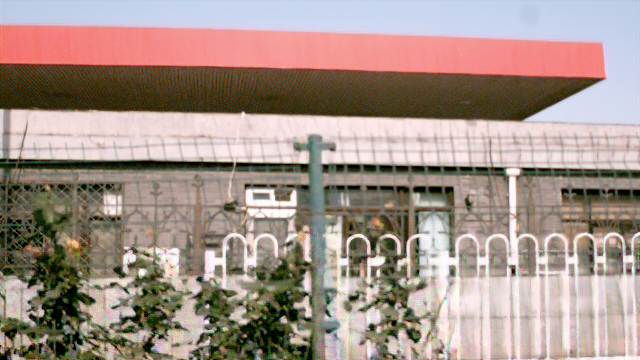}};
                \spy on \newssxxsr in node [left] at \ssyysr;
				\spy [red] on \newssxxsz in node [left,red] at \ssyys;
				\end{tikzpicture}
			
			Temporal E-IC\\
			(25.56, 0.808, 0.0920)\vspace{0.5em}
    	\end{minipage}%
    \hfill
    	\begin{minipage}[t]{\imwidth\linewidth}
    		\centering
    	    \begin{tikzpicture}[spy using outlines={green,magnification=\ssmag,size=\ssizz},inner sep=0]
				\node {\includegraphics[width=\linewidth]{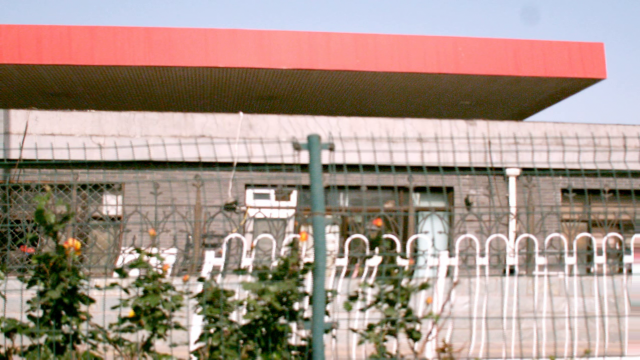}};
				\spy on \newssxxsr in node [left] at \ssyysr;
				\spy [red] on \newssxxsz in node [left,red] at \ssyys;
				\end{tikzpicture}
			Spatial E-IC\\
			(21.35, 0.730, 0.0734)\vspace{0.5em}
    	\end{minipage}%
    \hfill
    	\begin{minipage}[t]{\imwidth\linewidth}
    		\centering
    	    \begin{tikzpicture}[spy using outlines={green,magnification=\ssmag,size=\ssizz},inner sep=0]
				\node {\includegraphics[width=\linewidth]{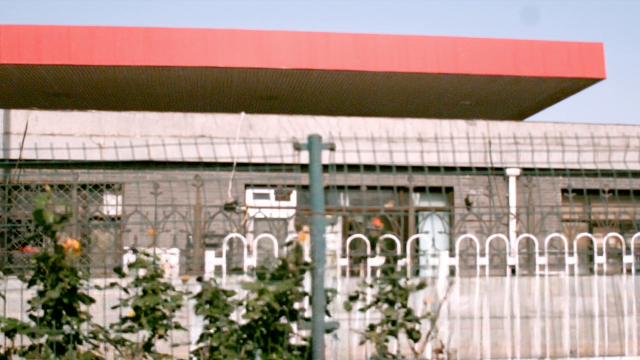}};
				\spy on \newssxxsr in node [left] at \ssyysr;
				\spy [red] on \newssxxsz in node [left,red] at \ssyys;
				\end{tikzpicture}
			Spatial \& Temporal E-IC\\
			(\textbf{27.05},  \textbf{0.852}, \textbf{0.0585})\vspace{0.5em}
    	\end{minipage}%
    \hfill
		\begin{minipage}[t]{\imwidth\linewidth}
			\centering
			\begin{tikzpicture}[spy using outlines={green,magnification=\ssmag,size=\ssizz},inner sep=0]
				\node {\includegraphics[width=\linewidth]{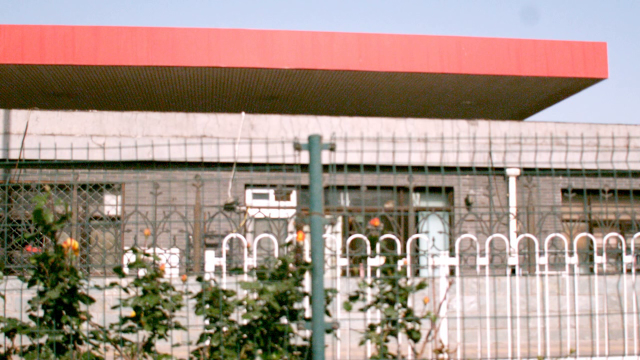}};
				\spy on \newssxxsr in node [left] at \ssyysr;
				\spy [red] on \newssxxsz in node [left,red] at \ssyys;
				\end{tikzpicture}
			GT\\
			(Inf.,  1.000, 0.0000)\vspace{.5em}
			
			
		\end{minipage}%
  \hspace*{.6cm}
  \caption{Qualitative ablations of spatial and temporal E-ICs in \myname.}\vspace{-1em}
    \label{Ablation_module}
  \end{figure*}

\subsection{Comparison on the Real-world Datasets}\label{Comparison on the Real-world Datasets}

We also conduct experiments on real-world datasets, \ie, Gev-RS-Real~\cite{zhou2022evunroll} and DRE in \cref{sec:datasets}. Qualitative comparisons are made to the state-of-the-art frame-based method CVR \cite{fan2022context} and event-based method EvUnroll\cite{zhou2022evunroll}. 
\cref{fig:gev-real} illustrate comparisons on the single frame restoration, while the performance on the video reconstruction is shown in \cref{mul_image_real,dre_mul_image} respectively. We can observe that both CVR and EvUnroll are trained on synthetic datasets and thus suffer from degraded performance in real-world scenes due to the "synthetic-to-
real" gap.
Specifically, CVR produces inaccurate rectification and shape distortions due to violated motion assumptions, while EvUnroll yields obvious artifacts and blurred details due to the violation of learned modality correspondence in real-world scenes. In contrast, SelfUnrolls utilize self-supervised learning on real datasets, effectively bridging the “synthetic-to-real" gap and producing visually satisfying results without texture and shape distortion. More results on the real-world datasets can be found at \url{\website}.

\begin{table}[htb]
\caption{Ablation study of E-ICs modules on the {Gev-RS} dataset. \textcolor{red}{\textbf{Bold}} numbers represents the best performance.}
\centering
\begin{tabular}{ccccc}

\hline
Spatial & Temporal & PSNR$\uparrow$     & SSIM$\uparrow$    & LPIPS$\downarrow$   \\ \hline
\checkmark      &           & 28.67 & 0.877 & 0.0311 \\
     &\checkmark           & 31.67 & 0.916 & 0.0227 \\
\checkmark     & \checkmark          & \textcolor{red}{\textbf{32.26}}  & \textcolor{red}{\textbf{ 0.926}} & \textcolor{red}{\textbf{0.0204}} \\ \hline
\end{tabular}
\label{ablation_module}
\end{table}

\begin{table}[hbpt]
    \centering
    \caption{Ablation study of supervisions on the {Gev-RS} dataset. Case \#0 represents the model without any supervision, and the metrics of Case \#0 which are calculated between the input RS frame and the GS reference are given as a reference. \textcolor{red}{\textbf{Bold}} and \textcolor{blue}{\underline{underlined}} numbers represent the best and the second-best performance.}
    \resizebox{\linewidth}{!}{
    \begin{tabular}{c|cccccccccc}
\hline
        \multirow{2}{*}{Case} & \multirow{2}{*}{$\mathcal{L}_{lc}$} & \multirow{2}{*}{$\mathcal{L}_{cc}$} & \multirow{2}{*}{$\mathcal{L}_{tc}$} & \multicolumn{3}{c}{SelfUnroll-S/-M} \\ \cline{5-7}
                &  &  &  & PSNR$\uparrow$ & SSIM$\uparrow$ & LPIPS$\downarrow$ \\ \hline
                \#0    &        &                   &                   &   19.01/19.01    &   0.673/0.673    &   0.0751/0.0751   \\ 
            \#1    &  \checkmark      &                   &                   &   17.30/17.45    &   0.449/0.439    &   0.1110/0.1344      \\ 
             \#2    &  &           \checkmark        &                   &       19.01/19.48 &  0.670/0.671     & 0.0761/0.1102    \\
             \#3    &  &                   &        \checkmark           &      30.60/30.85 &    0.898/0.905   &  0.0277/0.0275  \\
          \#4 &\checkmark       &      \checkmark             &                   & 29.23/29.47      &  0.877/0.882     & 0.0944/0.0971      \\
           \#5   &\checkmark    &                   &     \checkmark              &  \textcolor{blue}{\underline{31.30}}/\textcolor{blue}{\underline{31.56}}     &  \textcolor{blue}{\underline{0.911}}/\textcolor{blue}{\underline{0.915}}     &  \textcolor{blue}{\underline{0.0243}}/0.0236 \\
           \#6   &    &          \checkmark         &         \checkmark          &    30.98/31.37   &  0.906/0.914     &  0.0248/\textcolor{blue}{\underline{0.0229}}   \\
           \#7 &   \checkmark   &         \checkmark          &       \checkmark            &    \textcolor{red}{\textbf{32.26}}/\textcolor{red}{\textbf{32.62}}    &   \textcolor{red}{\textbf{0.926}}/\textcolor{red}{\textbf{0.932}}     &   \textcolor{red}{\textbf{0.0204}}/\textcolor{red}{\textbf{0.0200}}  \\ \hline
        
    \end{tabular}
    }
    \label{tab:ablation_loss}
\end{table}

\input{Figs/Ablation_loss.tex}
\subsection{Ablation study}
In this subsection, we perform a diverse set of ablation studies to investigate the contribution of each loss in our self-supervised learning framework and the importance of each module of the proposed network.
\par

\subsubsection{Importance of Spatial \& Temporal E-ICs} 
The overall \myname-S network is composed of spatial and temporal E-ICs, \ie, $\operatorname{{E-IC}_{S}}$ and $\operatorname{{E-IC}_{T}}$. The spatial connection $\operatorname{{E-IC}_{S}}$ achieves RS2GS transformation by estimating per-pixel motions and is thus with brightness consistency but may suffer from geometric distortions caused by motion errors from noisy events, as shown in \cref{Ablation_module}. On the other hand, the temporal $\operatorname{{E-IC}_{T}}$ module compensates for brightness change between RS and GS images and is thus robust to motion distortions but may suffer from unrealistic artifacts and chromatic aberrations, as shown in \cref{Ablation_module}. Fusing spatial and temporal E-ICs gives the best quantitative and qualitative results, as shown in \cref{Ablation_module} and \cref{ablation_module}.

\par

\def\imwidth{0.33}

\def\ssxxs{(1.0,0.1)} 
\def\ssxxsr{(1.12,-0.65)} 

\def\ssxxsz{(1.05,0.1)} 

\def\ssyys{(1.40,-1.85)}
\def\ssyysr{(-0.15,-1.85)}

\def\hqfred{(0.8,-0.5)}
\def\hqfgreen{(0.2,-0.8)}

\def\hqfredz{(0.725,-0.507)}
\def\hqfgreenz{(0.095,-0.82)}

\def\hqfredpos{(1.8,-2.35)}
\def\hqfgreenpos{(-0.05,-2.35)}

\def\ssizz{1.3cm}
\def\sswidth{0.245\textwidth}
\def\ssmag{2}
\def\scc{(2.12,1.4)}
\begin{figure}[!t]
\footnotesize
	\centering
    	\begin{minipage}[t]{\imwidth\linewidth}
    		\centering
    		
			\begin{tikzpicture}[spy using outlines={green,magnification=\ssmag,size=\ssizz},inner sep=0]
				\node {\includegraphics[width=\linewidth]{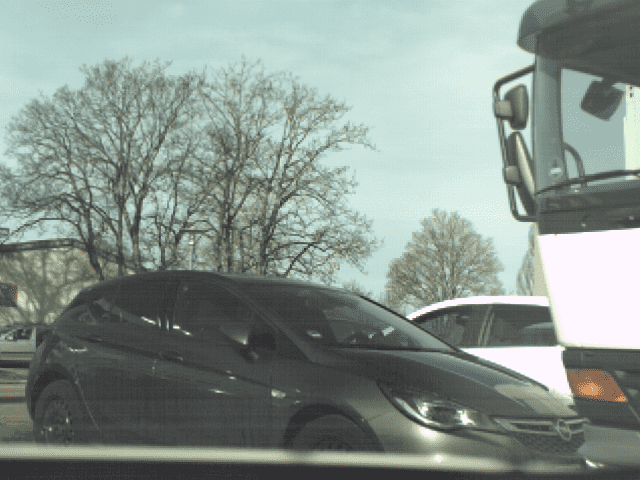}};
				\spy on \ssxxsr in node [left] at \ssyysr;
				\spy [red] on \ssxxsz in node [left,red] at \ssyys;
				\end{tikzpicture}
             RS Image $I_{1}^{RS}$\\
            (16.18, 0.599, 0.1692)\vspace{0.5em} 
    	\end{minipage}%
    \hspace*{0mm}
    	\begin{minipage}[t]{\imwidth\linewidth}
    		\centering
    		\begin{tikzpicture}[spy using outlines={green,magnification=\ssmag,size=\ssizz},inner sep=0]
				\node {\includegraphics[width=\linewidth]{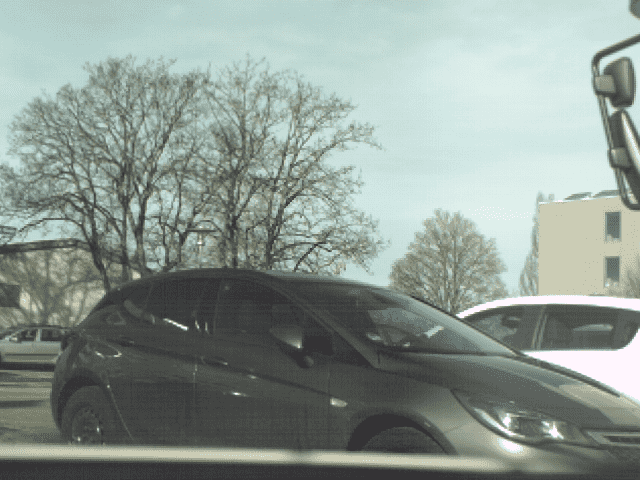}};
				\spy on \ssxxsr in node [left] at \ssyysr;
				\spy [red] on \ssxxsz in node [left,red] at \ssyys;
				\end{tikzpicture}
			 RS Image $I_{2}^{RS}$\\
			(17.04, 0.643, 0.1853)\vspace{0.5em}
			
    	\end{minipage}%
    \hspace*{0mm}
    	\begin{minipage}[t]{\imwidth\linewidth}
    		\centering
            \begin{tikzpicture}[spy using outlines={green,magnification=\ssmag,size=\ssizz},inner sep=0]
				\node {\includegraphics[width=\linewidth]{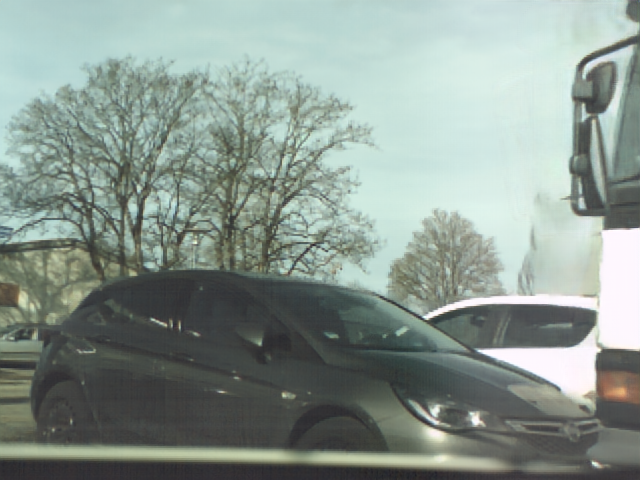}};
				\spy on \ssxxsr in node [left] at \ssyysr;
				\spy [red] on \ssxxsz in node [left,red] at \ssyys;
				\end{tikzpicture}
			SelfUnroll-S ($I_{1}^{RS}$)\\
			(30.64, 0.917, 0.0521)\vspace{0.5em}
    		
    	\end{minipage}%
    \\
    	\begin{minipage}[t]{\imwidth\linewidth}
    		\centering
    		\begin{tikzpicture}[spy using outlines={green,magnification=\ssmag,size=\ssizz},inner sep=0]
				\node {\includegraphics[width=\linewidth]{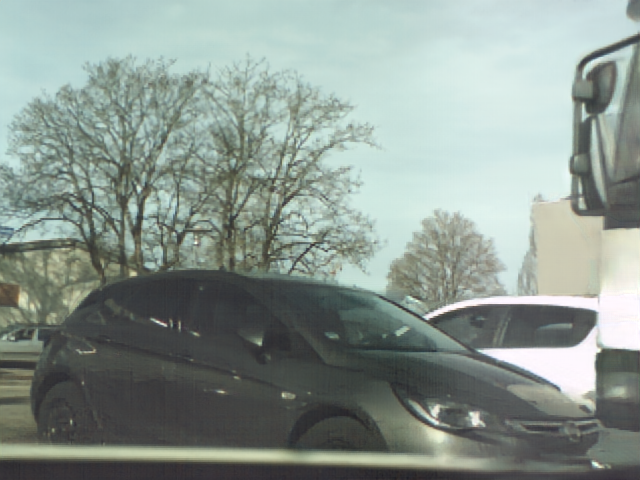}};
				\spy on \ssxxsr in node [left] at \ssyysr;
				\spy [red] on \ssxxsz in node [left,red] at \ssyys;
				\end{tikzpicture}
			SelfUnroll-S ($I_{2}^{RS}$)\\
			(29.49, 0.910, 0.0597)\vspace{0.5em}
    	\end{minipage}%
    \hspace*{0mm}
    	\begin{minipage}[t]{\imwidth\linewidth}
    		\centering
    		\begin{tikzpicture}[spy using outlines={green,magnification=\ssmag,size=\ssizz},inner sep=0]
				\node {\includegraphics[width=\linewidth]{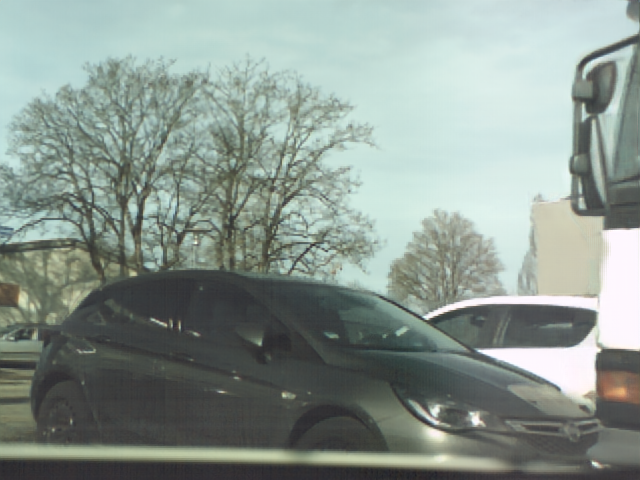}};
                \spy on \ssxxsr in node [left] at \ssyysr;
				\spy [red] on \ssxxsz in node [left,red] at \ssyys;
				\end{tikzpicture}
			SelfUnroll-M\\
			(\textbf{31.51}, \textbf{0.925}, \textbf{0.0503})\vspace{0.5em}
    	\end{minipage}%
     \hspace*{0mm}
    	\begin{minipage}[t]{\imwidth\linewidth}
    		\centering
    		\begin{tikzpicture}[spy using outlines={green,magnification=\ssmag,size=\ssizz},inner sep=0]
				\node {\includegraphics[width=\linewidth]{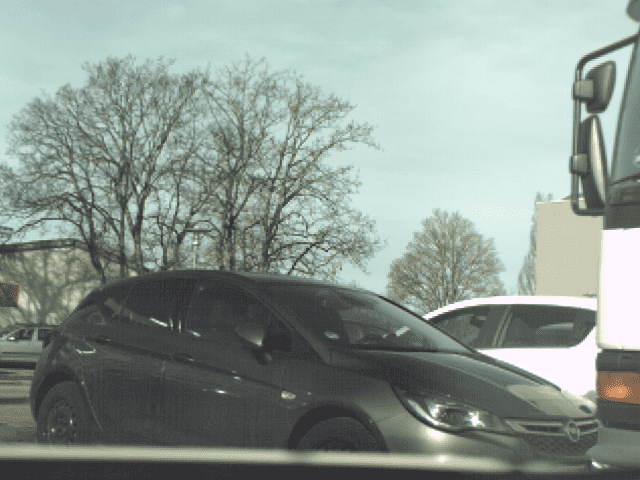}};
                \spy on \ssxxsr in node [left] at \ssyysr;
				\spy [red] on \ssxxsz in node [left,red] at \ssyys;
				\end{tikzpicture}
			GT\\
			(Inf., 1.000, 0.0000)\vspace{0.5em}
    	\end{minipage}%
    
    \centering
	\caption{Quantitative and qualitative ablations of MOA on the Faset-RS dataset. Metrics performance (\ie, PSNR$\uparrow$/SSIM$\uparrow$/LPIPS$\downarrow$) on estimating the inter-frame global-shutter image. $I_{1}^{RS}$ (previous) and $I_{2}^{RS}$ (next) are two consecutive RS frames. The occluded objects reconverd from SelfUnroll are color distorted. The overall quality of the recovered image is upgraded by MOA. }
	\label{fig:OA}
\end{figure}
\subsubsection{Necessity of Loss Combination}
From absolute differences to the ground-truth GS reference, it is observed in \cref{Ablation_loss} that removing either $\mathcal{L}_{lc}$ or $\mathcal{L}_{tc}$ results in RS distortions, indicating their contribution to RS correction. Specifically, $\mathcal{L}_{tc}$ directly introduces the determined inputs, \ie, two consecutive RS frames as the strong supervision to guide E-ICs to learn spatial and temporal transitions, which ultimately benefits the RS2GS task. {$\mathcal{L}_{lc}$ provides the only constraint in the GS domain and thus is also helpful in the RS2GS task. However, $\mathcal{L}_{lc}$ alone cannot guarantee brightness consistency as it does not impose a constraint between the outputs of E-ICs and the original images. On the other hand, although $\mathcal{L}_{cc}$ does not directly help E-ICs produce GS images, it contributes to brightness consistency and improves the performance of E-ICs.} Overall, \cref{tab:ablation_loss} and \cref{Ablation_loss} show that combining all three lossed leads to the smallest absolute error, validating the necessity of learning RS2GS with $\mathcal{L}_{lc}$, $\mathcal{L}_{cc}$, and $\mathcal{L}_{tc}$ simultaneously.

\subsubsection{Superiority of Multi-frame Fusion}
The above quantitative and qualitative analyses in \cref{Comparison on the Synthetic Datasets,Comparison on the Real-world Datasets} have demonstrated that SelfUnroll-M outperforms SelfUnroll-S due to the complementarity of the previous and next frame information. To study how the MOA module handles the information loss introduced by the occlusion, we compare the performance of SelfUnroll-S and SelfUnroll-M when facing the serious occlusions, \eg, the orange headlight and the distant building in \cref{fig:OA}. Using a single RS image ($I_{1}^{RS}$/$I_{2}^{RS}$) as input for SelfUnroll-S produces incorrect textures and colors in attempts to recover regions occluded by foreground objects in moving scenes, \eg, the disappearing building in SelfUnroll-S ($I_{1}^{RS}$) and the gray headlight in SelfUnroll-S ($I_{2}^{RS}$) in \cref{fig:OA}. The MOA module we propose can adaptively fuse the GS results generated by SelfUnroll-S ($I_{1}^{RS}$/$I_{2}^{RS}$) to alleviate the impact of occlusion.

\begin{figure}[!htp]
     \centering
    \includegraphics[width=\linewidth]{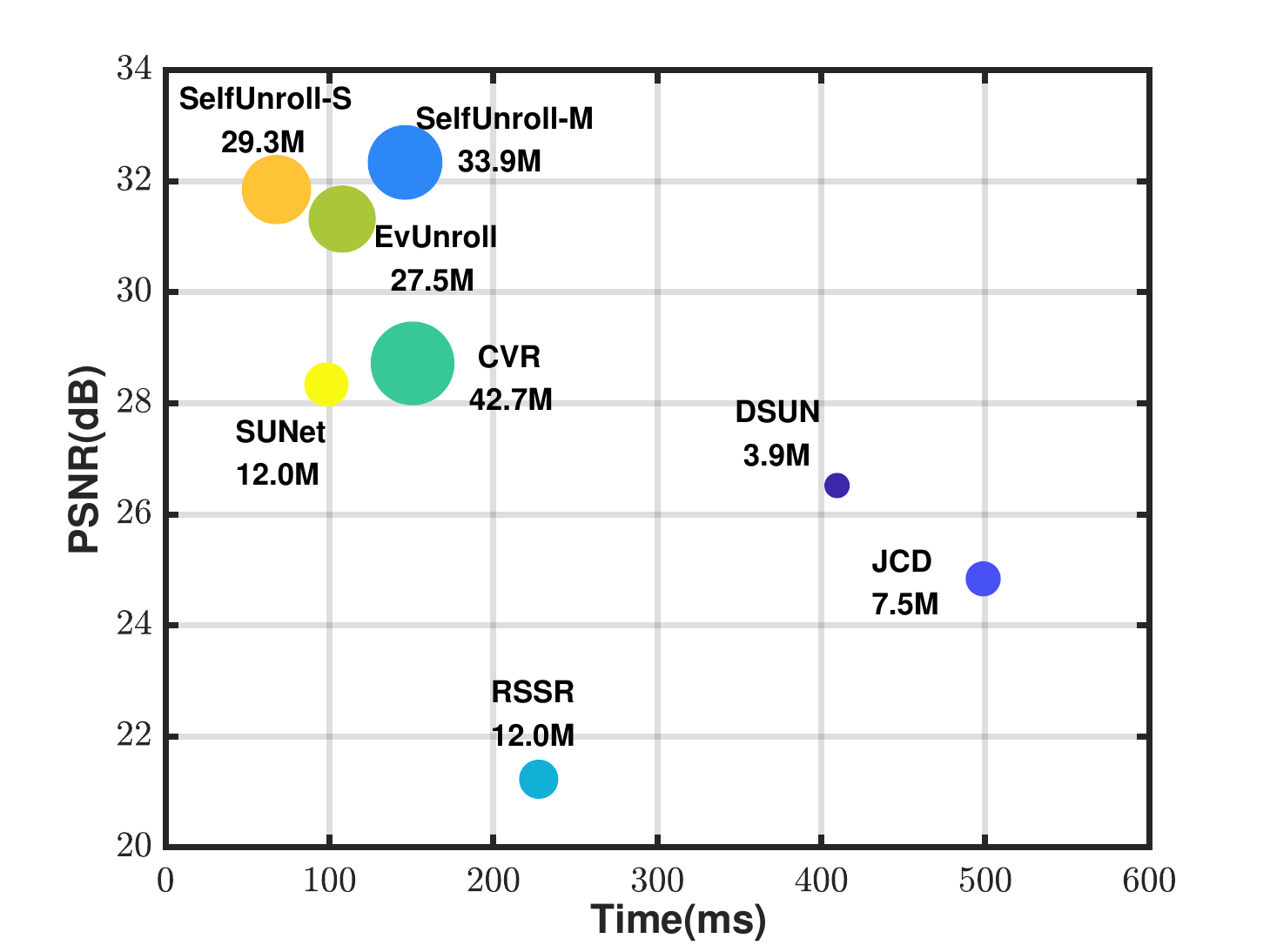}
    \caption{ \textbf{PSNR \vs Inference time, size $\propto$ parameters} on Fastec-RS dataset. The inference time is the length of time used to reconstruct a GS image with a resolution of 640$\times$480. The size of the blobs is proportional to the number of network parameters. The model in the top left corner has a better performance.}
    \label{fig:infertime}
\end{figure}

\subsection{Complexity to Performance Analysis}
We further evaluate the complexity of our proposed approaches and other RS2GS methods by feeding RS images with the same spatial resolution $640\times 480$ and implementing them over the same NVIDIA GeForce RTX 3090 GPU. The comparison results on complexity to performance diagram are given in \cref{fig:infertime}. It is shown that our SelfUnroll-S can infer GS prediction with the shortest time ($\approx 68$ ms) while SelfUnroll-M performs with increased complexity since it requires an additional MOA module as shown in \cref{fig:net-arch}. SelfUnroll-M performs better than SelfUnroll-S while both achieve state-of-the-art RS2GS performance.

Regarding the model size, CVR has the most network parameters and gives the best performance among frame-based RS2GS methods. Due to the utilization of events, the event-based methods have smaller model sizes but outperform CVR by a large margin. Among the three event-based methods, our proposed SelfUnroll-S gives higher GS reconstruction PSNR than EvUnroll even though they have comparable model sizes. The RS2GS performance can be further improved by SelfUnroll-M with increased model sizes and input frames.

 Overall, our proposed SelfUnroll-S/-M is highly effective and efficient in restoring the GS frame and facilitating the high temporal reconstruction of RS2GS.

\section{Limitations and Future Works}
 Our proposed SelfUnroll method is based on the principle of estimating the spatial and temporal transformation between two sharp images. However, it is limited in its ability to handle motion blur, as shown in the first row of \cref{fig:failure2}.  Additionally, as illustrated in the second row of \cref{fig:failure2}, our method is unable to mitigate the flicker effect\cite{lin2023deflickercyclegan}  caused by the variation in brightness during RS exposure time. While we have successfully removed the distortion, the streaks caused by the flicker effect remain present. We plan to address these issues in the future.

\def\imwidth{0.32}
\def\cimwid{0.14}

\def\zuoxia{(-0.3,-0.9)}
\def\youshang{(0.2,0.45)}

\def\ssyy{(-0.8,-0.85)}
\def\ssizz{0.5cm}
\def\sswidth{0.245\textwidth}
\def\ssmag{3}
\def\scc{(2.12,1.4)}

\begin{figure}[!htp]
\footnotesize
	\centering

  \begin{minipage}[t]{\imwidth\linewidth}
    		\centering
			\begin{tikzpicture}[spy using outlines={rectangle,green,magnification=\ssmag,size=\ssizz},inner sep=0]
				\node {\includegraphics[width=\linewidth,cframe=black .05mm]{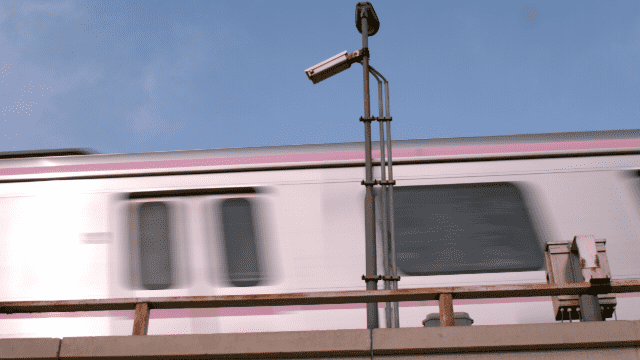}};
				\end{tikzpicture}
    	\end{minipage}%
    \hfill
    	\begin{minipage}[t]{\imwidth\linewidth}
    		\centering
    		\begin{tikzpicture}[spy using outlines={rectangle,green,magnification=\ssmag,size=\ssizz},inner sep=0]
				\node {\includegraphics[width=\linewidth,cframe=black .05mm]{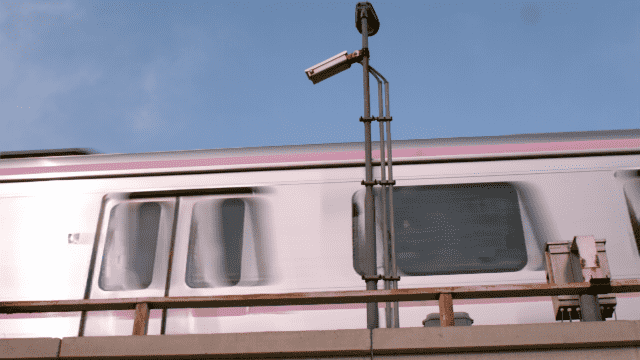}};
			\end{tikzpicture}
			
    	\end{minipage}%
    \hfill
    \begin{minipage}[t]{\imwidth\linewidth}
    		\centering
    		\begin{tikzpicture}[spy using outlines={rectangle,green,magnification=\ssmag,size=\ssizz},inner sep=0]
				\node {\includegraphics[width=\linewidth,cframe=black .05mm]{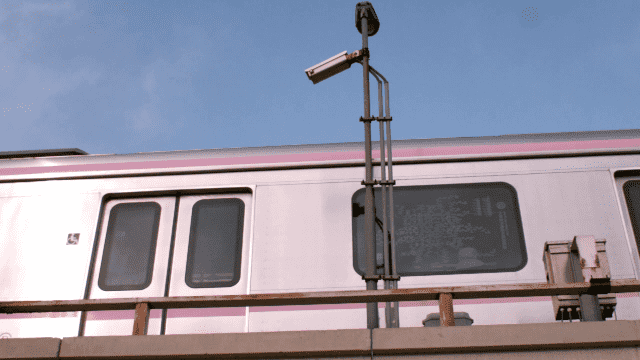}};
			\end{tikzpicture}
			
    	\end{minipage}%
   
\vspace{.5mm}

\begin{minipage}[t]{\imwidth\linewidth}
    		\centering
			\begin{tikzpicture}[spy using outlines={rectangle,green,magnification=\ssmag,size=\ssizz},inner sep=0]
				\node {\includegraphics[width=\linewidth,cframe=black .05mm]{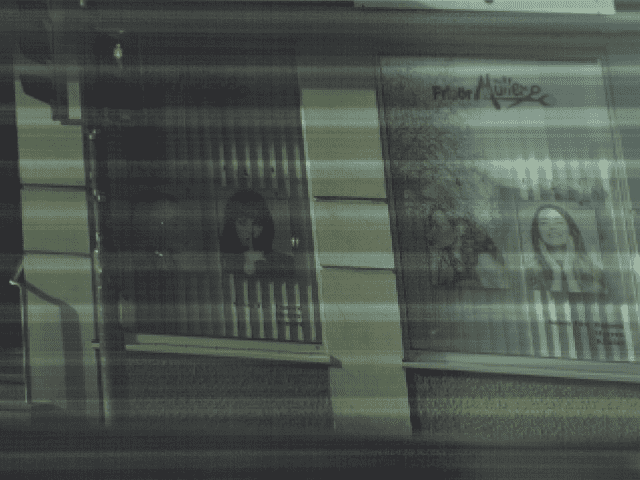}};
				\end{tikzpicture}
            (a) RS image \vspace{0.5em}
    	\end{minipage}%
    \hfill
    	\begin{minipage}[t]{\imwidth\linewidth}
    		\centering
    		\begin{tikzpicture}[spy using outlines={rectangle,green,magnification=\ssmag,size=\ssizz},inner sep=0]
				\node {\includegraphics[width=\linewidth,cframe=black .05mm]{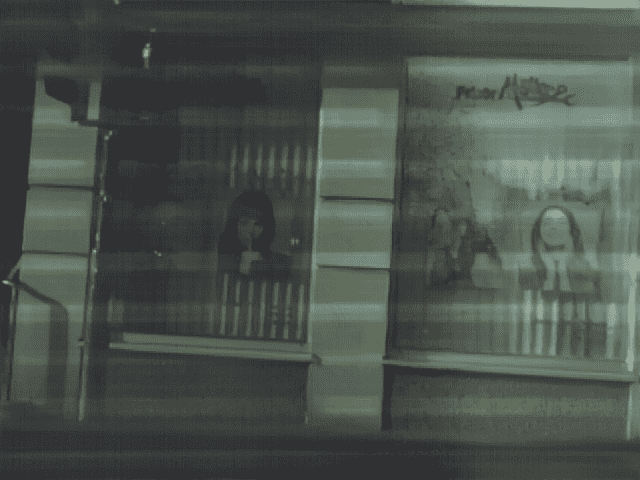}};
			\end{tikzpicture}
			
			(b) SelfUnroll-M	\vspace{0.5em}
    	\end{minipage}%
    \hfill
    \begin{minipage}[t]{\imwidth\linewidth}
    		\centering
    		\begin{tikzpicture}[spy using outlines={rectangle,green,magnification=\ssmag,size=\ssizz},inner sep=0]
				\node {\includegraphics[width=\linewidth,cframe=black .05mm]{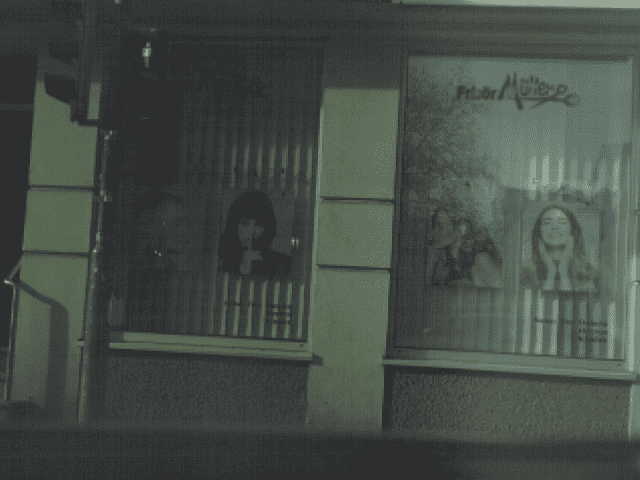}};
			\end{tikzpicture}
			
			(c) GT	\vspace{0.5em}
    	\end{minipage}%

      \vspace{-1em}
	\caption{Failure cases: a blurry RS image with a train of high-speed motion (the 1$^{st}$ row) and an RS image with light flicker effect (the 2$^{nd}$ row).}\vspace{-1em}
	\label{fig:failure2}
\end{figure}

\section{Conclusion}
In this paper, we propose a novel event-based RS2GS method SelfUnroll to simultaneously achieve the RS image correction and  the high temporal GS video reconstruction. Our method is based on the Event-based Inter/Intra-frame Compensator (E-IC), which establishes a unified spatial and temporal connection between two images with different exposures, thus improving the RS2GS transformation. Based on E-IC, we first propose the SelfUnroll-S network to restore GS images from a single RS image with the aid of events. 
Then we extend SelfUnroll-S to SelfUnroll-M where the Motion and Occlusion Aware (MOA) module is designed to tackle the occlusion problem by fusing the temporal information.
We further propose a self-supervised learning framework for both SelfUnroll-S and SelfUnroll-M that allows for learning with real events and RS images, thereby reducing the cost of collecting ground-truth GS images and bridging the performance gap between synthetic and real datasets.
To validate the effectiveness of our proposed methods, we develop a real-world RS image dataset that contains both events and RS frames for various indoor and outdoor scenes. Our experimental results on synthetic and real datasets demonstrate the superiority of our proposed SelfUnroll methods over existing state-of-the-art methods.


%





\ifCLASSOPTIONcaptionsoff
  \newpage
\fi



%

\bibliographystyle{ieee_fullname}

\end{document}